\documentclass[nonacm, format=acmsmall, review=false, screen=true]{acmart}

\usepackage{pdflscape}
\usepackage{url}
\usepackage{verbatim}

\usepackage{graphicx}
\usepackage{subfig}
\usepackage{refcount}
\usepackage{footnote}
\usepackage{multirow}
\usepackage{listings}
\usepackage{multicol}
\usepackage{framed}
\usepackage{listings}
\usepackage[skipbelow=\topskip,skipabove=\topskip]{mdframed}
\mdfsetup{roundcorner=0}
\usepackage{etoolbox}
\usepackage{eurosym}
\usepackage{pifont}
\usepackage{algorithmicx}
\usepackage{algorithm}
\usepackage{algpseudocode}

\usepackage{enumitem}
\usepackage{textcomp}
\usepackage{mathtools}

\DeclareMathOperator*{\argmin}{argmin}

\def\itemrange#1{%
\let\oldlabelenumi=\labelenumi
\addtocounter{enumi}{1}%
\edef\labelenumi{(\theenumi--\noexpand\theenumi)}%
\addtocounter{enumi}{-1}%
\addtocounter{enumi}{#1}%
\item
\let\labelenumi=\oldlabelenumi}

\algnewcommand\algorithmicinput{\textbf{Input:}}
\algnewcommand\INPUT{\item[\algorithmicinput]}

\algnewcommand\algorithmicoutput{\textbf{Output:}}
\algnewcommand\OUTPUT{\item[\algorithmicoutput]}

\acmJournal{CSUR}

\setcopyright{acmlicensed}

\begin{document}

\title{AutonoML: Towards an Integrated Framework for Autonomous Machine Learning}
\subtitle{A Comprehensive and Synthesising Review of Concepts in AutoML Research and Beyond}

\author{David Jacob Kedziora}
\affiliation{%
  \institution{Complex Adaptive Systems Lab, Data Science Institute, University of Technology Sydney}
  \city{Sydney}
  \state{New South Wales}
  \postcode{2007}
  \country{Australia}
}
\email{david.kedziora@uts.edu.au}

\author{Katarzyna Musial}
\affiliation{%
  \institution{Complex Adaptive Systems Lab, Data Science Institute, University of Technology Sydney}
  \city{Sydney}
  \state{New South Wales}
  \postcode{2007}
  \country{Australia}
}
\email{katarzyna.musial-gabrys@uts.edu.au}

\author{Bogdan Gabrys}
\affiliation{%
  \institution{Complex Adaptive Systems Lab, Data Science Institute, University of Technology Sydney}
  \city{Sydney}
  \state{New South Wales}
  \postcode{2007}
  \country{Australia}
}
\email{bogdan.gabrys@uts.edu.au}

%
%

%
%

\keywords{Automated machine learning (AutoML); Bayesian optimisation; Sequential model-based optimisation (SMBO); Combined algorithm selection and hyperparameter optimisation (CASH); Multi-component predictive systems; Predictive services composition; Neural architecture search (NAS); Automated feature engineering; Meta-learning; Concept drift; Dynamic environments; Multi-objective optimisation; Resource constraints; Autonomous learning systems}

\renewcommand{\shortauthors}{Kedziora et al.}

\begin{abstract} Over the last decade, the long-running endeavour to automate high-level processes in machine learning (ML) has risen to mainstream prominence, stimulated by advances in optimisation techniques and their impact on selecting ML models/algorithms. Central to this drive is the appeal of engineering a computational system that both discovers and deploys high-performance solutions to arbitrary ML problems with minimal human interaction. Beyond this, an even loftier goal is the pursuit of autonomy, which describes the capability of the system to independently adjust an ML solution over a lifetime of changing contexts. However, these ambitions are unlikely to be achieved in a robust manner without the broader synthesis of various mechanisms and theoretical frameworks, which, at the present time, remain scattered across numerous research threads. Accordingly, this review seeks to motivate a more expansive perspective on what constitutes an automated/autonomous ML system, alongside consideration of how best to consolidate those elements. In doing so, we survey developments in the following research areas: hyperparameter optimisation, multi-component models, neural architecture search, automated feature engineering, meta-learning, multi-level ensembling, dynamic adaptation, multi-objective evaluation, resource constraints, flexible user involvement, and the principles of generalisation. We also develop a conceptual framework throughout the review, augmented by each topic, to illustrate one possible way of fusing high-level mechanisms into an autonomous ML system. Ultimately, we conclude that the notion of architectural integration deserves more discussion, without which the field of automated ML risks stifling both its technical advantages and general uptake.
\end{abstract}
\maketitle
\section{Introduction}
\label{sec:Intro}

The field of data science is primarily concerned with the process of extracting information from data, often by way of fitting a mathematical model. Data science, as an umbrella term for techniques drawn from various disciplines, is agnostic as to who or what is driving that extraction. Indeed, while much effort has been dedicated to codifying effective workflows for data scientists~\citep{fapi96}, e.g.~the Cross-Industry Standard Process for Data Mining (CRISP-DM)~\citep{chcl00} and others~\citep{kumu06, stbu21}, there is no inherent restriction that forces any phase of the process to be manually applied.

\begin{figure}[!htb]
  \centering
  \includegraphics[width=\linewidth]{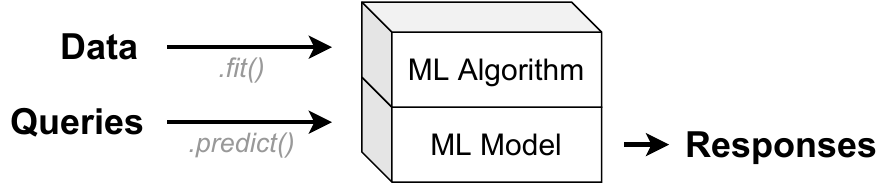}
  \caption{A simple representation of a machine learning (ML) model, which takes in user queries and outputs responses. The ML model is tuned by an ML algorithm, which processes training data.}
  \label{Fig:Solution}
\end{figure}

Certainly, in the modern era, one element of data mining and analysis is almost ubiquitously automated: model training. At one point in time, this was considered a novel advance, with computers only just becoming capable of running model-updating algorithms, as depicted in Fig.~\ref{Fig:Solution}, without human intervention. In fact, this form of automation was considered such a paradigm shift that it birthed the term `machine learning' (ML) in the 1950s~\citep{sa59}, provoked debate on its ``moral and technical consequences''~\citep{wi60, sa60}, and merged into the evolution of modern data analysis~\citep{tu62}. Advances since then, both fundamental and technological, have all but cemented computational dominance for model training.

Now, more than 60 years beyond the dawn of ML, associated techniques and technologies have diffused through society at large. While advances in graphics processing units (GPUs) and big data architectures are credited with popularising deep neural networks (DNNs), abundant black-box implementations of the backpropagation method have also played their part. In effect, the need for human expertise to develop complex inferential models has been lessened, and the last decade has consequently witnessed data science moving towards democratisation~\citep{boko19}. The 2012 journal article that is frequently credited with kicking off the DNN era sums it up well: ``What many in the vision research community failed to appreciate was that methods that require careful hand-engineering by a programmer who understands the domain do not scale as well as methods that replace the programmer with a powerful general-purpose learning procedure.''~\citep{krsu12}

\begin{figure}[!htb]
  \centering
  \includegraphics[width=\linewidth]{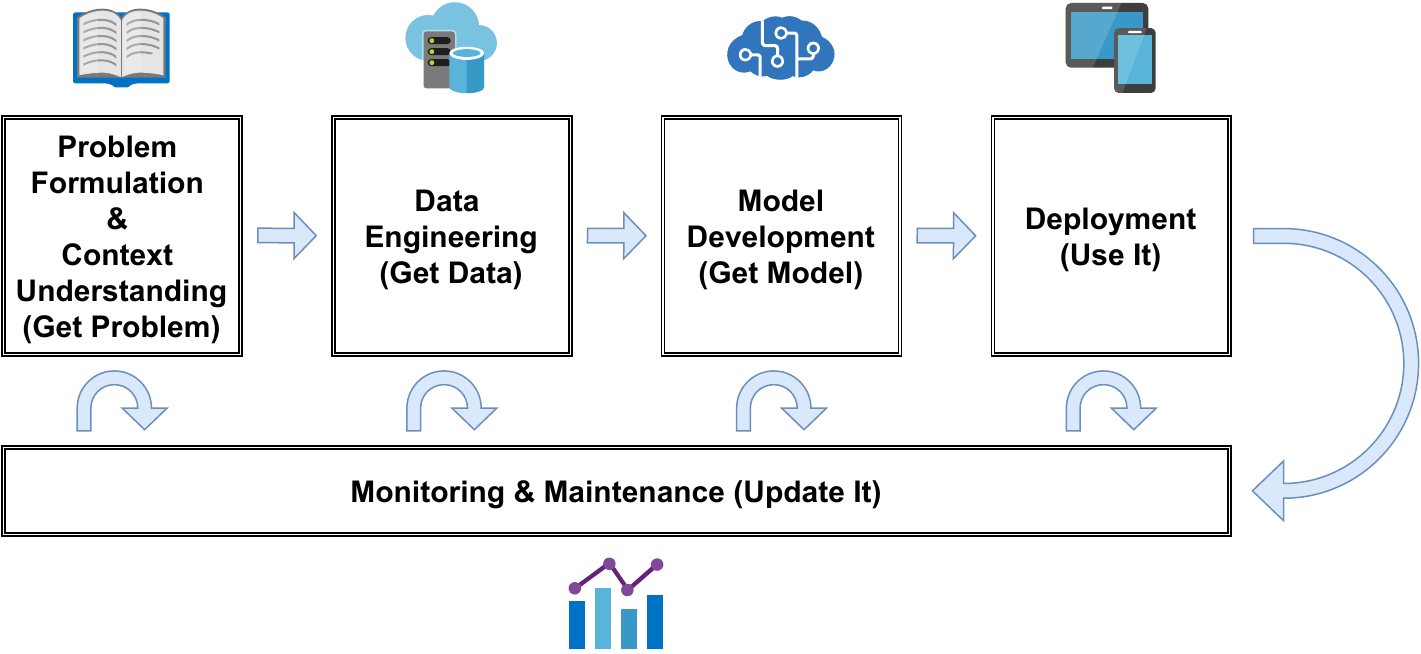}
  \caption{Schematic of the workflow involved in designing, constructing, deploying and maintaining an ML model. Crucially, monitoring and maintenance are required to enable continuous learning, a prerequisite for autonomous machine learning (AutonoML).}
  \label{Fig:Workflow}
\end{figure}

Despite all this, common practice has not yet seen widespread mechanisation along the rest of the data science chain~\citep{vega19}. Granted, given a specific choice of model, it is routine to train, validate and deploy the model on clean and informative datasets with `fit' and `predict' functionality, depicted in Fig.~\ref{Fig:Solution}, that is provided by numerous coding packages. However, even within the model-development phase, the typical data scientist still has to select a model, an evaluation metric and a training/validation strategy, all subject to human bias. Once the entire ML workflow is considered, shown at high level in Fig.~\ref{Fig:Workflow}, it becomes even more clear just how much an ML application relies on manual operations, which is not ideal. The data-engineering phase, for instance, is a notorious time sink for human involvement~\citep{ciku05, kumu06}.

The field of `automated machine learning' (AutoML)~\citep{huko19, yawa18, hezh21, zohu21, baal18, trwa19} has firmly established itself in recent years as a response to this; AutoML endeavours to continue mechanising the workflow of ML-based operations. It is motivated by the idea that reducing dependencies on human effort and expertise will, as a non-exhaustive list,
\begin{itemize}
    \item make ML and its benefits more accessible to the general public,
    \item improve the efficiency and speed of finding ML solutions,
    \item improve the quality and consistency of ML solutions,
    \item enforce a systematic application of sound and robust ML methodologies,
    \item enable quick deployment and reuse of ML methodologies,
    \item compartmentalise complexity to reduce the potential for human error, and
    \item divert human resources to more productive roles.
\end{itemize}
In fairness, the field as a whole must also grapple with the risks of automation, including
\begin{itemize}
    \item increased obscuration of ML technical debt~\citep{scho15},
    \item inappropriate or unethical usage as a result of ML illiteracy~\citep{boko19},
    \item interpretability issues in high-stakes contexts~\citep{ru19}, and
    \item adverse socio-economic impacts such as job displacement~\citep{wasi19}.
\end{itemize}
These are complex topics that are deserving of their own extensive discussions.

In this review, we focus primarily on the technical aspects of AutoML. Specifically, we motivate and discuss synthesising major threads of existing AutoML research into a general integrated framework. Unlike other published reviews, we also broaden the scope to capture adjacent research that has been, to date, barely or not at all associated with the AutoML label, particularly in recent scientific literature. In effect, the aim of this review is to help inspire the evolution of AutoML towards `autonomous machine learning' (AutonoML), where architectures are able to independently design, construct, deploy, and maintain ML models to solve specific problems, ideally limiting human involvement to task setup and the provision of expert knowledge. That does not mean that humans should be excluded from ML-based decision making, and there is certainly a necessary debate to be had about degrees of autonomy that are appropriate in various contexts. However, for the cases where society judges the benefits of automation outweigh the disadvantages, it is worth considering how AutonoML may be achieved.

Importantly, we do not profess the optimality of any particular AutonoML architecture. Given the breadth and malleability of the field, it is unlikely that any one framework proposal will perfectly subsume all existing AutoML systems, let alone future advances. On the other hand, the merits and challenges of integration are best discussed with reference to concrete schematics. Thus, the literature survey in this monograph is accompanied by the graduated development of a conceptual framework, exemplifying the potential interplay between various elements of AutonoML. As a necessity, Section~\ref{Sec:Basics} lays the initial groundwork for this example architecture by considering the fundamentals of ML, abstracting and encapsulating them as the lowest level of automatic operations; this is where the `fit' and `predict' functionality displayed in Fig.~\ref{Fig:Solution} resides.

The survey of AutoML begins in earnest within Section~\ref{Sec:CASH}, which discusses the role of optimisation in automating the selection of an ML model/algorithm and associated hyperparameters. Section~\ref{Sec:MCPS} then discards an assumption that the ML model need be monolithic, reviewing optimisation research for extended pipelines of data operators. This generalisation enables Section~\ref{Sec:NAS} to examine the optimisation of neural architectures specifically, while Section~\ref{Sec:Features} focusses on the pre-processing elements of an ML pipeline, exploring the automation of feature engineering. Subsequently, Section~\ref{Sec:Meta} investigates how model search can be upgraded by meta-knowledge, leveraging information from external ML experiments, while Section~\ref{Sec:Ensemble} identifies the importance of ensembles and discusses how their management can be mechanised.

Notably, in terms of the ML workflow shown by Fig.~\ref{Fig:Workflow}, the previously listed six sections focus heavily on the model-development phase -- with a touch of data engineering -- because this is the space in which the majority of AutoML research exists. Nonetheless, recent times have witnessed a significant broadening of scope along the entire workflow, and the next five sections document current research perspectives on the role of automation with respect to model deployment, model maintenance, and even problem formulation.

Thus, the paradigm of AutonoML is finally introduced in Section~\ref{Sec:Dynamic}, defined by the ability to adapt models within dynamic contexts. This stimulates an examination within Section~\ref{Sec:Eval} of how the quality of any one solution should even be determined. Section~\ref{Sec:Resource} then examines the challenge of automating operations in low-resource settings, while Section~\ref{Sec:User} reviews efforts to reintegrate expert knowledge and user control back into autonomous systems. The survey ends with an acknowledgement in Section~\ref{Sec:AGI} of the drive towards one-size-fits-all AutonoML, i.e.~the quest for general applicability. Finally, Section~\ref{Sec:Discussion} concludes with a discussion on the overarching technical challenges of integration, while remaining cognisant of the fact that understanding and fostering general engagement with resulting technologies are complex endeavours in their own right.
\section{Machine Learning Basics}
\label{Sec:Basics}

At its core, standard ML is driven by the following premise: given a structured set of possible `queries', $\mathcal{Q}$, and a corresponding space of possible `responses', $\mathcal{R}$, there exists a mapping from one to the other that is maximally desirable for some intended purpose, i.e.~an ML task. For instance, one may seek a function from bitmaps to truth values for image classification, a function from historic arrays to predicted scalar values for time-series forecasting, or a function from sensor data to control signals for robotic operations. Naturally, for problems of interest, the optimal mapping is typically unknown across the entire space of $\mathcal{Q}$. Data scientists thus construct an ML model, $M:\mathcal{Q} \to \mathcal{R}$, to approximate this function, generalising where necessary, and then employ an ML algorithm to configure the model. If the algorithm is any good at its job, it will systematically acquire experience, learning from interactions with data so that the accuracy of the approximation improves. This process is called model development.

\textbf{Modes and Examples of ML Operations.} Importantly, an ML model may exist in one of many operational modes across the duration of its existence. Terminology and definitions vary greatly, but these include:
\begin{itemize}
    \item Training -- A mode in which the parameters of the ML model, typically in its initial state, are tuned by its associated ML algorithm. Any data that supports this process is called training data.
    \item Validation -- A mode in which the accuracy of the ML model, given a selection of parameter values, is evaluated. The ML model converts incoming queries in $\mathcal{Q}$ to corresponding responses in $\mathcal{R}$, which are compared against expected values. The result is typically used to tune the `hyperparameters' of an ML model/algorithm and estimate predictive performance; see Sections~\ref{Sec:CASH} and \ref{Sec:Eval}, respectively. Validation data used in this process must be distinct from training data and must also be bundled with expected responses.
    \item Testing -- A mode similar to validation, but where evaluated accuracy serves to judge the performance of a finalised ML model/algorithm for the ML task at hand. Testing data used in this process must be distinct from training/validation data and must also be bundled with expected responses.
    \item Deployment -- A mode in which the ML model is put to practical use. It converts any incoming query in $\mathcal{Q}$ to a corresponding response in $\mathcal{R}$; these responses are used externally to fulfil the intended purpose of the ML model, e.g.~decision-making.
    \item Adaptation -- A mode in which the parameters of the ML model, previously trained and typically deployed, are re-tuned by its associated ML algorithm as a reaction to ML-model quality decreasing over time. This mode requires multiple tests and streamed data; see Section~\ref{Sec:Dynamic}.
\end{itemize}
As a note, these modes are not necessarily independent. For instance, a rapid-deployment scenario may begin using an ML model for practical purposes while that model is still being trained. Regardless, any genuine AutonoML system must facilitate all the above modes of operation.

\begin{figure}[!htb]
  \centering
  \includegraphics[width=\linewidth]{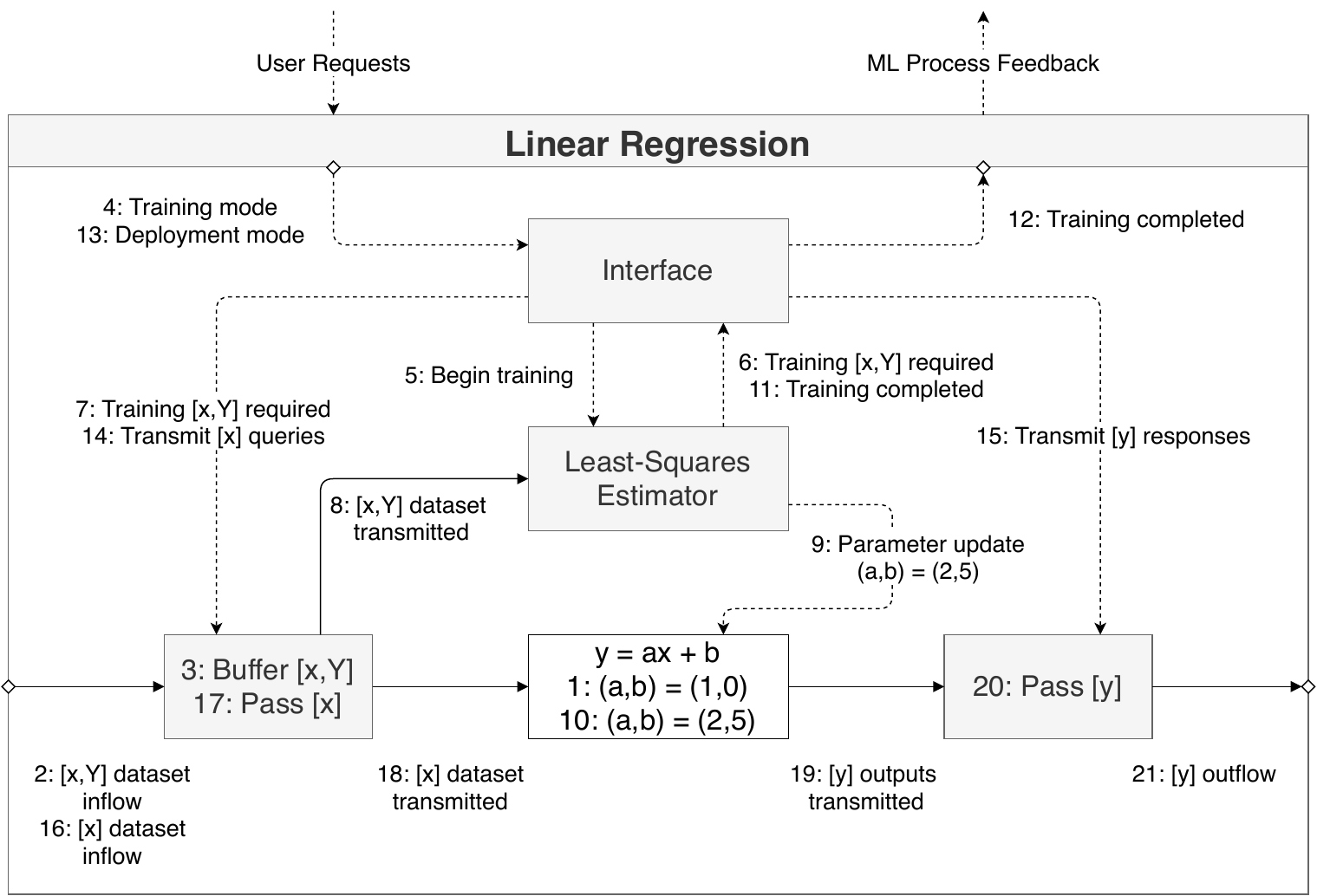}
  \caption{An example of an ML system processing a linear regression task, i.e.~training and deploying a linear model subject to a least-squares estimator (LSE). Steps are numbered chronologically and detailed in Section~\ref{Sec:Basics}. Solid arrows depict dataflow channels. Dashed arrows depict control and feedback signals.}
  \label{Fig:ExampleLSE}
\end{figure}

To exemplify some of these modes, Figure~\ref{Fig:ExampleLSE} provides an example of how model training and ensuing model deployment can be managed by an automated system that acts in response to user requests. In this example, a data scientist is tackling the standard task of linear regression, having selected the simple linear function $y=ax+b$ and a least-squares estimator (LSE) as their ML model and ML algorithm, respectively. The training of the linear predictor proceeds chronologically as follows:
\begin{enumerate}[leftmargin=1.5cm]
    \item The linear model is initialised with default parameters, e.g.~$(a,b) = (1,0)$.
    \itemrange{1} The user imports a training dataset into the ML system, which is collected in an inflow-data buffer, awaiting further orders. Each data instance contains a model input, $x$, and an expected output, $Y$. As befits a regression task, $Y$ is a continuous variable.
    \item The user orders the ML system via interface to begin training.
    \itemrange{3} The interface directs the LSE to begin training. The LSE signals back its training requirements. The interface, processing this, signals the inflow buffer to transmit the training dataset to the LSE, which it does so. This is equivalent to calling a `fit' method in a standard ML coding package.
    \itemrange{1} The LSE, according to its internal biases, calculates parameter values corresponding to the best linear fit for all provided $(x,Y)$ coordinates. It signals the model to update its parameters to new values, e.g.~$(a,b) = (2,5)$.
    \itemrange{1} The LSE emits a training completion signal that is passed to the user.
\end{enumerate}
The subsequent deployment and use of the linear predictor proceeds chronologically as follows:
\begin{enumerate}[leftmargin=1.5cm]
    \addtocounter{enumi}{12}
    \item The user now orders the ML system via interface to deploy the trained model.
    \itemrange{1} The interface directs the inflow-data buffer to transmit incoming queries of the form $x$ to the linear model. The resulting outputs of the form $y$ are allowed to propagate onwards, e.g.~to the user.
    \itemrange{2} The user imports a query dataset of $x$ values into the ML system, which is directly passed to the linear model. This is equivalent to calling a `predict' method in a standard ML coding package.
    \itemrange{2} The model calculates and emits associated response values, $y$. These are propagated onwards, e.g.~to the user.
\end{enumerate}

\begin{figure}[!htb]
  \centering
  \includegraphics[width=\linewidth]{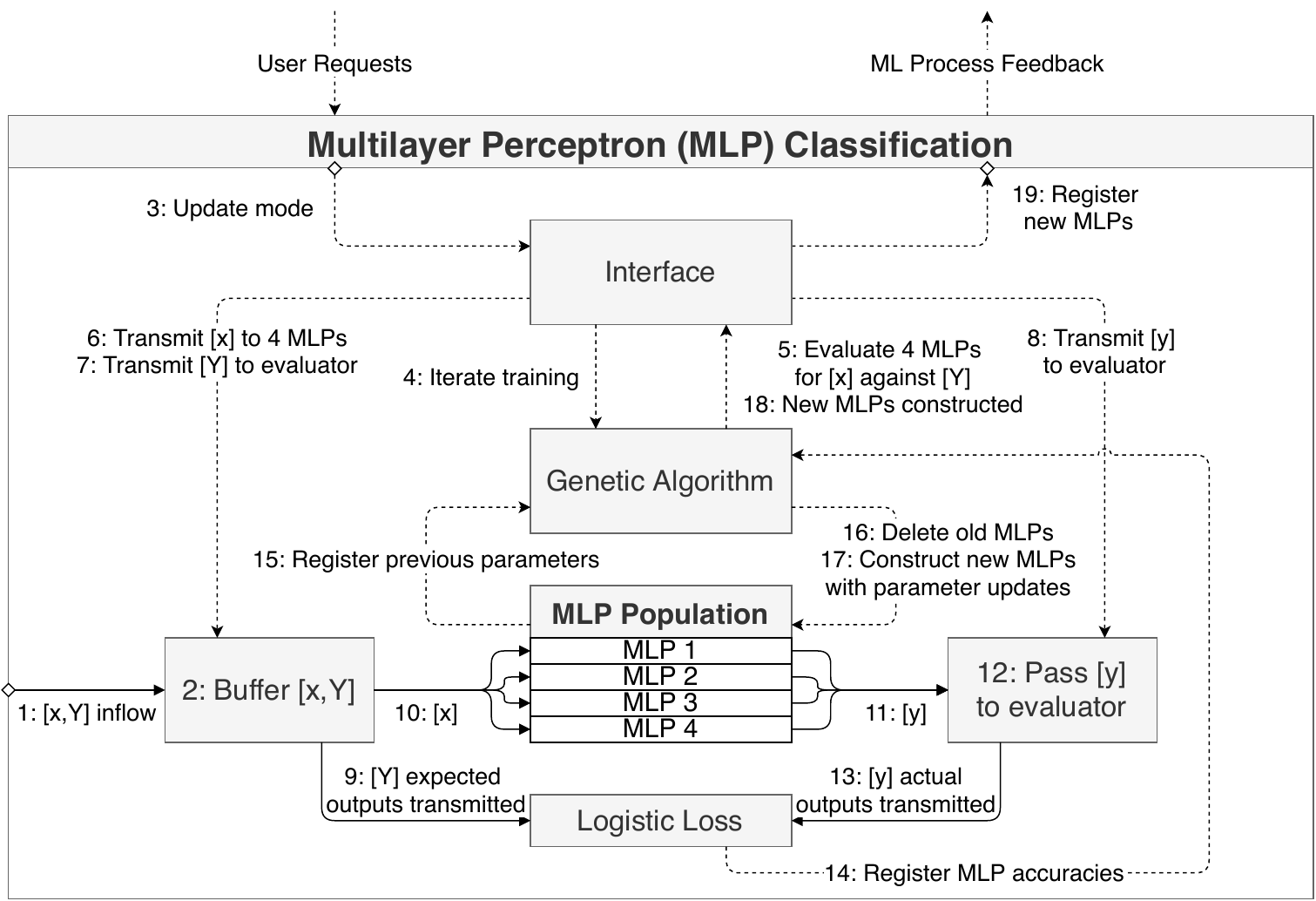}
  \caption{An example of an ML system processing a multi-layer perceptron (MLP) classification task, i.e.~updating a population of MLPs subject to a genetic algorithm (GA). A logistic-loss function is used to evaluate MLP accuracy. Steps are numbered chronologically and detailed in Section~\ref{Sec:Basics}. Solid arrows depict dataflow channels. Dashed arrows depict control and feedback signals.}
  \label{Fig:ExampleGA}
\end{figure}

Figure~\ref{Fig:ExampleGA} depicts a more complicated example of model development for an ML classification task, still managed by an automated system driven by user requests. Here, a data scientist has selected a multi-layer perceptron (MLP) as their ML model, training it via genetic algorithm (GA). Unlike the LSE of the previous example, the GA operates in an iterative manner and requires its parameter selections to be evaluated as part of the training process; the user has thus selected a classification-appropriate logistic-loss evaluator for this purpose. Due to the complexity of portraying a dynamic process within a static representation of ML architecture, Fig.~\ref{Fig:ExampleGA} only shows one iteration of training. It assumes that the user has already acquired a set of four MLP instances, each with their own unique values for parameters, i.e.~neural connection weights and biases. The update of the MLP `population' proceeds chronologically as follows:
\begin{enumerate}[leftmargin=1.5cm]
    \itemrange{1} The user imports a training dataset into the ML system, which is collected in an inflow-data buffer, awaiting further orders. Each data instance contains a model input, $x$, and an expected output, $Y$. As befits a classification task, $Y$ is a categorical variable.
    \item The user orders the ML system via interface to begin updating the model.
    \itemrange{1} The interface directs the GA to apply an iteration of training. The GA signals back its training requirements.
    \itemrange{3} The interface directs the inflow buffer to transmit available training inputs, $x$, to the MLP population. The interface also signals expected and actual MLP outputs, $Y$ and $y$, respectively, to be sent to the logistic-loss evaluator; $Y$ is transmitted immediately.
    \itemrange{3} For each MLP within the current population, the model transforms input $x$ into categorical variable $y$, which is then compared against expected output $Y$. A logistic-loss score is determined per MLP.
    \itemrange{3} The GA, registering the parameters and associated accuracy scores of each MLP model, selects parameters for a new population of MLPs. This selection is driven by elitism, crossover, mutation, and other principles relevant to the GA.
    \itemrange{1} The algorithm emits a feedback signal to the interface, detailing the completed process. The interface propagates this signal back to the user or any other external `listeners'.
\end{enumerate}
It should be noted that standard GA training continues to reiterate steps 4--19 for the same training inputs and expected outputs until some stopping criterion is met, such as a time limit. Subsequently, only the best performing model from the population is typically selected for deployment at the end of training, although an implementation may choose to keep the other models in memory for various reasons. 

These are but two examples among a great diversity of ML tasks, yet they already reveal commonalities to be incorporated into any inclusive framework. Specifically, an ML model is essentially a data transformation, whether as complex as an MLP or as simple as a linear function. Correspondingly, an ML algorithm is a parameter controller for this data transformation; it has full knowledge of the model and, during a model development phase, can assign values to all of its parameters.

\begin{figure}[!htb]
  \centering
  \includegraphics[width=\linewidth]{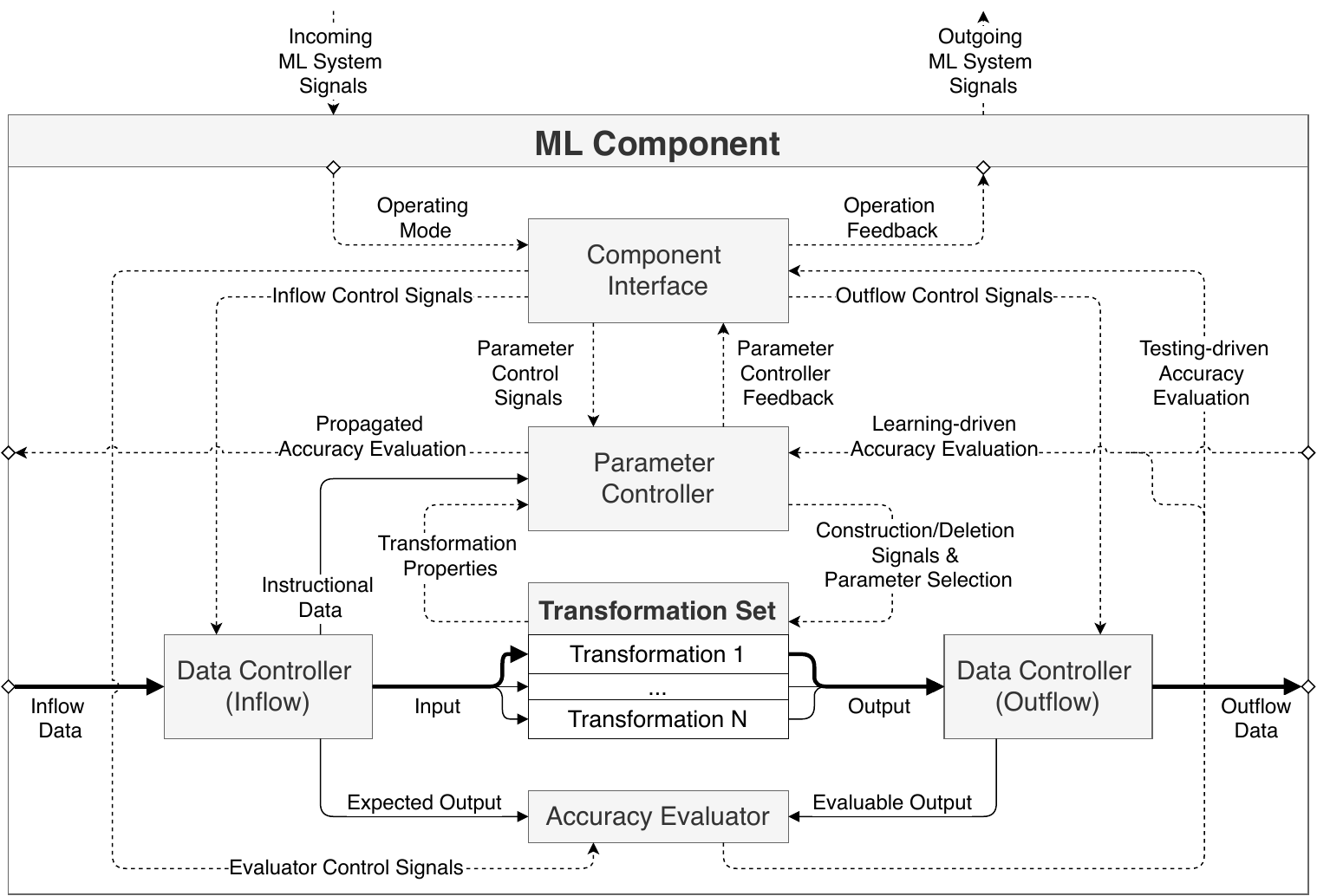}
  \caption{A general schematic of an ML component, wrapped around a data transformation and its parameter controller. A component interface communicates with external systems and the parameter controller, relaying directions to a pair of data controllers. The data controllers channel inflow and outflow data according to model development/deployment needs, while an accuracy evaluator assesses transformation quality based on supervised-learning data. Accuracy evaluations are directed to the parameter controller or the interface depending on whether they are sought as part of a learning or testing process. Solid arrows depict dataflow channels; thick arrows denote dataflow during deployment. Dashed arrows depict control and feedback signals.}
  \label{Fig:Component}
\end{figure}

\textit{\textbf{Framework - The ML Component.}} With these notions in mind, we now introduce the schematic shown in Fig.~\ref{Fig:Component}, which we propose to represent the foundation for an automated ML system: an `ML component'. Crucially, this concept should already be familiar to any AutoML developer, serving as an encapsulation that is amenable to substitution and, in advanced frameworks, concatenation; see Section~\ref{Sec:MCPS}. For simplicity, we will currently assume that a predictive system revolves around a solitary ML component. Additionally, while the schematic may appear complex, much of it is but a wrapper dedicated to providing a flexible interface for its two central elements: the data transformation and its associated parameter controller. Certainly, as the aforementioned LSE and GA examples show, not all depicted signals and dataflow channels are necessary, let alone active, in all contexts. Similarly, it is generality that motivates the use of a set-based container around the data transformation when coupling to a parameter controller. This is justified by Fig.~\ref{Fig:ExampleGA} demonstrating that some ML algorithms work with multiple copies of an ML model. In essence, while we do not necessarily advocate Fig.~\ref{Fig:Component} as the perfect way to structure an ML component, any alternative framework for truly generic AutoML should facilitate the same depicted processes, communications, and supported scenarios.

We begin detailing Fig.~\ref{Fig:Component} by describing and justifying core interactions. Automated model development requires the parameter controller to be cognisant of all transformation properties, as well as possess the ability to tune parameters and, in some cases, be able to construct/delete transformation objects. Notably, these qualities are already implicit within many existing ML coding packages that treat a model object as an output of an algorithm object. However, the Fig.~\ref{Fig:Component} schematic suggests that a data transformation and parameter controller should inherently be on equal footing; there is no theoretical reason why an ML model cannot be initialised alongside an ML algorithm using default values, e.g.~the identity function $y=1x+0$ for linear regression. Even grown/pruned models with a variable number of parameters, sometimes called constructivist/selectivist models~\citep{gale05}, have initial states. Default values may not be predictively powerful, but at least this representation supports the `anytime' validity of an ML model that is required by rapid-deployment AutonoML; see Section~\ref{Sec:Dynamic}.

Now, standard ML assumes that a data scientist is responsible for plugging in a combination of ML model and ML algorithm that is well suited to their objective and data, before driving the ML component through the various high-level operating modes described earlier. Each phase of operation may involve a different way of directing data around the component. For instance, certain ML algorithms may need to operate on data during initial training, such as the LSE portrayed in Fig.~\ref{Fig:ExampleLSE}, while ML-model outputs may need to be held back from a user prior to a deployment phase, as exemplified by the GA-based training shown in Fig.~\ref{Fig:ExampleGA}. The schematic in Fig.~\ref{Fig:Component} therefore includes an inflow-data controller (IDC) and an outflow-data controller (ODC) to disallow or distribute dataflow as required. In turn, these requirements are dictated by a component interface, which interprets instructions from external systems, or the user, and then translates them into lower-level directives that are disseminated as control signals. The interface also returns any relevant operational feedback to the source of external requests.

A generic ML component must additionally support an accuracy evaluator, allowing the quality of a data transformation to be assessed by comparing its outputs against expectation whenever available. In the Fig.~\ref{Fig:Component} schematic, this module is wrapped up within the ML component and managed by the interface. As with the parameter controller and associated transformation, the contents of this evaluator vary by ML task, e.g.~the logistic-loss calculator in Fig.~\ref{Fig:ExampleGA}. Moreover, it is possible that an ML component need not always instantiate the accuracy evaluator; certain ML models/algorithms seek novel patterns in inflow data for which the traditional concept of accuracy makes no sense.

\textbf{High Level Modes and Low Level Representations.} Returning to the topic of operating modes, what a user wants at a high level may not always project neatly into the lower level of an ML component. Even variations between ML models/algorithms can result in divergent processes, as demonstrated by the differences in training between the aforementioned LSE and GA examples; the latter employs calls to an accuracy evaluator as part of its internal learning procedure. This is why the component interface needs to translate concepts such as training and deployment, with regard to specific classes of ML model/algorithm, into combinations of atomic low-level directives. The following is one such proposal:
\begin{itemize}
    \item `Update' -- Request the parameter controller to continually tune the transformation set until some stopping criteria is reached. Direct the IDC to distribute inflow data in any way that supports the parameter-controller algorithm.
    \item `Validate' -- Direct the IDC to pass input data to the transformation set, then direct the IDC and ODC to distribute expected output and actual output, respectively, to the accuracy evaluator.
    \item `Propagate' -- Direct the IDC to pass input data to the transformation set, then direct the ODC to propagate the output further downstream.
\end{itemize}
In any implementation, these calls will likely be paired with further arguments to, for example, specify which buffered batches of streamed inflow data should be used for each process.

Given this basis set of low-level operations, a high-level request for model deployment is the most trivial to translate, mapping simply to a `propagate' call. Steps 13--15 for the LSE example in Fig.~\ref{Fig:ExampleLSE} demonstrate the relevant signals, while steps 16--21 depict deployed-model dataflow. The only implementational challenge is ensuring that the IDC can filter inflow data appropriately and, in the case of transformation plurality, identify the right ML model to feed with input data. All of this can be handled by the interface relaying signals from the parameter controller over to the IDC, specifically detailing the properties of the transformation set.

Similarly, high-level desires to validate or test an ML model, usually differing by which subsets of available data are used in the process, map directly to the `validate' directive. The accuracy evaluator is responsible for this procedure, comparing transformed inputs with expected outputs. Both higher-level modes, i.e.~validation and testing, usually serve to inform the user about the quality of an ML model/algorithm, which is why Fig.~\ref{Fig:Component} depicts `testing-driven' evaluations being sent to the component interface for external dissemination. This contrasts with the case of `validate' calls being part of model training, wherein accuracy evaluations are `learning-driven' and sent directly to the parameter controller, so as to advise the ML algorithm in tuning ML-model parameters. It is important to stress here that, beyond a small set of direct formulaic procedures, such as the LSE, most ML algorithms do technically iterate through some form of accuracy evaluations. Thus, this `learning-driven' signal is more a depiction of whether the feedback is explicit and exposed, e.g.~error calculations during DNN training, or implicit and black-boxed, e.g.~optimisations done by standard quadratic-programming solvers. This is an important nuance, as ML algorithms that expose that feedback loop are more flexible to outside control, with potential to improve or degrade algorithmic performance. Certainly, it has been shown that rolling validatory mechanisms into the learning process can result in dramatically different training runs, even with the same ML model/algorithm/evaluator~\citep{ga04}. As for a mechanised perspective, different ways of driving the same underlying ML algorithm with learning-driven accuracy evaluation could be encapsulated by an extra hyperparameter; see Section~\ref{Sec:CASH}.

As an additional point of note, the high-level validation of an ML model/algorithm is typically involved, commonly requiring repeat training/validation runs on multiple resamplings of inflow data. Two examples of this process are k-fold cross-validation and out-of-bootstrap validation. Alternatively, multiple stochastic iterations can be avoided with methods that construct a representative sample~\citep{buga10, buga11, buga13}, i.e.~a subset that preserves the statistics of the superset, but these require additional dataset analysis. Again, we do not advocate for any particular implementation here; developers may decide whether responsibility for the management of a cross-validatory process, for example, falls to the component interface or to a higher level. However, the cleanest delegation of responsibility suggests that the interface should only direct operations per received data sample, with the management and aggregation of all resamplings taking place outside of the ML component, as elaborated in high-level schematics within Section~\ref{Sec:CASH}.

Ultimately, operation modes that involve model training/retraining, i.e.~the `update' directive, are where complexities truly arise. For a general ML system, the interface must be flexible enough not just to direct the parameter controller to tune the transformations but also to understand what it needs for that tuning procedure. This communication via parameter control signals and parameter controller feedback is demonstrated by steps 5--6 for the LSE example in Fig.~\ref{Fig:ExampleLSE}. In general, training requirements can vary dramatically, given that numerous parameter selection strategies exist, with differences depending on the properties of dataflow, selected model and intended task. However, without any inputs beyond the properties of the ML model itself, an ML algorithm can do nothing but blindly guess at transformation parameters. Hence, the simplest extension is to allow the IDC to pass `instructional data' to the parameter controller, as demonstrated by step 8 from the LSE example. Instructional data can be thought of as training data, but also crucially includes data used to evolve a model during adaptation, an important concept explored within Section~\ref{Sec:Dynamic}. Notably, not all ML algorithms need instructional data, as the GA example in Fig.~\ref{Fig:ExampleGA} demonstrates. However, to constitute a learning process, random parameter guesses must eventually be improved by some form of data-driven feedback, whether immediately calculated by an accuracy evaluator or supplied by a delayed evaluation downstream. This is why the Fig.~\ref{Fig:Component} schematic provides two sources for learning-driven accuracy evaluations.

\textbf{Diverse Forms of Learning.} Once mechanisms for instructional dataflow and accuracy evaluations are fully incorporated, the ML component in Fig.~\ref{Fig:Component} becomes capable of encompassing an extensive variety of ML scenarios. These include the following types of learning:
\begin{itemize}
    \item Unsupervised Learning -- An ML algorithm generates an ML model purely from instructional data, without being provided any external guidance in the form of an expected $\mathcal{Q}\to\mathcal{R}$ mapping. The ML model is thus completely subject to the internal biases of the ML algorithm. Although there are methods to quantify model quality with respect to the intent behind certain algorithms, there is no intrinsic accuracy to evaluate, i.e.~there is no `ground truth'.
    \item Supervised Learning -- ML-model inputs arrive packaged with expected outputs, i.e.~labels or target signals. An ML algorithm can learn from these directly, as instructional data, and/or indirectly, via accuracy evaluations that compare model outputs with expected outputs.
    \item Reinforcement Learning (RL) -- Expected $\mathcal{Q}\to\mathcal{R}$ mappings are not available, but the quality of different parameter selections are still comparable, with an indirect `reward' metric guiding improvement. Some RL strategies can derive this heuristic from instructional data, but, more commonly, model outputs must be assessed downstream with the evaluation then returned to the algorithm.
\end{itemize}
Hybrids of these scenarios are also represented, such as semi-supervised learning, for which training data arrives as a mixture of labelled and unlabelled instances. Semi-supervised learning typically leverages statistical information from the latter to improve the predictive knowledge afforded by the former. In all cases, learning procedures are mostly internalised by the parameter controller, if not entirely, no matter if they formulaically compute ML-model parameters, e.g.~the LSE, or do so in iterative fashion, e.g.~k-means clustering.

As a side note, it is worth highlighting that the representational generality provided by the Fig.~\ref{Fig:Component} schematic does mean that instructional data and transformation-input data can often occupy different vector spaces. This is obvious in most forms of supervised learning, where ML algorithms train on $\mathcal{Q}\times\mathcal{R}$ while ML models operate in $\mathcal{Q}$-space. Unsupervised learning algorithms are much more varied. While k-means clustering may both train and deploy on $\mathcal{Q}$, the Apriori algorithm used in association learning is a radically contrasting example; the ML algorithm operates on sets of objects, while the ML model operates on queries depicting implication rules. The two do not inherently embed within the same vector space. Consequently, within the Fig.~\ref{Fig:Component} representation, the interface-supported IDC must be capable of identifying and redirecting heterogeneously structured data into the appropriate channels.

At this point, there are a few miscellaneous comments to add about the ML component depicted in Fig.~\ref{Fig:Component}. First of all, just as accuracy evaluations can be received from downstream, they can also be augmented and propagated upstream, hence the inclusion of the `propagated accuracy evaluation' arrow in the schematic. This enables representation of automatic differentiation techniques, such as backpropagation, but these cannot be discussed until considering multiple components, done in Section~\ref{Sec:MCPS}. Secondly, the ability to keep multiple ML-model instances within memory via the transformation set implies that, with appropriate upgrades for the ODC, homogeneous ensembling could be implemented internally within an ML component. Discussion on ensembles in general is delayed until Section~\ref{Sec:Ensemble}. Finally, closer inspection of the GA example in Fig.~\ref{Fig:ExampleGA} suggests that the deletion/construction of MLP objects in steps 16--17 is superfluous when simply changing the parameter values of the existing MLP population would be sufficient. However, in so doing, the example demonstrates via steps 18--19 how an ML component could communicate with an external resource manager, distributing new model instances across available hardware. Resource management in the ML context is discussed in Section~\ref{Sec:Resource}.

\textbf{Beyond the Fundamental ML Component.} The schematic in Fig.~\ref{Fig:Component} broadly captures standard ML operations, including learning that is unsupervised, supervised, and reinforcement-based. Within this context, automation of the parameter controller, i.e.~model training, is essentially what spawned the field of ML. However, it is important to note that the same ML model can be tuned by many different ML algorithms. For instance, a linear function may be tuned via formulaic parameter estimation, e.g.~with the LSE or Theil–Sen regression, or via an iterative method, such as is typically done for training a perceptron. Additionally, with so many forms of data transformation to choose from, the question arises: given a particular ML task, which ML model/algorithm should a data scientist use? The attempt to answer this question in a systematic manner is the foundation of AutoML.

\section{Algorithm Selection and Hyperparameter Optimisation}
\label{Sec:CASH}

The modern use of the abbreviation `AutoML' to represent the field of automated machine learning arguably stems from a 2014 International Conference on Machine Learning (ICML) workshop. Specifically, while automating high-level ML operations has broad scope and thus an extensive history, as will be discussed in other sections, it was around this time that, fuelled by a series of advances in optimisation strategies, it became not just feasible but also effective to leave the selection of an ML model/algorithm up to a computer. Accordingly, with the optimisation community seeding the modern narrative of automation in ML, this is a logical starting point when discussing the evolution of standard ML into AutoML.

First, though, a caveat: while we have made an explicit distinction between an ML model and an ML algorithm, or a data transformation and parameter controller in Fig.~\ref{Fig:Component}, scientific literature in the field of ML can be loose with terminology~\citep{sm08}. Given how closely models are coupled with their training algorithms, e.g.~`support vector machine' (SVM) often referring to both, sections of the ML community use `algorithm selection' and `model selection' interchangeably. In practice, model/algorithm selection refers to a data scientist or high-level system swapping out an entire ML component for another, seeking a pairing of ML model and ML algorithm that is optimal for the task at hand.

\textbf{A Problem of Choice.} Semantics acknowledged, the algorithm-selection problem, surveyed more generally elsewhere~\citep{keho19}, is far older than the current wave of AutoML research; its conception is often attributed to the 1970s~\citep{ri76}. It was originally posed in fairly abstract terms, seeking a general framework that, given a set of problems and a set of problem-solving algorithms, could identify an optimal algorithm for each problem, according to some specified measures of performance. However, at the time and within the ML context, there was no feasible procedure to both systematically and efficiently search such an arbitrarily large space. After all, applying an ML algorithm and training a single ML model could already take significant time, especially on the hardware of that era. Nonetheless, with theory and technology progressing over the decades, research attention turned to a limited form of model selection: hyperparameter optimisation (HPO)~\citep{clde15, yash20}.

Whereas a parameter of an ML model is a degree-of-freedom tuned by an ML algorithm during model development, a hyperparameter is a characteristic of either the model or algorithm that is set to a fixed value at initialisation. For example, with respect to a typical DNN trained via backpropagation and stochastic gradient descent (SGD), neural connection weights are parameters, while network size and the SGD learning rate are both hyperparameters. As a side note, distinctions have occasionally been drawn between model-based and algorithm-based hyperparameters, e.g.~the latter being called `training tunables'~\citep{cuga18}. Whatever the case, it is quickly evident that hyperparameters already define an extensive space of variation within a fixed class of models/algorithms. Systematically comparing two distinct types of model, like an SVM against an MLP, adds yet another layer of challenge. It is for this reason that, right up until the dawn of the modern AutoML era, researchers have sometimes used `model selection' in a more limited sense, i.e.~strictly in reference to tuning hyperparameters for fixed model types~\citep{chtr13, or14}.

Critically, hyperparameters can have a major effect on the suitability of a data transformation for any particular dataset. This would seem obvious, and ML novices are often taught early on to explore model performance under hyperparametric variation, yet this lesson remains commonly enough ignored to justify warnings against the use of default hyperparameter values~\citep{baca17}. Emphasising the surprising novelty of this point, recent publications in the field of software defect prediction have likewise announced that default classifier settings can perform poorly~\citep{tamc16, quch18, tamc19}. In fairness, identifying that hyperparameters can define a poor model/algorithm is far easier than finding good ones; this is the goal of HPO.

Typically, HPO is expressed as a minimisation problem for a loss function, $L$, as calculated for an ML model/algorithm, $A$, that is applied to sets of training and validation data, $D_\mathrm{train}$ and $D_\mathrm{valid}$, respectively. Model/algorithm $A$ is itself dependent on a set of hyperparameter values drawn from a structured space, i.e.~$\lambda \in \Lambda$. Some form of cross-validation is also often involved, drawing $D_\mathrm{train}$ and $D_\mathrm{valid}$ $k$ times from a larger dataset to statistically average loss. With this notation, HPO is written as
\begin{equation}
\label{Eq:HPO}
    \lambda^* \in \argmin\limits_{\lambda \in \Lambda}\frac{1}{k}
    \sum_{i=1}^{k}L(A_\lambda, D_\mathrm{train}^{(i)}, D_\mathrm{valid}^{(i)}).
\end{equation}
The most standard approach to this problem, especially when manually applied, is a grid search through hyperparameter space. Another relatively simple alternative is random search, which tends to find more accurate models much more efficiently~\citep{bebe12}. Some packages, e.g.~H2O AutoML~\citep{h217}, are content to use grid/random search for model selection on the balance of performance versus complexity.

\textbf{The Bayesian Breakthrough.} The development of HPO algorithms that incorporate exploitation, not just exploration, has historically been quite gradual. Many attempts would restrict themselves to hyperparameters with continuity and differentiability conditions; these include gradient-based methods~\citep{be00} and bi-level optimisations over combined parameter/hyperparameter spaces~\citep{beku08}. However, among HPO efforts, Bayesian-based approaches began to appear particularly promising, often assessed against frequentist methods due to the overlap between ML and the field of statistics~\citep{kala04, gusa10}.

Bayesian optimisation (BO)~\citep{shsw16} is usually considered to have been birthed in the 1970s~\citep{mo75}, although foundational concepts were explored much earlier. A core idea behind the approach, introduced at least a decade earlier~\citep{ku64}, is to use stochastic processes to approximate unknown functions. These approximators are sometimes called surrogate models or response curves~\citep{josc98}. Plenty of literature has been dedicated to analysing surrogates and their utility~\citep{huho09, huba10, du15}, although Gaussian processes remain a traditional standard. Regardless, once a surrogate has been chosen, BO uses this approximation to drive iterative evaluations of the unknown target function, with each exploration coincidentally tightening the fit of the surrogate. Technically, each evaluation is advised by an `acquisition function' based on the surrogate, which estimates the benefit of evaluating any point along the unknown function. A commonly used metric for this is `expected improvement', although alternatives have been proposed, such as `entropy search'~\citep{hesc12}, which focusses on maximising the information gained about an optimum rather than simply moving closer towards it. More in-depth details about BO can be found in tutorials elsewhere~\citep{brco10}.

In the meantime, in parallel to BO gradually developing as a feasible approach for standard HPO, the ML community began to return to the question of full model/algorithm selection, with the 2006 Neural Information Processing Systems (NIPS) conference organising an early challenge dedicated to this topic~\citep{gusa11}. As a result, possibly the first method for full model search was published not long after~\citep{esmo09}. It was based on particle swarm optimisation (PSO) and even tackled the more ambitious problem of multi-component ML models, discussed later in Section~\ref{Sec:MCPS}.

However, from a model-search perspective, the genesis of the modern AutoML era is often synonymous with crucial advances in BO. Specifically, while some HPO strategies could handle issues caused by mixing continuous and categorical hyperparameters, conditional values were more challenging. For instance, if one hyperparameter denotes the kernel that an SVM employs, a value of `polynomial' requires a polynomial-degree hyperparameter to also be chosen. Similarly, a `radial basis function' value opens up the use of a Gaussian-width hyperparameter. In effect, the hyperparameter space for many models/algorithms resembles a complex hierarchical tree. Nonetheless, once these conditional hyperparameters could properly be treated~\citep{huho09a}, a new method for generic HPO was published, namely Sequential Model-based Algorithm Conﬁguration (SMAC)~\citep{huho11}, which uses random forests as surrogates for unknown loss functions that operate on hierarchical domains. Incidentally, the SMAC publication also popularised the phrase `sequential model-based optimisation' (SMBO) to describe the Bayesian-based procedure. Two search methods produced soon after, namely the Tree-structured Parzen Estimator (TPE) approach~\citep{beba11} and Spearmint~\citep{snla12}, are both considered to be popular SMBOs, although the latter was notably not designed to handle conditional parameters.

Full model search was quickly identified as a natural extension of HPO. In 2013, the combined algorithm selection and hyperparameter optimisation problem (CASH) was formalised~\citep{thhu13}, written as
\begin{equation}
\label{Eq:CASH}
    A^*_{\lambda^*} \in \argmin_{A^{(j)} \in \mathcal{A},\; \lambda \in \Lambda^{(j)}}\frac{1}{k}
    \sum_{i=1}^{k}L(A_\lambda^{(j)}, D_\mathrm{train}^{(i)}, D_\mathrm{valid}^{(i)}),
\end{equation}
where the optimisation is applied across all ML models/algorithms of interest, $A^{(j)} \in \mathcal{A}$, and their associated hyperparameter spaces, $\Lambda^{(j)}$. However, the important insight bundled into CASH was that combinations of ML model and ML algorithm could be described as just another categorical hyperparameter at the root of a hierarchical tree~\citep{thhu13, bego15}. This meant that generic SMBO algorithms like SMAC and TPE could be applied directly to varying model/algorithm types. Accordingly, 2013 marks the release of the first AutoML package based on SMBO, namely Auto-WEKA~\citep{thhu13}.

\textit{\textbf{Framework - Solving CASH.}} Crucially, Auto-WEKA popularised a form of modularity that has essentially been ubiquitous across AutoML packages, wherein the mechanism for solving HPO/CASH acts as a high-level wrapper around low-level ML libraries, these often being produced by third-party developers. Indeed, as the name implies, Auto-WEKA applies its selection method to a pool of classification algorithms implemented by the WEKA package. In effect, or at least if implemented well, the optimisation routine can be abstracted away from ML-model particulars.

\begin{figure}[!htb]
  \centering
  \includegraphics[width=\linewidth]{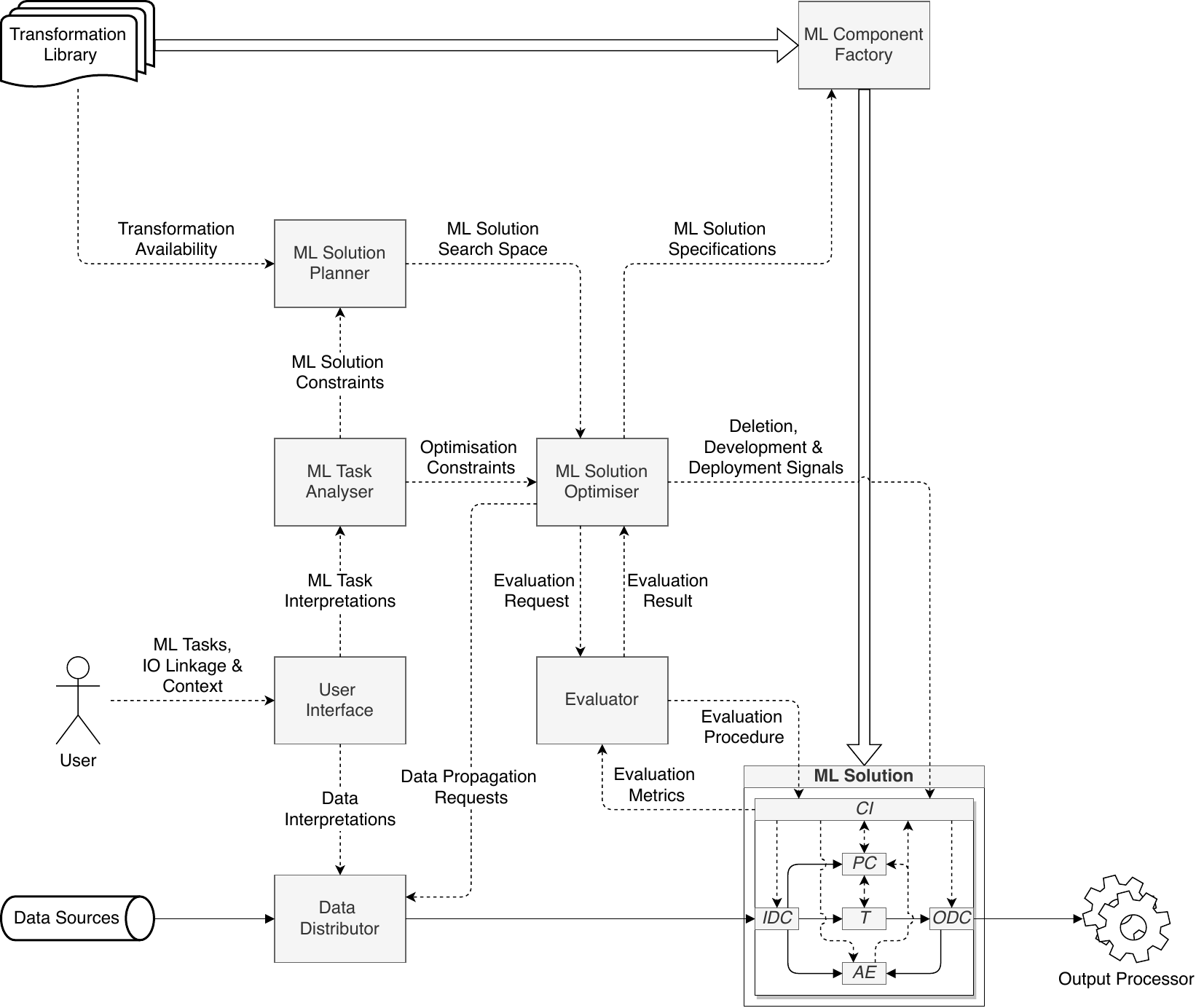}
  \caption{A high-level schematic of a minimalist AutoML system, demonstrating how a CASH-solver can be incorporated in place of manual model selection within an ML platform. An interface allows a user to describe a desired ML task, along with inputs and outputs to the system. Once the task is interpreted, with relevant constraints passed onwards, a solution planner advises a CASH-solving optimiser in terms of transformation availability and associated ML-solution search space. The optimiser proceeds to iteratively select candidates for an ML solution, instructing a factory to instantiate relevant ML components. An evaluator serves to assess these candidates, while a data distributor feeds training/validation data to the ML components during assessment. The distributor also propagates queries once optimisation concludes and an ML solution is deployed. Every ML component contains a data transformation (T), a parameter controller (PC), a component interface (CI), controllers for inflow/outflow data (IDC/ODC), and an accuracy evaluator (AE). Dashed arrows depict control and feedback signals. Solid arrows depict dataflow channels. Block arrows depict the transfer of ML components.}
  \label{Fig:ModelSelection}
\end{figure}

For illustrative purposes, a variant of this design pattern is shown in Fig.~\ref{Fig:ModelSelection}, applying model selection via optimisation to a library of transformations and their associated training algorithms. In this example, a user interface (UI) for the AutoML system fields user requests, such as the desire to run a classification task. Abstractly put, the user must also link the system to input/output (IO) and ensure that this IO is sufficiently described, e.g.~where the data sources are, what their formats are, where model responses should be sent, and so on. This provided context allows developmental/query data to be propagated as requested by a data-distributor module.

Central to Fig.~\ref{Fig:ModelSelection}, and any modern AutoML system, is the ML-solution optimiser, e.g.~an implementation of SMAC. Systems vary in how much flexibility a CASH-solver is afforded, so the schematic includes an ML-task analyser that passes optimisation constraints onwards, e.g.~time allowed per iteration. Often, this analysis can also provide some hard constraints on the ML solution to be expected, e.g.~excluding regression-based models for a classification task. These constraints must be examined in the light of which data transformations are available to the AutoML system, so an ML-solution planning module is tasked with actually detailing the search space to the optimiser.

From here, the encapsulation of ML models/algorithms with their low-level control/management mechanisms, detailed in Section~\ref{Sec:Basics}, makes HPO convenient to depict. During optimisation, the CASH-solver iteratively directs a factory to instantiate ML components according to multi-dimensional hyperparameter guesses, with types of ML model/algorithm included in the selection process. Each ML component becomes a candidate ML solution, with the optimiser deciding when to delete it. During each iteration of optimisation, a high-level evaluator is tasked with assessing the candidate; it directs the ML component via interface to undertake an evaluation procedure, using data that is simultaneously sent to its IDC. As Section~\ref{Sec:Basics} implied, this is where it seems most logical to automate cross-validatory mechanisms. Finally, when the optimiser hits its stopping criteria, it ensures that the optimal ML component it found is constructed and then signals it to train on a complete set of developmental data, requested from the data distributor. When this is finished, it signals the ML solution to deploy, requesting the data distributor to pass on any incoming queries, and the ODC of the ML component subsequently outputs all responses.

As before, we do not claim that this is necessarily the best way to partition AutoML processes into modules and network them up, particularly from an implementational perspective. Deeper commentary on the challenges of integration is provided in Section~\ref{Sec:Discussion}. Moreover, bespoke frameworks arguably do not need this level of complexity, e.g.~a restrictive transformation library may not require a planner to constrain search space. However, if the field of AutoML is dedicated towards automating an increasing spectrum of user-desired ML operations, all concepts depicted in Fig.~\ref{Fig:ModelSelection} must be integrated in some fashion.

\textbf{Modern Optimisation Methods.} Returning to SMBO methods in general, research and development continued in the years following the release of SMAC~\citep{lu16a}. One of the earliest aims of the sub-field was to decide whether some SMBO strategies could find optimal models faster and more accurately than others. As Eqs (\ref{Eq:HPO}) and (\ref{Eq:CASH}) show, this is not a trivial question; a loss function is critically dependent on the data used to train and validate a model. Thus, Python-based benchmarking library HPOlib was proposed to standardise this~\citep{egfe13}. It was used in conjunction with several search strategies built around SMBO package Hyperopt~\citep{beya13}, which only implements TPE by default, to make preliminary assessments~\citep{beko14}, although these served more as a proof of concept than strong recommendations for any one SMBO method. Subsequent attempts by the ML community have attempted to mitigate the resource costs of benchmarking~\citep{eghu15}, discuss ways of ranking the SMBO methods~\citep{demc16}, and assess strategies within specific contexts such as multi object tracking~\citep{mama18}. Of course, Hyperopt is only an early entry among several libraries that provide SMBO implementations. For example, an R-based machine-learning toolbox for model-based optimisation (mlrMBO) extends into multi-objective optimisation~\citep{biri17}, while Python-based Sherpa prioritises parallel computation~\citep{heco18}.

Other non-Bayesian strategies have also shown promise for both HPO and CASH. Evolutionary algorithms such as PSO and GA have been heavily investigated~\citep{esmo09, cane14, gasa18}, shown to perform better than gradient-based methods in certain nonlinear contexts~\citep{diko18}, although their advantages are most discernible for multi-component ML models, discussed in Section~\ref{Sec:MCPS}. A lot of recent focus has also gone into reinforcement-based strategies. Although RL becomes particularly pertinent for neural network design~\citep{zole17, bagu17}, discussed in Section~\ref{Sec:NAS}, the long-studied multi-armed bandit (MAB) problem~\citep{th33, sc10} has been linked in general fashion to HPO. This has led to a number of MAB-based solvers being proposed, e.g.~the multi-armed simultanous selection of algorithm and its hyperparameters (MASSAH) method~\citep{effi16} and the more recent Hyperband~\citep{lija18}. Accommodating the HPO setting does often require certain MAB assumptions to be tweaked, such as treating the performance `reward' of selecting an `arm', i.e.~model, as non-stochastic~\citep{jata16}. Other research has continued relaxing assumptions, such as exploring nonstationary contexts where the performance reward may change over time~\citep{mcbe20}; this is perhaps relevant to the selection of ML models/algorithms in the context of adaptation, discussed in Section~\ref{Sec:Dynamic}. Notably, attempts have also been made to fuse Bayesian theory with MAB strategies, such as using correlations between arms to advise subsequent arm-pulls~\citep{hosh14}, and one such culmination of these efforts is the BO-HB method~\citep{fakl17}, which aims to balance the exploitation of Bayesian optimisation with the exploration of Hyperband. Then there are even more exotic variants for CASH, such as combining RL with Monte Carlo tree search (MCTS)~\citep{drkr18}.

\textbf{How to Search Better.} Regardless of CASH-solver chosen, it is commonly accepted that searching for a task-optimal ML component, i.e.~model/algorithm, is extremely computationally expensive, likely to take hours if not days on modern hardware. Accordingly, it has been suggested that, even in the more limited case of HPO, where research once prioritised over-fitting avoidance, the big-data era has shifted focus to search efficiency~\citep{maes19}. Meta-learning, discussed in Section~\ref{Sec:Meta}, has been suggested as one possible way to boost the selection of ML models/algorithms, such as pruning hyperparameter space based on previous experience~\citep{wisc15}. However, in the absence of external information, other speed-up mechanisms have been sought.

Some researchers have suggested that the easiest way to reduce search space is to simply design useful ML models/algorithms with fewer hyperparameters~\citep{hulu15}. In the same vein, there exists an entire stream of work dedicated to hyperparameter-free optimisation~\citep{or14, orpa16}, where, for example, an algorithm analogised by `coin betting' can outright remove the need to tune learning rate when training DNNs~\citep{orto17}. Alternatively, if the number of hyperparameters is necessarily fixed, decomposing CASH once more into smaller independent subproblems, as is done via the alternating direction method of multipliers (ADMM), has its appeal~\citep{lira20}. Others have analysed the hyperparameters themselves, assessing their impact on ML models/algorithms either in general~\citep{prbo19} or in specific contexts, such as with random forests~\citep{prwr19}. The hope is that identifying `tunable' hyperparameters, i.e.~ones that model performance is particularly sensitive to, will allow other settings to be ignored, constraining search space. In fact, determining a good hyperparameter subspace, occasionally with the recommendation of priors, is the motivation behind several analysis-of-variance (ANOVA) studies~\citep{huho14, wulu18, rihu18, shri19}. However, these investigations often acknowledge the choice of training datasets as complicating factors, making it difficult to form generic statements on hyperparametric influence beyond those results taken to be obvious, e.g.~epoch number is very important to deep learning.

Complementary to constraining search space is cutting off unpromising forays via early termination. In simplest form, this is abandoning a search if the performance of an evaluated ML model/algorithm is not bettered after a subsequent number of selections~\citep{wulu18}. A more involved approach for CASH-solvers that parallelise evaluations is to initially test a large sample of hyperparameter configurations, maximising exploration, and then gradually decrease the size of subsequent samplings, thus honing in on the best-performing models~\citep{lija18, fakl17}. This is often bundled with a resource allocation strategy, e.g.~of training time, so that a fixed budget is spread less thinly across fewer candidate models when they are expected to be better performing.

Engineering search-boosts can also target lower-level aspects, such as the training/validation time it takes to evaluate a single candidate model. For instance, in the 1990s, Hoeffding races were proposed as one way to discard poorly performing models quickly during cross-validation. Specifically, confidence bounds on model accuracy would be updated per iteration of training/validation, with any model being immediately eliminated from contention if its maximum bound fell below the minimum value of any competitor~\citep{mamo94}. Feedback-driven ML algorithms, such as backpropagation-based SGD, are similarly open to interruptions prior to completion; learning curve extrapolations have been explored as quick estimates of final model performance~\citep{dosp15}, allowing for early termination.

Naturally, the logical extreme of speeding up training/validation is to constrict the data itself. Even a small subsample can indicate the expected performance of a fully-trained model~\citep{pe00}, especially if the sample is density-preserving and reflects the characteristics of the full dataset~\citep{buga10, buga11, buga13} or involves some bootstrapping technique~\citep{an17, andu17}. Hence, similar to a graduated allocation of training time, some HPO strategies use an increasing amount of training data for sequential model evaluations, honing in on better performing regions of hyperparameter space after cheap exploratory estimates~\citep{wafe15}. Sometimes, for CASH, this is coupled with other iterative updates, such as progressively decreasing the pool of ML model/algorithms to explore~\citep{lu16a, zelu17}.

Further efficiency upgrades may be possible depending on CASH-solver specifics. For instance, the acquisition function used by SMBO methods is a prime target for augmentation. Metrics such as `expected improvement per second' and similar variants have been shown to guide BO away from regions of hyperparameter space where evaluations are estimated to be time-expensive~\citep{snla11}. A more recent example involving a method called Fast Bayesian Optimisation of Machine Learning Hyperparameters on Large Datasets (FABOLAS) gives itself the freedom to select how much of a dataset a model should be trained on; the subsample size is factored into its entropy search metric, forcing BO to balance the desire for more instructional data against the desire for cheap model training~\citep{klfa17}. Notably, although the idea veers towards meta-learning principles discussed in Section~\ref{Sec:Meta}, entropy search has also previously been boosted by factoring correlations between related tasks~\citep{swsn13}.

On a final practical note, upon appreciating that the field of HPO is constantly evolving, developers intent on designing AutoML systems may ask: what CASH-solver should I plug into my codebase according to the state of knowledge as of the early 2020s? According to recently published comprehensive benchmarking, the state-of-the-art answer is BO-HB, assuming well-behaved datasets where random subsamples are representative of the whole~\citep{yash20}. Alternatively, for ML tasks with data sources that do not obey those statistics, the same survey advises BO strategies for small hyperparameter spaces and PSO for large ones. These suggestions appear to be a good rule of thumb, although the sparsity and limitations of benchmarking studies still preclude any encompassing conclusions. Of course, the literature evaluating AutoML mechanisms and packages~\citep{baal18} will presumably grow over time, especially with the ongoing development of supporting infrastructure, e.g.~OpenML~\citep{vari14, bica17}, MLaut~\citep{kaki19}, etc.

\textbf{Extending the Scope of CASH.} The notion of AutoML, as it has been popularised, is inextricably linked to the topic of optimisation. Whereas standard ML and its search for model parameters can be encapsulated within an ML component, AutoML has sought ways to automate the selection of an ML component itself. Accordingly, ML software is rarely considered an AutoML package unless it has implemented some form of CASH-solver at a high level, and ongoing research continues to seek efficiency gains in this area. However, in under a decade, AutoML has come to represent ambitions far beyond traditional CASH. A recent publication has even pointedly suggested that HPO may not even be necessary for high-performance AutoML~\citep{ermu20}. In effect, the scope of the field is both fluid and expansive, encompassing anything that may benefit the desires of an ML practitioner; this is not limited to maximising ML-model accuracy and minimising runtime. For instance, some users and ML tasks may require ML models/algorithms with significant complexity. In these cases, one ML component may no longer be an effective representation for an entire predictive system.

\section{Multi-component Pipelines}
\label{Sec:MCPS}

The purpose of ML, as expressed in Section~\ref{Sec:Basics}, is to algorithmically ingest data in some fashion to inform the production of a maximally desirable mapping from a query space to a response space, i.e.~$M:\mathcal{Q} \to \mathcal{R}$. While we have thus far implicitly treated $M$ as a simple singular object, the reality is that ML tasks can involve substantially complicated ground-truth functions; users may need to employ correspondingly complex ML models to have any reasonable chance at a good approximation. The problem with this is that arbitrarily general nonlinear mappings are intractable. Complex ML tasks can only be attacked by chaining together simpler data transformations that are individually well understood and better behaving, if not outright linear. Of course, it is still a significant challenge to form an `ML pipeline' of well-tuned components that, when concatenated, acts as a good ML model for a task at hand. So, with AutoML successful in tackling standard CASH, the natural extension is to automate the construction of a multi-component predictive system (MCPS)~\citep{sabu16b, sabu19}.

However, as before, it is worth highlighting semantic ambiguities first. An example of a standard three-component ML pipeline is one that transforms data defined in a query space, pre-cleaned for convenience, through two intermediate `feature' spaces and into the response space, i.e.~$P_{1+2+3}:\mathcal{Q} \to \mathcal{F}_1 \to \mathcal{F}_2 \to \mathcal{R}$. Some data scientists would consider the final segment of this ML pipeline, sometimes called a predictor, to be an ML model, e.g.~an evaluable classifier/regressor, while the previous two segments would be called preprocessors, e.g.~feature engineers. Deep-learning specialists may alternatively call the entire ML pipeline an ML model, in recognition of feature-generating layers being internal to a convolutional neural network (CNN), as well as the fact that a singular backpropagation-based algorithm typically trains all of its layers. We avoid this debate; the abstract language of `ML components', `data transformations' and `parameter controllers' enables a conceptual AutoML architecture to represent an encompassing range of ML scenarios.

\begin{figure}[!htb]
  \centering
  \includegraphics[width=\linewidth]{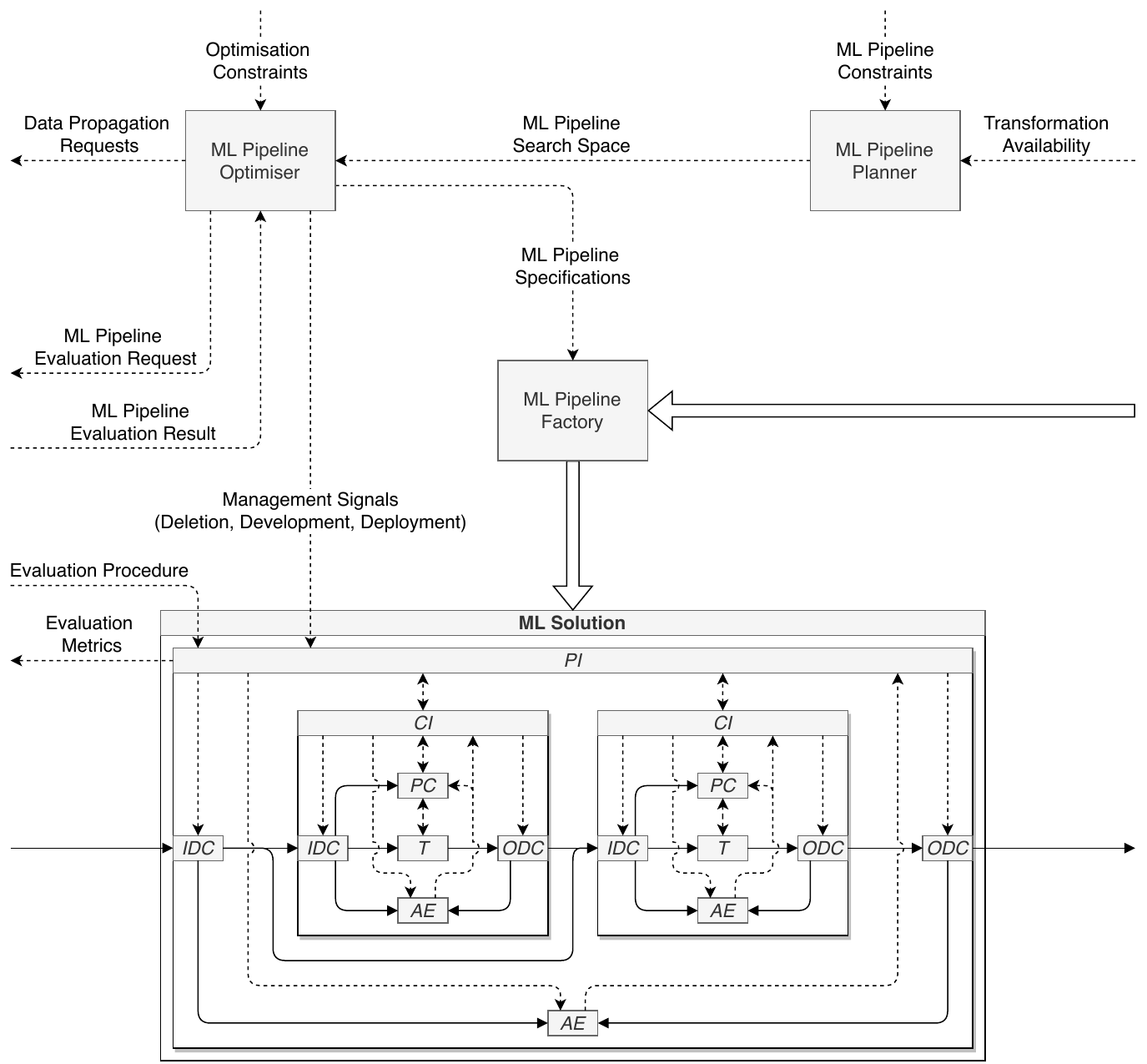}
  \caption{A schematic of a two-component ML pipeline acting as an ML solution within an AutoML system, with immediate modular environment pulled/renamed from Fig.~\ref{Fig:ModelSelection}. A pipeline interface (PI) relays communications between external systems and each component interface (CI), which in turn communicates with the parameter controller (PC) for each data transformation (T). The PI also manages a pipeline-specific inflow-data controller (IDC), which supplies all the IDCs of individual components, and likewise manages a pipeline-specific outflow-data controller (ODC), which receives output from the tail end of the ML-component arrangement. The ML pipeline also maintains its own accuracy evaluator (AE), mimicking the control structure of an ML component. Dashed arrows depict control and feedback signals. Solid arrows depict dataflow channels. Block arrows depict the transfer of ML components/pipelines.}
  \label{Fig:PipelineControl}
\end{figure}

\textit{\textbf{Framework - The MCPS Upgrade.}} Given the ML-component wrapper introduced in Section~\ref{Sec:Basics}, it is relatively simple to fashion a higher-level analogue. Specifically, Fig.~\ref{Fig:PipelineControl} demonstrates how the ML solution within Fig.~\ref{Fig:ModelSelection} can be upgraded into ML-pipeline format. All associated modules in the high-level schematic can now be presumed to fashion entire pipelines, as opposed to solitary ML components; these include the planner, optimiser, and factory. As for the suggested design of an ML pipeline, we recommend similar elements of wrapper-based control to those provided by the ML component. First of all, the ML pipeline should contain an interface that relays communications between external modules and individual components. Optimising an ML pipeline becomes a complex balancing task with respect to the optimisation of individual ML components, and advanced HPO strategies catering to an MCPS must be actionable. This means individual ML-component evaluations, if available, should be disseminated to external modules as well. On that note, while the predictive accuracy of ML pipelines will typically mirror the accuracy of their final component, this is not always the case, particularly if the ML-pipeline output is an aggregation involving multiple ML components. Essentially, installing an accuracy evaluator at the pipeline level allows an ML pipeline to be assessed independently of any ML-component evaluations. Finally, pipeline-specific controllers for inflow/outflow data are also advised, with the pipeline IDC directing dataflow to all component IDCs. This allows expected outputs to be shuttled to wherever an accuracy evaluator is in play, while also allowing each individual ML component to be fine-tuned, assuming that training data can be provided within its relevant feature space.

\begin{figure}[!htb]
  \centering
  \includegraphics[width=\linewidth]{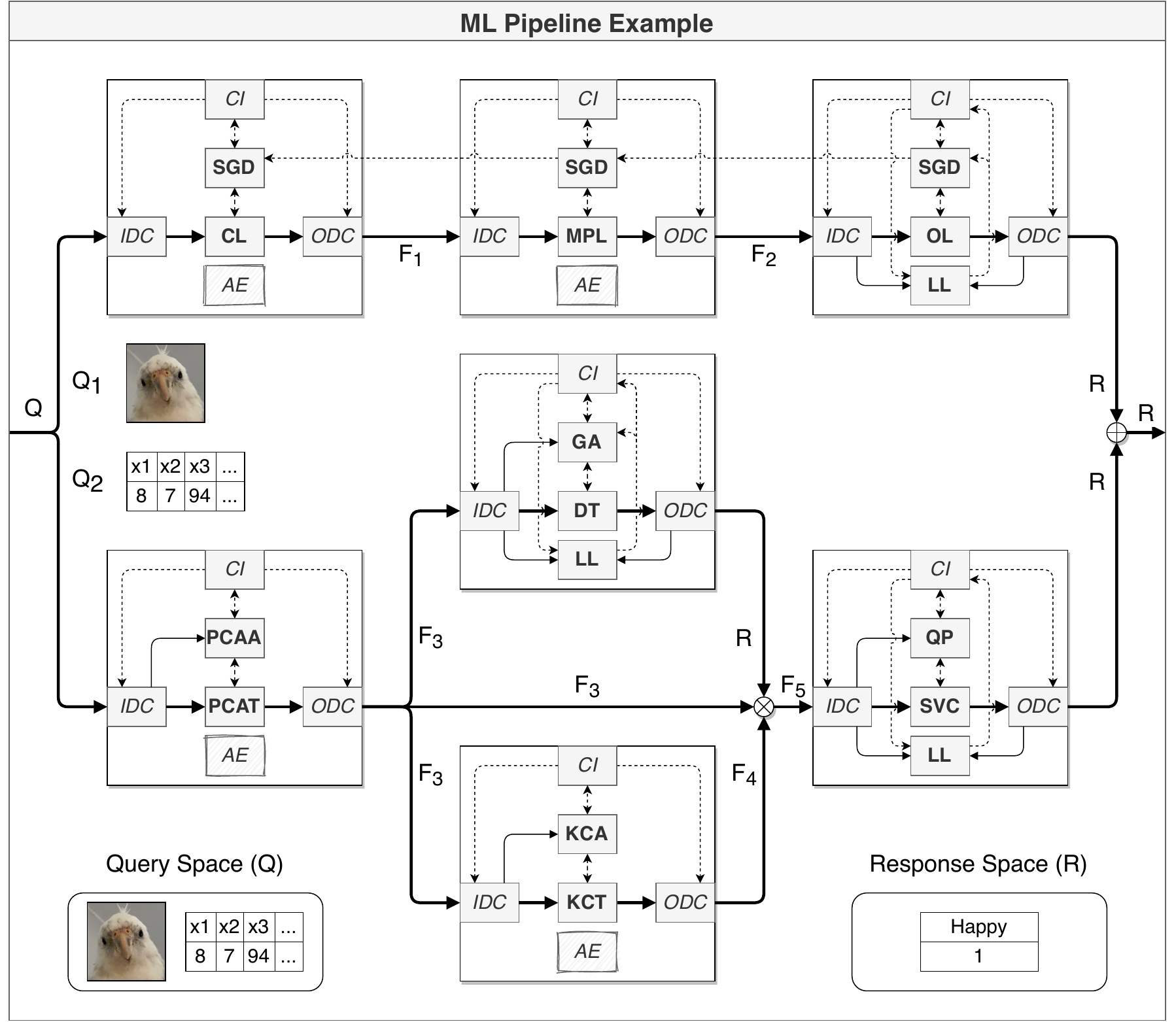}
  \caption{An example of a complex seven-component ML pipeline, mapping queries of mixed image/tabular format to a binary classification response. Each ML component is an instantiation of Fig.~\ref{Fig:Component} in compressed representation, containing a component interface (CI), controllers for inflow/outflow data (IDC/ODC), and an optional accuracy evaluator (AE). Other ML-component abbreviations denote a convolutional layer (CL), max-pooling layer (MPL), output layer (OL), stochastic gradient descent (SGD), logistic loss (LL) calculator, principal-component-analysis transformation/algorithm (PCAT/PCAA), k-means-clustering transformation/algorithm (KCT/KCA), decision tree (DT), genetic algorithm (GA), support vector classifier (SVC), and quadratic programming (QP). Solid arrows depict dataflow channels; thick arrows denote dataflow during deployment. Dashed arrows depict control and feedback signals. Labels along component-external dataflow channels define the spaces that data inhabits at those points; they are detailed in Section~\ref{Sec:MCPS}. The `$\times$ junction' concatenates data from different spaces. The `$+$ junction' averages data within the same space. Pipeline-level control elements, other than query-based dataflow, are omitted from the diagram for clarity.}
  \label{Fig:PipelineExample}
\end{figure}

In principle, the ML-pipeline paradigm enables arbitrary complexity. An example is provided in Fig.~\ref{Fig:PipelineExample}, with most aspects of pipeline control omitted to avoid clutter. Here, each incoming instance of data, defined in $\mathcal{Q}$-space, is a concatenation of an image in $\mathcal{Q}_1$-space and tabular data in $\mathcal{Q}_2$-space. With ML-component IDCs acting as gatekeepers, possibly declaring their permissible inflow spaces to the ML-pipeline IDC via indirect means, the data is essentially partitioned between parallel tracks. Each image is passed through a CNN of three layers, i.e.~convolutional (CL), max-pooling (MPL) and output (OL), so as to be classified in a binary response space, $\mathcal{R}$. This net transformation passes through two intermediate feature spaces, $\mathcal{F}_1$ and $\mathcal{F}_2$. The tabular data instead takes the alternate track, being refined by principal component analysis (PCA) and mapped into feature space $\mathcal{F}_3$. A decision tree (DT) makes an early attempt at classification, $\mathcal{F}_3 \to \mathcal{R}$, while k-means clustering tries to generate a new feature in $\mathcal{F}_4$-space. These outputs are concatenated with the original refined data into a new feature space, $\mathcal{F}_5 = \mathcal{F}_3 \times \mathcal{F}_4 \times \mathcal{R}$; this data is then classified by an SVM, mapping $\mathcal{F}_5$ back to $\mathcal{R}$. A final averaging across the outputs of parallel tracks, i.e.~the image classifier and the tabular-data classifier, results in a final response.

Importantly, this example demonstrates how an ML solution can involve a mix of strategies. For instance, neural networks can be part of an ML-pipeline search, elaborated in Section~\ref{Sec:NAS}, if error evaluations are allowed to propagate between ML components, which was facilitated within Section~\ref{Sec:Basics}. Simple versions of ensembling strategies like `stacking' are also possible on this level~\citep{chwu18}, as shown with the SVM operating on DT outputs. Moreover, the example represents a diversity of algorithms in terms of how they tune their corresponding data transformations. For instance, the depicted optimiser for quadratic programming (QP) does not explicitly take feedback from an evaluator, training its associated support vector classifier (SVC) on instructional data in a principled implicit manner. On the other hand, the depicted stochastic gradient descent (SGD) optimiser bases its updates of associated CNN layers on explicit logistic-loss comparisons between CNN outputs and actual labels. The key point here is that the component representation is significantly flexible; while DTs can be constructed via implicit procedures, expanding out until reaching a threshold of node purity, Fig.~\ref{Fig:PipelineExample} suggests that using a GA to build a DT, driving evolution by explicit evaluations of a rule-based tree, is accommodated too. As for the presence of PCA and clustering, they re-emphasise that unsupervised techniques can be combined with supervised learning. Again, with solid design and implementation on the base level, an ML pipeline can be endlessly inclusive, whether or not the complexity is justified. At the very least, shunting preprocessing operations into the ML pipeline enables AutoML to automate feature engineering~\citep{bo17}, discussed further in Section~\ref{Sec:Features}.

\textbf{Tackling the Selection Problem.} From a historical perspective, automatically selecting a good ML pipeline for an ML task is not a new idea. Several recommender systems were developed in the 1990s and 2000s that suggested series of data-transforming processes; these were often based on meta-learning, discussed in Section~\ref{Sec:Meta}. Likewise, prior to CASH being formalised, there was at least one proposal for an MCPS-based ML architecture~\citep{kaga09}.

It is not unexpected then that full-model-search pioneers tackled ML pipelines from the start. For instance, one PSO-based attempt optimised over a hyperparameter space that encoded whether or not to use individual preprocessors~\citep{esmo09}; this strategy would later be tweaked to divide HPO and ML-pipeline selection between the PSO and a GA, respectively~\citep{supf12, supf13}. Similarly, MLbase, a contemporary of Auto-WEKA, focussed on constructing and optimising `learning plans' from series of data operators~\citep{krta13}. Oddly, given that tree-based SMAC can handle ML pipelines natively~\citep{fekl15, fekl19}, Auto-WEKA seems almost anomalous in limiting preprocessing to feature selection~\citep{thhu13, koth19}, but, in fairness, its development may have prioritised ensembling instead, a topic discussed further in Section~\ref{Sec:Ensemble}.

Since those earlier years, AutoML research has definitively expanded from HPO through CASH to ML-pipeline optimisation~\citep{brgi17}. Some coding packages specifically promote these features, such as AutoWeka4MCPS~\citep{sabu16, sa17}, built directly on top of Auto-WEKA. Notably, appreciating that complex pipelines can be challenging to interpret for users, AutoWeka4MCPS applies Petri net representation~\citep{aa98} to a constructed MCPS for the sake of transparency~\citep{sabu17}. Then there are AutoML systems that fully embrace ML pipelines with theoretically arbitrary complexity, such as the GA-driven Tree-based Pipeline Optimization Tool (TPOT)~\citep{olba16}, which has further explored the parallelisation of feature generation/selection as part of a prostate-cancer case study~\citep{olur16}. The appeal of such an inclusive approach, as indicated by the Fig.~\ref{Fig:PipelineExample} example, comes from the fact that relaxing ML pipelines from linear chains to general directed acyclic graphs (DAGs) increases their representative potential. Attempts to generalise backpropagation for arbitrary ML pipelines have similarly supported DAGs~\citep{miba17}.

\textbf{The Challenges of Concatenation and Complexity.} Designing an MCPS has many issues that must be addressed. Chief among them is ensuring ML components can be validly combined together into an ML pipeline. This is less of a problem when data is clean, numerical and vectorised. Messy real-world data, on the other hand, remains a challenge for learning systems, both adaptive~\citep{zlbi12} and otherwise. Indeed, the Challenges in Machine Learning (ChaLearn) 2015-2016 AutoML competition found that all participants but one were unable to deal with sparse data~\citep{guch16, sugu18}. It is then no surprise that folding missing-value imputation into an ML pipeline requires careful thought, as exemplified by TPOT-based research~\citep{gasa17, game18}. Alternatively, accepting that some pipelines will fail and quickly identifying them is another tactic, which is employed by the recently published `AVATAR' method~\citep{ngma20, ngga20}; this approach avoids the computational cost of setting up and executing an entire ML pipeline by evaluating a simpler proxy.

In practice, most AutoML systems avoid ML-pipeline generality from the outset, using hard constraints to avoid both invalid compositions and unnecessarily bloated search spaces; the planning module in Fig.~\ref{Fig:PipelineControl} enshrines this approach. For instance, if a data scientist faces an ML task involving image-based inputs, they will probably benefit from using convolutional neural layers. Likewise, dataset normalisation is likely to be a transformation employed early on within an ML pipeline. In fact, forming ontologies of operators and linking them with useful workflows has been a priority of the data-mining research field even before the current wave of AutoML~\citep{kise09}. Thus, using similar concepts, MLbase provides one such example in which a `logical learning plan' acts as a template for a `physical learning plan', i.e.~an instantiated ML pipeline~\citep{krta13}. Likewise, whereas TPOT constrains its pipeline search with a Pareto front that factors in the number of ML components within a solution~\citep{olba16}, an alternative GA-driven AutoML framework named `REsilient ClassifIcation Pipeline Evolution' (RECIPE) uses a grammar for finer control~\citep{sapi17}.

Notably, the problem of building optimal ML pipelines within hard constraints shares overlaps with the field of automated planning~\citep{jiro12}. Accordingly, the concept of hierarchical task networks (HTNs) has been adopted to systematically describe valid ML pipelines, whether in the design of data-mining tools~\citep{kise09, kise12} or more recent AutoML systems like ML-Plan~\citep{mowe18a, mowe18}. Researchers behind ML-Plan imply that, due to the recursive nature of HTNs, it is still possible to explore ML pipelines with arbitrarily long preprocessing workflows, matching the generality of TPOT without wasting evaluation time~\citep{wemo18}.

Whatever the design, the actual optimisation of ML pipelines remains challenging, given that the typical CASH already corresponds to optimising a one-component pipeline. Genetic programming~\citep{ko90} is one particularly favoured technique in this context as it natively works with ordered sequences of operators; it is used in TPOT~\citep{olba16} and RECIPE~\citep{sapi17} among other AutoML systems. An asynchronous version of this evolutionary algorithm is implemented by the Genetic Automated Machine learning Assistant (GAMA)~\citep{giva19}. However, SMBOs remain employed as an alternative, with Fast LineAr SearcH (FLASH) as a recent attempt to upgrade Bayesian optimisation for ML pipelines, separating CASH from pipeline search and applying a linear model to error propagation across ML components~\citep{zhba16}. A subsampling method called Bag of Little Bootstraps (BLB) has likewise been used to speed up BO-based pipeline search~\citep{an17, andu17}. Naturally, as with CASH, there are also yet other approaches, such as the MCTS used in ML-Plan~\citep{mowe18a}.

However, to date, there is no conclusively best-performing search strategy for ML pipelines. It is not even clear how an MCPS complicates hyperparameter space, with one exploration of fitness landscapes finding frequent disperse optima and situations where basic grid/random search methods are highly competitive~\citep{gasa18}. Another investigation supports the multiple-optima finding, with repeated ML-pipeline optimisations producing inconsistent results depending on how the search was initialised~\citep{sabu19}. This is not entirely unexpected; the multiplicity of distinct but similarly performing models has attained the nickname of `the Rashomon effect'~\citep{br01}. Nonetheless, it is still an open question as to how AutoML should select an MCPS when faced with this conundrum, let alone whether exhaustive searches are worth the computational resources when an easily found local optimum is `good enough'.

\textbf{Blurring the Boundaries of the MCPS Paradigm.} Ultimately, as optimisation strategies for ML pipelines improve, these principles of automation continue to envelop new settings. For instance, the Hyperopt library has been applied to signal processing pipelines~\citep{hakr16}, while causal impact analysis operators have been considered as extensions to the standard pool of ML components~\citep{huhu17}. Finally, a recently published system named AutoML-Zero has pushed atomicity to the extreme, applying genetic programming to simple mathematical operators so as to build up predictors and learning algorithms from scratch; this notably includes a rediscovery of backpropagation driven by gradient descent~\citep{reli20}. Such edge cases can be challenging to fit within an encompassing framework, including the illustrative one built up in this review, as, despite the pipelining similarities, AutoML-Zero could arguably be described as a one-component HPO problem, just with an extraordinarily high-dimensional hyperparameter space. All the same, this indicates how the diversity of modern AutoML systems can blur the lines of simple categorisation.

\section{Neural Architecture Search}
\label{Sec:NAS}

In the era of DNNs, it was inevitable that the principles of AutoML would be co-opted in automating deep learning. Indeed, the `AutoDL' abbreviation has already been adopted by a recent ChaLearn competition, AutoDL 2019-2020, to draw a symbolic line between the use of tabular data, i.e.~the traditional focus of pioneering AutoML, and data domains that deep learning has excelled in processing~\citep{ligu19}, such as images, videos, speech, and text. The process of automatically finding optimal DNNs and related neural networks for ML tasks has similarly attained the distinct label of `neural architecture search' (NAS)~\citep{elme19, wira19, hezh21, rexi20}.

\textbf{Deep Networks and ML Pipelines.} Neural networks have a special status amongst ML models, in part due to the `expressive power' they are afforded by the universal approximation theorem. A standard feedforward network (FFN) with unrestricted size and a nonlinear activation function, e.g.~the rectified linear unit (ReLU), can fit a continuous function arbitrarily well. Accordingly, there are claims that the performance of several state-of-the-art DNNs comes from their capacity for extremely deep representations~\citep{hezh16}. The theoretical impact of layer width upon approximating-potential has also been studied~\citep{lupu17}. In essence, connectionist models are powerful enough that a deep-learning specialist need not necessarily interface with adjacent topics.

From a conceptual perspective, however, it is difficult to assess whether NAS is truly a separate category of AutoML, despite often being treated as its own chapter of the story~\citep{huko19}. For one thing, society had only just begun to shift its focus to deep learning~\citep{krsu12} during the first wave of AutoML-package releases; it is no surprise that wrappers around ML libraries designed before 2010, e.g.~Auto-WEKA~\citep{thhu13} and Auto-sklearn~\citep{fekl15}, treat neural networks in a limited way. In contrast, AutoML systems that focus on NAS profit off of more recently engineered foundations, e.g.~Auto-Keras~\citep{jiso19} and Auto-PyTorch~\citep{mekl19}. The distinction between AutoML and AutoDL thus appears somewhat superficial, even if a DNN, as a self-contained model, typically has a far more complicated hyperparameter space than any standard alternative, such as an SVM. This is true, but it unnecessarily confounds reductionist approaches to ML automation.

It is arguable instead that NAS ultimately collapses to the problem of ML-pipeline search. Indeed, as discussed in Section~\ref{Sec:MCPS}, non-output neural layers can be considered equivalent to standard ML preprocessors. That is why the schematic of the ML component in Fig.~\ref{Fig:Component} explicitly includes error propagation signals; this allows a DNN to be broken apart into the pipeline representation exemplified by Fig.~\ref{Fig:PipelineExample}. Of course, in fairness, modern neural networks are diverse in nature. Any encompassing AutoML framework requires careful engineering to represent as many scenarios as possible, e.g.~handling internal state to cater for recurrent neural networks (RNNs). Indeed, training RNNs, e.g.~via the DAG-unravelling method of `truncated backpropagation through time', may require more careful consideration if each layer is to be treated as an independent ML component. However, these are mostly matters of implementation.

\textbf{Strategies for NAS.} Notably, while NAS has become a prominent aim of ML research within the last few years, efforts in neural-network selection stretch back at least two decades earlier. Evolutionary algorithms were a popular choice for investigating network design, as shown by a survey from 1999~\citep{ya99}, and other early investigations even explored Bayesian theory for making architectural comparisons~\citep{ma92}. As for codebases, an ML suite named Learn-O-Matic~\citep{sefe12} was described in 2012 as using RL-based `policy gradients with parameter based exploration' (PGPE)~\citep{seos10} to optimise the structure of feedforward networks. This system, although contemporaneous with Auto-WEKA, appears to have flown under the radar within the larger ML community. This is not unexpected given the sheer volume of deep-learning publications; it can be arbitrary as to which advances stick out to the community and inform best practice, let alone standard practice. Sure enough, in similar fashion to standard HPO, both grid search and random search remain default approaches to building neural networks, with the former demonstrated by a mammogram classification study~\citep{fome15}.

That all acknowledged, 2017 witnessed several conference papers that caught public attention and have stimulated intensifying interest in NAS. One of these proposed using an RNN `controller' to construct a `child' network, layer after layer, by sequentially recommending values for structural hyperparameters~\citep{zole17}, e.g.~filter width/height per CNN layer. The RNN controller is trained via RL, driven by performance assessments of the resulting child network when applied to an ML task of choice. In terms of the illustrative framework conceptualised within this review, specifically the elements shown in Fig.~\ref{Fig:PipelineControl}, the child network is essentially an ML pipeline, while the RNN controller is equivalent to an ML-pipeline optimiser. This kind of approach, searching for an optimal sequence of building instructions, also inspired the `MetaQNN' strategy~\citep{bagu17}, which is based on Q-learning applied to Markov decision processes.

Naturally, numerous other search strategies were quickly applied to NAS. Some of them were continuations of long-term neuro-evolutionary research; these were shown to produce DNNs that are highly competitive with other state-of-the-art networks~\citep{reag19}, as, for instance, judged on typical Canadian Institute For Advanced Research (CIFAR) benchmarks~\citep{remo17, sush17}. Alongside the standard principles of sampling high-performance candidates from a population, so as to iteratively breed new generations of neural networks, these mechanisms typically involve encoding networks by some genetic representation, so that mutation operators are capable of both adding/removing layers and modifying structural/training hyperparameters. Such evolutionary NAS methods have been used to, for example, design better transformer architectures~\citep{sole19}. Alternatively, complementing GAs and RL, another major class of NAS approaches revolves around gradient optimisation. Differentiable ARchiTecture Search (DARTS) is an archetype of this strategy~\citep{lisi19}, which eschews discretisation and aims to relax network representation into a continuous space. This allows both connection weights and network architecture to be tuned as part of a single bi-level optimisation problem.

Not unexpectedly, the inevitable alignment of AutoML and NAS also brought BO methods into the fold. Auto-Net~\citep{mekl16} is a SMAC-based AutoML package released before the current upsurge in NAS interest. Despite, in principle, being capable of searching for fully-connected FFNs of arbitrary depth, its initial publication limited the number of layers to six for practical reasons. More recently, other groups of researchers from the same institution that published Auto-Net have criticised several prominent NAS methods, both RL-based~\citep{zole17} and GA-based~\citep{remo17}, for optimising network structure independently of training-algorithm hyperparameters~\citep{zekl18}. Unfortunately for the AutoML community, this partition has now become widespread enough that, in AutoDL convention, the term `hyperparameters' often refers only to the latter, which leads to a confusingly fuzzy scope for HPO; technically, NAS is a form of HPO, but AutoDL separates the two. In light of this convention, a publication presenting an upgrade of Auto-Net argues that NAS and `HPO' should be solved simultaneously. The upgraded package, Auto-PyTorch, utilises BO-HB instead of SMAC, while also extending configuration search space to include modern deep-learning techniques and structures~\citep{mekl19}. Motivated by similar concerns, BO-HB has also been used in a combined hyperparameter and architecture search (HAS) when constructing an RL policy network for the design of ribonucleic acid (RNA) molecules~\citep{rust19}. Since then, the importance of HAS has increasingly gained traction, with the recent publication of an `AutoHAS' approach likewise grappling with how to specify configuration space across all levels, albeit for a gradient descent strategy as opposed to BO~\citep{dota20}.

\textbf{Lightening the Computational Load.} Optimisation approaches aside, the topic of NAS proves most valuable to general AutoML research as a case study of an MCPS taken to extremes. Training a single neural network on non-tabular datasets can be computationally expensive, and this cost can balloon dramatically when reiterated numerous times throughout an exceedingly complex search space. A 2020 survey of modern NAS methods lists the number of GPU days each one took to learn from the CIFAR-10 and ImageNet datasets~\citep{rexi20}; values in the hundreds and thousands are not uncommon. Granted, there is debate in the NAS community about how to fairly report GPU days, especially when experiments are not run on standardised devices, but it remains a fact that computational costs are immense and require careful treatment~\citep{reag19, sole19}. They can be further inflated in the full HAS setting. Unsurprisingly, there has also been research into HPO for neural networks specifically, e.g.~constraining search spaces for residual neural networks (ResNets) by assessing the importance of individual hyperparameters~\citep{shri19}. Notably, proposed HPO tricks sometimes lean on factors specific to AutoDL, such as the fact that DNN weights are usually optimised by gradient descent. One investigated approach thus uses implicit differentiation to tune weights and hyperparameters in conjunction, rather than training one candidate model per each hyperparametric configuration~\citep{lovi20}.

Returning to NAS specifically, the obvious approach to make architectural search manageable is to artificially constrain allowable configurations. This harks back to the HTNs employed for general ML pipelines, as was described in Section~\ref{Sec:MCPS}. A popular practice in recent years is to construct networks with `cells'~\citep{wira19}, complex substructures that can be reused multiple times within larger templates. So, for instance, a CNN may consist of a number of cells interleaved with input/output and pooling layers; NAS strategies need to decide how many cells to stack together and what the contents of an individual cell are, but this is a significantly smaller search space than optimising cells independently. This approach was popularised by the NASNet search space~\citep{zova18}, although there have been many variations since, e.g.~ProxylessNAS~\citep{cazh19}. Nonetheless, it is an open question whether the simplicity afforded by cell-based search, alternatively called micro-NAS, is worth the loss of layer diversity afforded by `global' search~\citep{hula18}, sometimes called macro-NAS.

Beyond constraining search space, many efficiency-based research threads in NAS relate to establishing intelligent shortcuts in the ML-pipeline search process, such as recycling architectures and sharing trained parameter values between candidate networks~\citep{rexi20}. The popular notion of weight-sharing~\citep{phgu18}, where candidate models are subgraphs sampled from a larger DAG, was originally employed for an RL-based NAS method. The idea has since been fused with differentiable NAS~\citep{doya19}, and even randomly sampling weight-sharing models has proved highly competitive~\citep{lita19}. Another example of an efficiency boost is leveraging functionality-preserving morphisms, implemented by the Auto-Keras package, to iterate effectively through potential networks~\citep{jiso19}. Notably, in return, Auto-Keras seems to require a decent initial configuration, which will depend on the ML task at hand. In certain cases, it may be sufficient for a user to suggest this starting point, but other NAS approaches lean heavily into transfer learning to automate this jump-start. For instance, one proposed extension to the original RNN-controller scheme~\citep{zole17} is to include an embedding vector that represents a diversity of ML tasks, such that a fully trained RNN-controller is able to suggest a strong candidate DNN for any new ML task, based simply on which previously encountered ML task this new one is similar to~\citep{woho18}. Then there is few-shot learning, an extreme variant of transfer learning, where a well-performing but generic DNN is sought out to serve as a near-optimal initial configuration for a NAS attempt~\citep{fiab17}. Many of these speed-up mechanisms are based on or adjacent to meta-learning; see Section~\ref{Sec:Meta}.

\textbf{The Potential and Pitfalls of NAS.} For now, it is too early to make conclusive remarks about NAS as it relates to AutoML as a whole. While the fine details can vary dramatically, reviewed more broadly/deeply elsewhere~\citep{elme19, wira19, rexi20}, NAS appears to be a subset of ML-pipeline search; its elements are able to be represented by the abstractions within Fig.~\ref{Fig:PipelineControl}. However, the research area is constantly evolving, subject to a lot of attention, and novel network designs are constantly being proposed, with some examples mentioned in Section~\ref{Sec:AGI}. At some point, with an ongoing trend towards packaging network design for biomimetic neurons~\citep{dusu19}, the principles of NAS may even be extended to spiking neural networks, although how best to approach such a fusion remains extremely speculative. Regardless, as a nascent field stimulating unbridled exploration, NAS has shorter-term problems to address. For instance, robust evaluations of network topology and other NAS outputs~\citep{doli21} are rare. More troublingly, a recent benchmark comparison of random search with several state-of-the-art NAS strategies has found no competitive distinction, simultaneously challenging the effectiveness of constrained search spaces and weight sharing strategies~\citep{yusc20}. This result supports a previous assessment, likewise involving random search, that additionally bemoans a lack of reproducibility within the field in general~\citep{lita19}. These have accompanied similar criticisms about the lack of ablation studies, necessary to identify the novel aspects of NAS strategies that truly advance efficiency/performance, as well as a publication bias that does not highlight important negative results~\citep{gegi19}. 

Evidently, the research area of NAS remains far from fully mature. However, it is still linked with significant achievements and an ever-expanding body of work. Indeed, while the intent of this section has been to show that NAS does not contradict the basis of an integrated conceptual framework, there is much more to say on the topic that goes outside the scope of this monograph. Simply put, the practicalities and challenges of deep learning add their own flavour to AutoML, certainly along the entire ML workflow depicted in Fig.~\ref{Fig:Workflow}, and the resulting field of AutoDL is worthy of being reviewed elsewhere beyond the central focus of NAS.

\section{Automated Feature Engineering}
\label{Sec:Features}

Real-world data for any ML task is rarely structured in an informative and discriminative manner. One way to face this challenge is to design complex predictors capable of drawing highly nonlinear classification boundaries or regression slopes within these difficult data spaces, e.g.~SVMs with exotic kernels. Alternatively, one may seek to nonlinearly morph incoming data until it sits within a feature space that can be mapped to expected outputs in a simple fashion, perhaps even linearly. This latter approach is called feature engineering (FE), and its automation falls under the scope of AutoML research.

\begin{figure}[!htb]
  \centering
  \includegraphics[width=\linewidth]{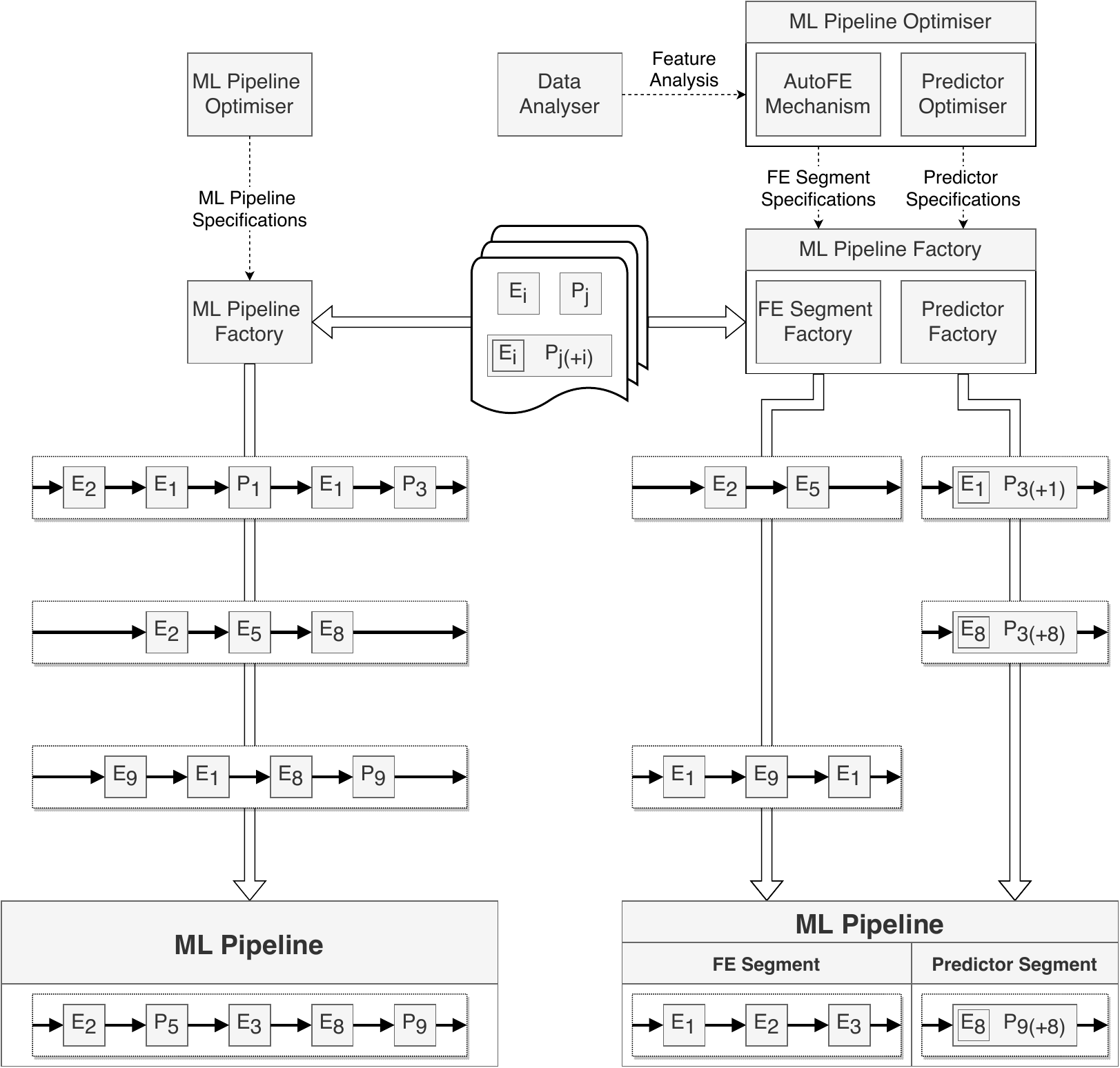}
  \caption{A demonstration of two different approaches to ML-pipeline search, with E\textsubscript{i} and P\textsubscript{j} representing pools of feature-engineering (FE) and predictor ML components, respectively. ML components marked as P\textsubscript{j(+i)} are complex predictors that internalise FE transformation E\textsubscript{i}. The left-side approach composes candidate pipelines freely, preferencing simple predictors. The right-side approach subsections pipelines between FE components and complex predictors, selecting segments independently. The right-side ML-pipeline optimiser is provided the feature analysis of data to support `filter-type' AutoFE mechanisms. Dashed arrows depict control and feedback signals. Solid arrows depict dataflow channels. Block arrows depict the transfer of ML components/pipelines.}
  \label{Fig:FEStrategy}
\end{figure}

\textit{\textbf{Framework - Preprocessor Management.}} In principle, like NAS, automated feature engineering (AutoFE) is related to, if not subsumed by, the topic of ML pipelines examined in Section~\ref{Sec:MCPS}. Any early-stage operation, e.g.~outlier handling or one-hot encoding, can be considered as a data transformation deserving of its own ML component. This immediately opens up FE-heavy ML pipelines to optimisation approaches discussed earlier. However, despite the overlap, there are still important nuances to consider. Broadly speaking, MCPS studies are motivated by extending HPO to extreme configuration spaces, without fixation on what those components are. For instance, defining a predictor to be a mapping onto response space $\mathcal{R}$, and thus being evaluable with respect to expected outputs, an ideal MCPS optimiser would have no problem searching arbitrary ML pipelines with early-stage predictors, as exemplified by the left side of Fig.~\ref{Fig:FEStrategy}. These candidates are not all invalid or suboptimal either; chaining predictors into stacked ensembles, as suggested by Fig.~\ref{Fig:PipelineExample}, may ultimately be the best solution for an ML task. Admittedly, in practice, a large fraction of MCPS research does deal with structural constraints, but these are a secondary consideration in service of making optimisation feasible.

In contrast, AutoFE is grounded in structural constraints, accepting the traditional notion that FE and prediction are unique and well-ordered sections of the data-processing sequence. In light of how intractable unconstrained ML-pipeline search can be, this attitude is not unreasonable. It is convenient to clump together both algebraic transformations and an expansive search through them under one feature-generation banner, just as it is convenient to have one monolithic process for combinatorial feature-subset selection. In light of this, the AutoFE approach can be described abstractly as an attempt to subdelegate the responsibility for MCPS optimisation, as demonstrated by the right side of Fig.~\ref{Fig:FEStrategy}.

However, because ML-pipeline optimisation is so novel, the concept of partitioning is under-explored. For example, a CNN is considered a complex predictor, where the initial layers are typically responsible for identifying features within images. The question arises: should these layers remain subject to a NAS procedure, or should they be selected/optimised externally as part of an AutoFE process? Expressed in another way, given three consecutive indivisible operators, i.e.~$U: \mathcal{Q}\to\mathcal{F}_1$, $V:\mathcal{F}_1\to\mathcal{F}_2$ and $W:\mathcal{F}_2\to\mathcal{R}$, should $V$ be targeted by FE optimiser $\Omega_\mathrm{FE}$ or predictor optimiser $\Omega_\mathrm{Pred}$? Ideally, there would be no difference, assuming that both optimisers treat $V$ as a degree of freedom, have access to the same CASH-related search space, and have perfect information regarding the evaluation metrics of the $W \circ V \circ U$ composition as a whole. In actuality, these assumptions are rarely true, leading to the following inequality:
\begin{equation}
    \Omega_\mathrm{Pred}(W \circ V) \circ \Omega_\mathrm{FE}(U) \neq \Omega_\mathrm{Pred}(W) \circ \Omega_\mathrm{FE}(V \circ U).
\end{equation}
In essence, it is not simple to decide whether the onus of wrangling a topological transformation, $\mathcal{F}_i\to\mathcal{F}_j$, should fall to AutoFE or predictor design.

Regardless, despite the ambiguities around ML-pipeline segmentation and the generality invited by the MCPS paradigm, hard boundaries between preprocessors and predictors remain heavily ingrained within the ML community. Many of the MCPS-based AutoML systems discussed in Section~\ref{Sec:MCPS} either implicitly or explicitly subdivide ML-pipeline search, as conceptualised in Fig.~\ref{Fig:FEStrategy}. As a result, jointly optimising both segments is considered somewhat novel~\citep{uksa18}. Additionally, with priority focus on selecting classifiers/regressors, many systems also lean towards complex predictors, where feature transformations are treated as an embedded hyperparameter rather than a unique and free-floating component. The right side of Fig.~\ref{Fig:FEStrategy} implies this; P\textsubscript{j(+i)} is a monolithic ML component, even though it represents P\textsubscript{j} learning from a data remapping performed by E\textsubscript{i}. This internalisation of feature-space transformation is not an intrinsic weakness, with the SVM kernel trick providing a well-known example of where it proves useful.

\begin{figure}[!htb]
  \centering
  \includegraphics[width=\linewidth]{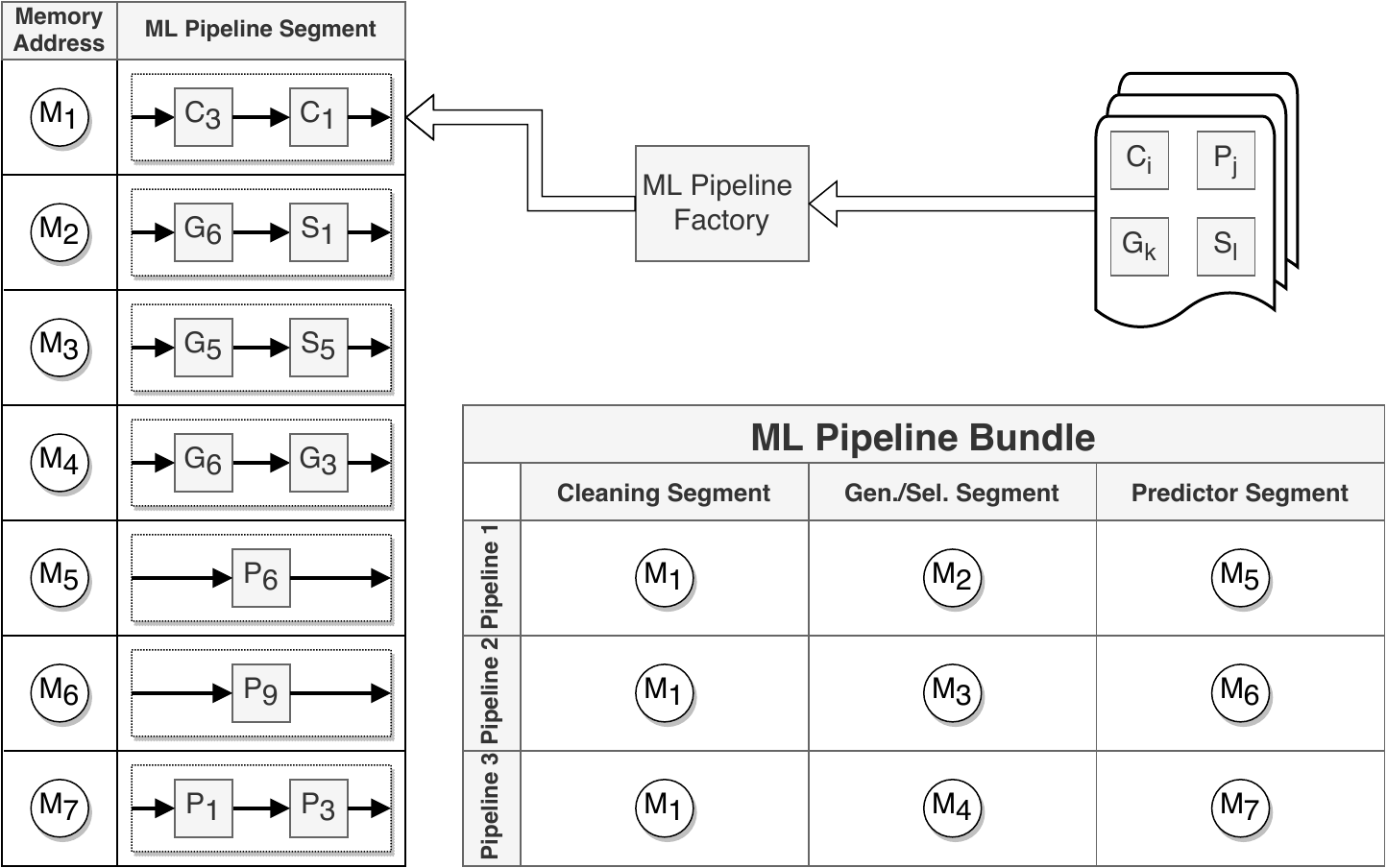}
  \caption{An example of sharing ML-component compositions across ML pipelines via pointer/reference. Label C\textsubscript{i} represents data-cleaning operations, P\textsubscript{j} represents predictors, G\textsubscript{k} represents feature generation, and S\textsubscript{l} represents feature selection. Solid arrows depict dataflow channels. Block arrows depict the transfer of ML components/pipelines.}
  \label{Fig:FEShared}
\end{figure}

Notably, although subdividing ML-pipeline search space trades off generality for manageability, this may not be as limiting to AutoFE as cell-based search may be to NAS. Preprocessing does tend to have a natural order, with data-cleaning operations typically required at the start of an ML pipeline. While there are several ways to, for instance, choose how to deal with missing values, with the options possibly represented by a tunable hyperparameter, there is no point testing ML-pipeline compositions with a late-stage imputation operator. Their validity, or lack thereof, can be inferred immediately. Accordingly, AutoFE very much profits from further structural constraints, perhaps in the form of the HTNs that were introduced in Section~\ref{Sec:MCPS}. In anticipation of ensembled ML pipelines, discussed in Section~\ref{Sec:Ensemble}, it is also worth highlighting that the preprocessing operations chosen for one composition are unlikely to differ from those for another, at least in relation to cleaning procedures. Any robust implementation of AutoML should consider working with pointers and references, so that, as Fig.~\ref{Fig:FEShared} demonstrates, the optimisation of one preprocessing segment carries over to all ML pipelines.

\textbf{A History of Generation and Selection.} Setting aside the discussion of how it fits within a larger AutoML environment, the automation of FE has been a research interest for several decades. The concept of deriving new data attributes from an existing set has taken several names; for instance, alongside `feature extraction'~\citep{ri76}, there was also `constructive induction'~\citep{mi86}. Much of the research associated with this topic, especially in the late 1980s and 1990s, is couched in the language of logic and expert systems, with feature transformations often applied in the form of Boolean operators~\citep{pa98}. Even so, several frameworks were developed early on in this space with the intention of automating feature generation via the chained application of constructive operators. The CITRE system is one such example, learning a decision tree from one collection of features to suggest a novel alternative set of logically compounded attributes, before retraining decision trees from the new feature set and recycling this process until converging to an optimum~\citep{mare89}. Although some of its contemporary frameworks would either ask a user to manage or just outright ignore the ballooning of the constructed feature set, the CITRE system was also among a few that pursued autonomous operations by ranking/pruning, with judgements of `quality' based on information-theoretic measures, e.g.~how usefully discriminative a new attribute is. This form of AutoFE proved beneficial in both accuracy gains and structural simplifications for the decision-tree predictors that were worked with.

Naturally, while some efforts focussed on how best to construct new features, optimal subset selection received its own attention. Central to this is the problem of determining which attributes of a data instance are most relevant to classification/regression. Excessive numbers of features can significantly slow down and worsen ML-model training via the curse of dimensionality, and correlations/redundancies can outright destroy the performance of certain predictors, e.g.~Naive Bayes classifiers. By the late 1990s, feature-selection methods were roughly categorised under `filter' and `wrapper' approaches~\citep{kojo97}. Filters involve determining useful attributes via dataset analysis, with no regard for the subsequent predictor. The aforementioned CITRE system somewhat exemplifies this, even though its quality-based filtering is technically embedded into the training of a decision tree. Not unexpectedly, surveys of the time list and debate many ways to assess the relevance of a feature, e.g.~the degree of unique correlation between class labels and the values of a feature~\citep{blla97}. In contrast, wrappers evaluate whether a feature is useful based on the error rate of a subsequent predictor. Thus, feature-selection filters can be considered fast and generic but also hit-and-miss, while wrappers are slow but accurate. As an aside, the right side of Fig.~\ref{Fig:FEStrategy} acknowledges the filter strategy by allowing advisory feature analysis to be received by the ML-pipeline optimiser.

In any case, while hardware and procedural details have improved over the decades, the overarching concepts of automating FE generation/selection have generally remained the same. Instead, pools of feature-constructing components have been expanded over time, exemplified by the Feature Incremental ConstrUction System (FICUS)~\citep{maro02}, which applies selection in a filter-based manner. Similarly, the Feature Discovery algorithm (FEADIS) promotes the inclusion of periodic functions, also serving as an alternative example of a greedy wrapper-style system, i.e.~one that checks whether each newly proposed feature improves predictor performance~\citep{dore12}.

\textbf{Modern AutoFE.} More recently, as of 2015, the Data Science Machine (DSM) and its Deep Feature Synthesis algorithm have acquired somewhat of a pioneering status in the modern AutoML wave, possibly due to their performance in several ML competitions~\citep{kave15}. The DSM is presented as an end-to-end system, thus going beyond AutoFE and also applying Bayesian HPO to a random-forest predictor. As an AutoML system designed for extracting features from relational databases, it served as an inspiration for the `one-button machine' (OneBM), which focussed on extending relational graphs to unstructured data~\citep{lath17}. This thread of research later led to training relational RNNs rather than searching pools of pre-specified operators for the purpose of representing optimal feature-generating transformations~\citep{lami18}. The DSM system was also acknowledged in the release of ExploreKit, which works with non-relational tabular data instead and, reminiscent of wrapper-based FEADIS, uses a pool of general operators to build up its features~\citep{kash16}. It would, in turn, serve as a direct inspiration for the `Feature Extraction and Selection for Predictive Analytics' (FESPA) method, which claims distinctiveness by implementing regression rather than classification, while also aiming to keep feature generation and selection as well-separated processes~\citep{bo17}.

As previously implied, AutoFE is dogged by the computational complexity of searching such a fine-grained space of operators, especially as most modern systems are wrapper-based and require significant time to train their predictors per pipeline evaluation. Neural networks have certain go-to methods, with auto-encoders sometimes employed to find compact feature-space representations of data instances, e.g.~in a competition studying online course dropouts~\citep{co16}. However, in both deep-learning and general contexts, research attempts explore optimisation methods to hone more efficiently onto optimal FE pipelines. Evolutionary algorithms are one such approach for feature selection~\citep{xuzh16}, with genetic programming being an obvious choice for chains of operators, in similar fashion to MCPS approaches discussed in Section~\ref{Sec:MCPS}. Example implementations have been proposed for AutoFE in specific applications, like compiler-based loop unrolling~\citep{lebo09}, or in general contexts, like with DNNs~\citep{he17}. Unsurprisingly, reinforcement-based strategies such as Q-learning have also been proposed, noting that the DSM and contemporaries are often bogged down by searching unnecessary portions of feature-transformation space~\citep{khsa17}. Beyond optimisation, other mechanisms have also been explored to boost efficiency. For instance, a tool called Zombie groups raw data by similarity prior to processing, so that useful data instances are prioritised by feature-generation code and, presumably, subsequent incremental learners~\citep{anan16}.

To date, AutoFE approaches have been applied in various domains. Natural language processing (NLP) is often more challenging than tabular data to generate features for, but efforts have been made to automatically extract features via entity-entity relationships present in semantic knowledge bases~\citep{chka11}. Similarly dealing with NLP, RaccoonDB is a system that accelerates feature extraction from social media for nowcasting~\citep{anan16}. These approaches often deal with immense online sources of data, which are not practical to load into local memory, and it is an open question how best a general AutoML implementation should interface with these inputs. As for tabular formats, wrapper-based feature extraction has been explored in the context of condition-based aircraft maintenance, where GA-based search seemed to be the best performing out of several tested optimisation routines~\citep{gega16}. In the clinical setting, as part of designing a `prediction tool using machine learning' (PredicT-ML), features have been condensed from datasets via numerous temporal aggregations and operators, subsequently selected in a filter-based manner with metrics such as information gain~\citep{lu16}. Elsewhere, a defensive publication has proposed using decision-tree techniques for feature selection in the context of training neural networks to predict data-centre outages~\citep{coro18}. This diversity of applications suggests that perhaps AutoFE may be more valuable to the broader community than predictor-specific CASH.

\textbf{AutoFE Meets Meta-learning.} To conclude this section, it is worth mentioning that many modern AutoFE approaches do not search for an optimal FE pipeline segment from scratch. For instance, the Cognito system, a framework using greedy exploration to traverse a tree of feature transformations~\citep{khtu16}, all prior to adopting an RL-based strategy~\citep{khsa17}, eventually implemented a `learner-predictor'; this module uses historical datasets to recommend FE transformations~\citep{khna16}. Similarly, ExploreKit~\citep{kash16} and FESPA~\citep{bo17} both augment their AutoFE processes with prior knowledge, if available. Moreover, there are recommender systems based on historic datasets that solely target feature generation~\citep{nasa17} or feature selection~\citep{waso13, pale17, pale21}. This idea of leveraging experience from beyond a current ML task is an attractive one, its potential benefits not limited to just AutoFE.

\section{Meta-knowledge}
\label{Sec:Meta}

Thus far, the previous sections have framed AutoML and the search for a task-optimal ML model as an optimisation problem. Whether in the context of an ML component or the context of an ML pipeline, whether for NAS or for AutoFE, many fundamental advances of this past decade have revolved around how to represent complex search spaces, how to constrain them appropriately, and how to apply search strategies efficiently. This focus is not surprising; optimisation is arguably the purest mechanism for identifying a good ML solution. It was certainly acknowledged as one way to approach algorithm selection, back when this problem was posed in the 1970s~\citep{ri76}. However, even in the modern era, training/testing each iteration of proposed ML model can take substantial time. So, given the theory/hardware limitations of the intermediate decades, making large-scale optimisation infeasible beyond ranking small sets of candidate solutions, ML practitioners had to rely on different tactics. In particular, the concept of meta-learning was popularised, mechanically leveraging knowledge gained from previous learning processes to support ML-model selection for a new task~\citep{br98}. As a form of AutoML before the `AutoML' abbreviation was coined, and briefly supplanted by communal interests in optimisation, meta-learning has reacquired status as a potentially empowering upgrade to standard model-searching systems~\citep{albu15, brgi17}.

\textbf{The Rationale for Recommender Systems.} The core foundations of meta-learning were motivated long before the term itself entered common parlance. For instance, when algorithm selection was initially codified as a research question, one proposed angle of attack was to find ways of classifying and categorising ML problems, thus identifying similarity-based groupings for which ML models/algorithms would be particularly effective~\citep{ri76}. This appeared to be a solid strategy; any hope in finding one super-algorithm that would be maximally performant for all settings was dashed by the publication of the no-free-lunch theorems in ML~\citep{wo96} and optimisation~\citep{woma97} during the 1990s. In essence, the theorems state that the average performance of all algorithms across all problem domains are equivalent. Algorithm selection can only be biased towards high performance by linking both a new ML problem and its context to prior experience and then acting on this `meta-knowledge'. This form of `learning to learn' appears biologically justified, with studies of pre-schoolers identifying that humans acquire the capacity to ``search for underlying commonalities'' at an early age~\citep{brka88}.

However, as with earlier sections, first a caveat: terminology is in flux and the boundaries of the meta-learning topic evolve over time. To add to the confusion, `meta-learning' has sometimes been used to describe ensemble-based approaches, where the `meta' prefix refers to classifiers composed of classifiers~\citep{chst93}. Even disregarding this obvious semantic discrepancy, multiple communities have taken the algorithm-selection problem and related meta-learning approaches in different directions, developing inconsistent terminology and arguably suffering from a lack of interdisciplinary communication~\citep{sm08}. Fine tweaks to definitions are also not uncommon; whereas meta-learning has always involved the transfer of meta-knowledge from different domains/problems, learning from a previous training run on the same dataset is now also considered under the same umbrella term~\citep{lebu15, brgi17}.

As a process, while meta-learning is often employed in one-off research investigations, there have been several proposals for how to encase its automation within an idealised general architecture~\citep{vigi04, grja07, jadu11, jagr11}. In similar vein to the illustrative framework developed in this review, these architectures aimed to systematise the principles underlying a preceding flurry of published systems, where the implementations were based on the Knowledge Discovery in Databases (KDD) process~\citep{fapi96}, a template for data exploration. Indeed, while KDD is a broad topic and associated toolkits are varied, many intelligent discovery assistants (IDAs) were designed to support the construction of data-processing pipelines via meta-learning principles, often providing recommendations by analysing datasets and relating them to previous experience. They have been surveyed and reviewed extensively elsewhere~\citep{va11, seva13, abad17}. Importantly, several were even capable of automatically recommending predictors, despite the lack of optimisation, thus acting as AutoML forerunners. These days, pure IDAs appear to have waned in relative importance, with the role of meta-learning evolving to, primarily, support CASH~\citep{va18}. This HPO-driven paradigm shift in the way AutoML uses meta-learning is made stark when comparing relevant workshops for the 20th and 21st European Conferences on Artificial Intelligence (ECAI), held in 2012~\citep{vabr12} and 2014~\citep{vabr14}, respectively.

\begin{figure}[!htb]
  \centering
  \includegraphics[width=\linewidth]{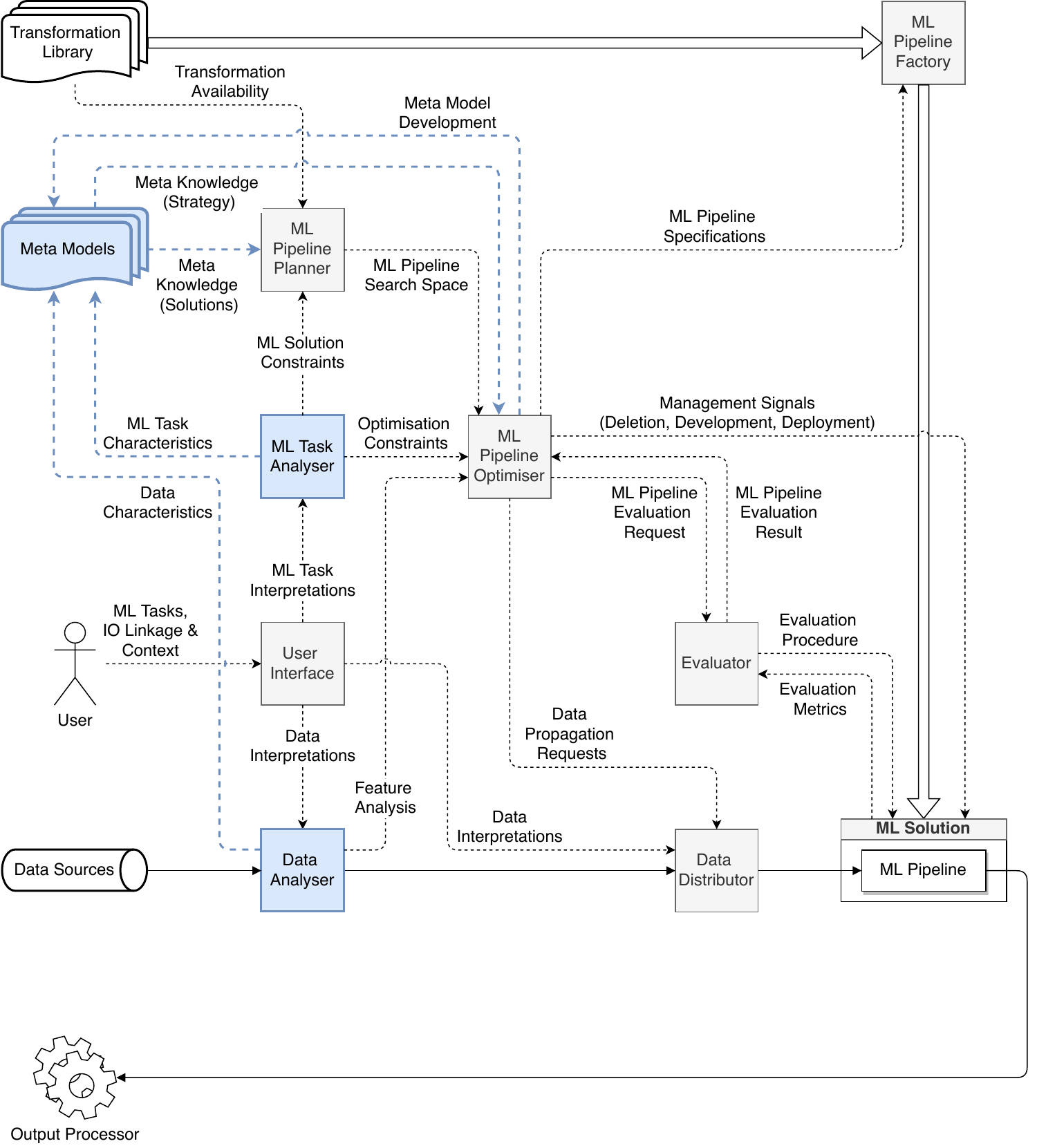}
  \caption{(Colour online) A high-level schematic of an improved AutoML system that employs meta-learning for search-strategy and solution recommendation. This architecture is an upgrade to that of Fig.~\ref{Fig:ModelSelection}, with blue highlights emphasising the main new additions related to meta-knowledge. Dashed arrows depict control and feedback signals. Solid arrows depict dataflow channels. Block arrows depict the transfer of ML components/pipelines.}
  \label{Fig:PriorKnowledge}
\end{figure}

\textit{\textbf{Framework - Exploiting Prior Experience.}} Before discussing meta-learning in depth, we provide an illustration, via Fig.~\ref{Fig:PriorKnowledge}, of how the concept may be integrated into an AutoML framework. In effect, this schematic is another sequential upgrade of the CASH-focussed system depicted in Fig.~\ref{Fig:ModelSelection}, expanding beyond the inclusion of ML pipelines that was shown in Fig.~\ref{Fig:PipelineControl}. The data-analysis module and its associated feature analysis, as recommended by Fig.~\ref{Fig:FEStrategy} to support subdelegated AutoFE optimisation, are also present. However, for this upgrade, the proposed analysis modules for ML tasks and data sources have the additional responsibility to extract metrics for task/dataset similarity. A library of `meta-models' stores the accumulation of previous experience, and, according to how recognisable the current context is, can provide an AutoML system with advice. Typically, this comes in one of two forms: solutions that worked well previously, or solution-finding procedures that worked well previously. Logically, these suggestions are best sent to the ML-pipeline planner or the ML-pipeline optimiser, respectively. Additionally, an ideal AutoML system should not just leverage meta-knowledge but also contribute to it. Hence, the optimiser, which has a view of its own performance along with that of any candidate solution, is charged with passing back results for the sake of development.

Naturally, the usual disclaimer applies, in that Fig.~\ref{Fig:PriorKnowledge} does not necessarily represent the only way to design a functional AutoML system. Critically, though, while CASH is unlikely to vary dramatically in integration, meta-learning represents a methodology as opposed to a method and is thus much harder to conclusively encapsulate. Any module, from UI to data distributor, could conceivably be designed to improve based on previous experience. Nonetheless, given that the vast majority of published research examines meta-knowledge in the context of ML-solution search, Fig.~\ref{Fig:PriorKnowledge} is sufficient to abstractly represent most approaches. Discussion around related limitations is thus reserved for Section~\ref{Sec:Discussion}.

\textbf{Meta-features and Meta-models.} When it comes to the details of meta-learning, the most common tactic is to seek similarity between datasets. Characterising them by useful `meta-features' is thus a necessary challenge and has been extensively explored. Obvious metrics include directly observable descriptors, e.g.~the number of data instances or features, as well as statistical measures, e.g.~variances, and information-theoretic meta-features, e.g.~attribute entropy~\citep{ge17}. For supervised-learning classification specifically, complexity measures aim to categorise a dataset by the nature of its separating boundaries, quantifying feature overlap, class separability, and geometries/topologies/densities of class-spanning manifolds~\citep{hoba02}. Analogous complexity measures have been proposed in the context of regression problems~\citep{loma17}. Naturally, data contexts that contain more structure provide more categorically supportive information, with discriminative characteristics proposed for both time series~\citep{lega10, lega10a, fujo14} and network-oriented datasets~\citep{rimu11}. These latter meta-features were later employed in an AutoML system specifically focussed on biological ecosystem networks~\citep{bamu18}. However, it should be noted that most meta-features described thus far arise from operations on the data itself. One ML-model recommender named AutoDi explores a different approach~\citep{vagr18}; alongside ML algorithms, datasets are assigned textual descriptors in the form of embedding-vectors that are derived from online corpora, e.g.~academic papers and Wikipedia.

In similar fashion to the filter/wrapper dichotomy in AutoFE, some strands of research have considered whether a prediction process should be involved in characterising an ML problem. For instance, several efforts have explored training archetypal predictors such as decision trees on a dataset, so as to use the model-based properties for dataset characterisation~\citep{be98}. More recently, in the context of automating semi-supervised learning, where not all data instances are labelled, outputs of clustering algorithms serve the same purpose as the aforementioned decision trees; cluster-based statistics of those resulting models are used as ML-task meta-features~\citep{liwa19}.

More common than describing a dataset by the properties of a trained ML model is describing it by the performance of that model. This approach, given the name `landmarking' in 2000~\citep{pfbe00}, can be risky if an ML algorithm is overly stochastic in the model it produces. However, it is argued that simple high-bias learners like linear discriminants or decision stumps are quick and robust estimates of ML-problem complexity~\citep{fupe01}. These landmarkers have thus found use alongside traditional dataset meta-features within meta-learning systems~\citep{cavi09}. It is worth mentioning though that a `sampling-based landmark', seemingly related, is an alternative that involves training a non-simplified ML model on a subset of data~\citep{sope01}. In effect, partial learning curves of accuracy versus sample size for sampling-based landmarkers can be used to further characterise datasets, but these curves also contain additional information about the relative performance of candidate ML algorithms, allowing performance estimates for a full dataset~\citep{lebr05, lebr10}. Moreover, knowing how ML algorithms compare on one dataset can be used to guide ranking strategies on another similar dataset~\citep{lebr12}. This idea has been exploited several times, with one study working to meta-learn pairwise comparisons of utility between ML algorithms~\citep{glma13}. It is also related to collaborative-filtering approaches, where patterns of model performance are assumed to correlate between similar datasets~\citep{smmi14, sugu18a}.

Naturally, with so many meta-features available, it is a challenge to select a set with high discriminative power and low computational cost. This is such an issue that there are hints of a new research thread arising: meta-feature engineering~\citep{pale21}. Regardless, once a set of meta-features have been selected for a class of ML problems, with each problem generally corresponding to an inflow dataset, a standard practice for meta-learning is to construct a meta-model, i.e.~a function from dataset meta-features to a recommendation variable, usually trained by ML. The development and use of such meta-models are depicted in Fig.~\ref{Fig:PriorKnowledge}. Often, a k-nearest neighbour (kNN) approach is used to leverage similarity in simple fashion, sometimes called an instance-based meta-learner~\citep{cuhu16}, while deriving decision rules is another simple option for sparse metadata~\citep{ah92}. However, the meta-model can be as sophisticated as desired. An archetypal example of meta-model construction is depicted within a study that evaluated optimal SVM kernels for 112 classification problems and subsequently trained a decision tree as a meta-classifier to map vectors of dataset characteristics to those optimal kernels~\citep{alsm06}. Just like a standard ML model, the meta-model could then be passed a query, i.e.~a new dataset transformed into a set of meta-feature values, and would then calculate a response, i.e.~a suggested SVM kernel. In similarly model-specific contexts, recommenders have been trained for other SVM hyperparameters~\citep{gopr12} and for decision tree induction~\citep{gr14}.

\textbf{AutoML and Meta-learning.} In the modern HPO-centred context of ML model/algorithm selection, meta-learning is commonly used to suggest a good initial point to start searching from. This process is called warm-starting, whereby a systematic start-from-scratch optimisation procedure is boosted by an externally derived heuristic. It is represented in Fig.~\ref{Fig:PriorKnowledge} by the solution-recommending meta-knowledge signal being propagated on to the optimiser. This concept of warm-starting is exemplified by a study wherein GA-based HPO is applied to a meta-learned initial population of hyperparameter values~\citep{resh12}. It has also occasionally been incorporated within fully implemented AutoML systems, e.g.~Auto-sklearn~\citep{fesp15}. However, there have been many nonstandard variations on the theme, such as recommending a warm-start candidate by first minimising across a weighted combination of previously encountered HPO loss functions~\citep{wisc15a}.

Meta-learning has been leveraged in several other unique ways to support HPO. For instance, the Automated Data Scientist distinguishes itself from Auto-sklearn by seeking to directly predict optimal hyperparameters, as opposed to suggesting a warm-starting configuration~\citep{nich18}. Elsewhere, meta-knowledge has been used to recommend regions of configuration space that should not be explored~\citep{wisc15}. Then there are proposed upgrades to SMBOs in particular, where the underlying surrogate functions are constructed across all encountered datasets rather than per each new ML task~\citep{babr13}. The original proposal was later improved upon to deal with a scaling challenge, i.e.~the fact that evaluation metrics are not directly comparable across ML problems~\citep{yoma14}. Variants of these concepts have been developed in an adaptive manner, exploiting external knowledge for SMBO while a new solution space is unknown, before gradually shifting to an acquisition function based on local knowledge~\citep{wisc16}. Whether the meta-learned surrogate should be shared or ensembled across all ML tasks has also been debated~\citep{wisc17a}. Again, all of these variant approaches underscore how difficult it is for a proposed AutoML framework to encapsulate the full range of meta-learning approaches, considering that prior knowledge can feed informatively into operations at almost any level.

Because meta-models have no particular constraint on their recommendation variable, provided that every ML problem in meta-$\mathcal{Q}$-space maps to an independent response value in meta-$\mathcal{R}$-space, meta-learning has been used to suggest elements of ML solutions beyond standard sets of hyperparameters. For instance, recommender systems have been designed for classifiers, both standard~\citep{brso03} and multi-label~\citep{chro11}, as well as regression-based predictors~\citep{nupe17}. Admittedly, this is not a particularly radical extension in light of the CASH paradigm, which treats ML algorithms as hyperparameters. Recent work has even used opportunistic meta-knowledge to recommend both the size and content of a predictor set, which then serves as the search space for HPO-solver SMAC within Auto-WEKA~\citep{ngke21}. Regardless, as Section~\ref{Sec:Features} hinted, FE operators can also be among meta-learned ML components~\citep{khna16, kash16, bo17}, whether generative~\citep{nasa17} or selective~\citep{waso13, pale17}. Certainly, a recent empirical study espouses the use of a meta-model for preprocessor selection~\citep{scgi18}, and the PRESISTANT system is an example of a modern IDA centred around this aspect of AutoML~\citep{bi18, biab18}. Moreover, recommender systems can even propose entire ML pipelines; Meta-Miner is one such system employing a meta-model that maps dataset meta-features to KDD-workflow characteristics, the latter serving to encode information relating to pipeline structure~\citep{nghi14}. Even more recently, pipeline recommendations have been supported by a meta-model in the form of a low-rank surrogate decomposition~\citep{yafa20}.

The examples go on. Meta-knowledge can be employed at any level of an AutoML architecture, and entire ensembles have been recommended based on standard and landmarker meta-features~\citep{koce17}, with the topic of ensembling elaborated in Section~\ref{Sec:Ensemble}. Likewise, in the context of adaptation, further described in Section~\ref{Sec:Dynamic}, meta-learning has proved useful in selecting the active predictor within a heterogenous ensemble~\citep{im20}. Even the parameters of an optimiser have been subject to tuning processes; notably, for CASH-solvers, these sit at a higher level than parameters/hyperparameters for ML models/algorithms. While there are direct strategies for automating optimiser-engineering, such as using RL to learn an update policy while the optimisation is running~\citep{lima16}, there are also approaches based on standard meta-models~\citep{agka17} or trainable neural architectures, e.g.~Long Short-Term Memory (LSTM) networks~\citep{ande16}. These effectively transfer settings for quickly convergent optimisers across similar problems. Admittedly, searching parameters for parameter-searchers can become a recursive problem, and it is an open question as to how much of an AutoML system should be automatically tuned.

As a side note, while most meta-models work towards ML model/algorithm selection on the basis of predictive accuracy, there are other metrics for evaluating ML pipelines; see Section~\ref{Sec:Eval}. In particular, the issue of ML-algorithm runtime has been visited several times within the field of meta-learning. In one direction, runtimes have been folded into meta-features, so that learning curves~\citep{sope01} become loss-time curves~\citep{riab15}. A ranking scheme for ML algorithms based on these ideas proved competitive with Auto-WEKA, especially given small time budgets for which the AutoML system could not fully exploit its optimisation capabilities~\citep{caab17}. Alternatively, in the opposite direction, meta-knowledge has been deployed to predict runtime~\citep{huxu14}. This approach of algorithmic runtime prediction is actually of general interest to researchers of higher-level scheduling. Examples of related work include a focus on Bayesian network-structure learning (BNSL) problems, where attempts have been made to meta-learn the runtime of a BNSL-solver algorithm, represented in turn by the proxy metric of BNSL-problem complexity~\citep{maka14, maka17a}. Schedulers themselves are also open to being meta-learned, e.g.~as judged by the metric of `robustness' in the context of dynamic loop scheduling~\citep{srma13}. However, a particularly good example of crossover between research fields is the 2017 Open Algorithm Selection Challenge; the task here was to construct a scheduling system to try out solvers for an unseen problem, with meta-knowledge available in the form of how long various solvers ran on previous problems. Notably, one entry ended up using Auto-sklearn to construct the meta-model~\citep{maka17}.

\textbf{The Context of DNNs.} At this point, it is worth discussing transfer learning, given that boundaries between the topic and meta-learning can be inconsistent and ill-defined in the literature, if not outright nonexistent~\citep{paya10}. Generally, transfer learning does not learn a meta-model mapping of an ML problem to a recommendation variable; it simply copies the value of a recommendation variable, i.e.~knowledge, from one ML problem to another, e.g.~an entire cell of CNN layers. There is an implicit assumption that the source/target ML problems are similar, but there is no meta-model around to quantify this, hence why a transferred variable, e.g.~the CNN cell, must usually be tuned afterwards, while an ideal meta-model could theoretically recommend a value that immediately accounts for differences in setting. Some ML practitioners may discuss semantics further, perhaps constraining the scope of the knowledge that can be meta-learned or transferred, but we avoid this debate. From the perspective of architectural design, transferrable knowledge is already contained within the meta-model library in Fig.~\ref{Fig:PriorKnowledge}, offered to the ML-pipeline planner as solution advice in the form of meta-knowledge, provided that an ML task at hand shares an identical and usually high-level characteristic to one encountered previously. For example, if two tasks focus on text translation, it is circumstantially reasonable to copy and paste a neural model from one to the other, essentially as a warm start, even if the human language involved varies.

Transfer learning is commonly associated with deep learning in the current day and age, often referring to the practice of using a neural network that is pre-trained for one setting as part of a new DNN, thus transferring useful feature representations. One study aims to support this domain adaptation by meta-learning the relation between dataset dissimilarity, based on meta-features, and the amount of tuning required for a transferred DNN substructure~\citep{albu19}. Transfer learning has also been proposed to extend RNN-based NAS~\citep{zole17}, although the concept is somewhat different and involves augmenting the RNN controller with a task-embedding strategy; the similarity of ML tasks when converted into embedded vectors drives correlated child-network design sequences~\citep{woho18}. For a general AutoML system, it is also of interest whether feature representations, i.e.~FE pipeline segments, can be transferred from unlabelled and even arbitrary data to supervised learning tasks. This idea has been previously explored under the name of `self-taught learning'~\citep{raba07}.

Given the current popularity of DNNs, many novel meta-learning advances are often framed within such contexts. Model-agnostic meta-learning (MAML) was proposed in 2017, where a network is trained across a set of similar ML tasks so that, when used to initialise a gradient-based search for a new ML task, few iterations are required to optimise the model~\citep{fiab17}. In effect, good initial DNN weights are meta-learned for rapid few-shot learning. The utility of MAML has been demonstrated in several contexts, e.g.~efficiently learning an RL policy for robotic control~\citep{fiyu17}.

\textbf{Lofty Ideals and Challenging Practicalities.} Ultimately, meta-learning provides sound foundations for upgrading almost any element of a modern AutoML system, at least in theory. The motivation is that, assuming previous experience is relevant to an ML task at hand, it would seem beneficial to incorporate meta-knowledge in data-based predictive/exploratory systems. Unsurprisingly, there is evidential support for this view; one strand of research involving a `meta-mining' module~\citep{nghi14} and an IDA named eProPlan/eIDA~\citep{kise12} found that the `best' workflows suggested by KDD system RapidMiner~\citep{rikl01, hokl16}, specifically ML pipelines composed of frequently used operators, were significantly outperformed by workflows ranked and recommended via meta-models. In effect, automation based on meta-learning generally produces better ML solutions than those manually selected by human practitioners. Nonetheless, the extent of its effectiveness cannot be assumed. Leveraging meta-learning requires several unclear design choices, so much so that meta-model selection has itself been the focus of meta-learned recommendation~\citep{cuhu16}. Concern has also been raised that standard sets of meta-features cannot seem to discriminatively subdivide the space of ML tasks according to preprocessing pipelines they should employ~\citep{geth17}. Then there is the issue that training a recommender system well requires significant amounts of data and experiments from similar contexts, which may simply not be available in practice. It is not even clear whether the obvious solution to this, i.e.~drawing meta-knowledge from other domains, is particularly effective~\citep{alga18}. Nor is it conclusive just how much use can be extracted from opportunistically derived meta-knowledge~\citep{ngke21}. In essence, the promise of meta-learning is there, but, as with many AutoML research threads, more critical analysis is required to properly evaluate the benefit of employing these mechanisms.

\section{Ensembles and Bundled Pipelines}
\label{Sec:Ensemble}

Every predictive ML task, i.e.~where the response space $\mathcal{R}$ is previously defined by a user, must involve a predictor at some stage, namely a transformation that converts some feature representation of queries in $\mathcal{Q}$ over to responses in $\mathcal{R}$. It can be a weakness, however, for an ML model to rely on only one predictor. An AutoML system may not have access to a strong enough learner in its pool of ML components, or, alternatively, perhaps each predictor is too prone to overfitting for a particular dataset. Worse yet, for continuous streams of data, relying on one predictor is especially brittle. True AutonoML needs to be efficiently adaptive, easily maintained even when one ML model fails. Thus, for many reasons, managing a multiplicity of ML pipelines and their ensembles is an important thread of research.

\textbf{The Why and How of Combining Models.} Aggregating multiple models is a powerful technique for controlling error. One review into forecast combination provides a Laplace quote to suggest that this was understood in the early 1800s~\citep{cl89}. Thus, in the modern era, a lot of discussion about ensembles is framed within the context of bias-variance decomposition (BVD), which was introduced to the ML field by the 1990s~\citep{gebi92}. In essence, the theory states that the failure of an ML model to mimic a desired function arises from three error terms:
\begin{itemize}
    \item Bias -- The inability of the ML model to fit data due to its simplifying assumptions. High bias is typically the cause of underfitting and commonly describes weak learners.
    \item Variance -- The inability of the ML model to generalise, given how much its fit changes for different data samples. High variance is typically the cause of overfitting.
    \item Noise (or Bayes error) -- An irreducible error related to how poorly sampled data represents the desired function.
\end{itemize}
Typically, BVD theory is often brought up in the context of ML-model limitations and bias-variance trade-off, although there is ongoing debate about where exactly this trade-off is applicable~\citep{nemi18}.

In terms of ensemble methods, there are three common types that find frequent use:
\begin{itemize}
    \item Boosting -- An ensemble is constructed in sequential manner, with each new predictor training on re-weighted data; these weights prioritise data instances that were poorly modelled by previous predictors. The output of the ensemble is usually provided as a weighted average across predictors. This entire process often reduces bias, i.e.~underfitting.
    \item Bagging -- An ensemble is constructed in parallel manner, with each predictor training on data sampled with replacement. The output of the ensemble is usually provided as an average across predictors. This entire process, also known as bootstrap aggregating, often reduces variance, i.e.~overfitting.
    \item Stacking -- An ensemble is constructed in layered fashion, where the outputs of base predictors are appended to their inputs. The next layer of predictors trains upon the concatenation produced by the previous layer.
\end{itemize}
Many ensemble methods are often homogeneous, i.e.~based on one predictor, with boosting/bagging often incorporating weak learners. However, heterogeneous ensembles have been proven effective in their own right~\citep{co16, riho15, riho17}, perhaps even more so~\citep{gagi08}. Ultimately, the predictive power of an ensemble is bound very closely to the diversity of its constituent learners. Ensuring independence between predictors, or even complementarity via negative-correlation learning~\citep{eaga07}, can have strong performance impacts.

\textbf{AutoML and Ensemble Strategies.} Notably, ensembling is not a foreign concept to AutoML development, with many systems being promoted specifically because of these strategies. For example, employing mlrMBO~\citep{biri17} as a CASH-solver, one effort automates gradient boosting~\citep{thco18}. Given that many AutoML packages search through a pool of ML models/algorithms, the resulting `autoxgboost' implementation is relatively unique in focussing on a single learner. Elsewhere, stacking methodologies have seen considerable uptake, utilised in the BO-based Automatic Frankensteining framework~\citep{wisc17} and the GA-driven Autostacker~\citep{chwu18}. Most recently, AutoGluon-Tabular joined this list, suggesting somewhat counter-intuitively that ensembling techniques could be more important than solving CASH for achieving state-of-the-art accuracies~\citep{ermu20}. Indeed, a common theme within related publications is that ensemble-based systems frequently appear to outperform their non-ensembled peers, with the AutoGluon release even providing an ablation study to emphasise the benefits of multi-layer stacking. However, the ablation study does not seem to account for any possible advantages afforded by manually well-selected hyperparameters, which could provide other AutoML systems a commensurate runtime discount. Further analysis with good benchmark design will be required to settle the debate.

What is clear is that the automation of ensembled ML models cannot succeed without well-developed strategies for selection and combination. Coalescing results across multiple ML predictors generally appears more beneficial than arbitrating amongst them~\citep{chst93}, although this assumes that each predictor is still informative in some way regarding the inflow-data distribution. This is not necessarily true for concept drift; see Section~\ref{Sec:Dynamic}. It also remains an open question regarding what the most effective way to build an ensemble is, with one study suggesting that greedily attaching heterogeneous ML models on the basis of individual performances can outperform boosting and bagging~\citep{cani04}. Unsurprisingly, the ensemble selection problem can also be treated as a variant of CASH. Certain SMBOs have been updated to automatically produce ensembles~\citep{lala14, lega16}, with associated extensions into regression problems and upgrades for dynamic ensemble-sizing~\citep{rodr18}. Likewise, GA-based methods have also been explored for ensembles~\citep{garu06}. Genetic programming has been applied to automatically evolve forecast-combination structures for airline data~\citep{leri12} and has been further investigated alongside meta-learning strategies~\citep{koce17}. In fact, there exists an entire genetic-programming framework named GRAmmar-DrIven ENsemble SysTem (GRADIENT), which automatically creates combination trees of base-level predictors~\citep{tsga12}. A two-tier approach would later be proposed as an alternative to this, where high-level ensembles act as fuzzy combinations of low-level ensembles~\citep{tsga13}.

Two-tier ensembling, alongside its multi-layer generalisation, is worth further discussion. While a lot of research has traditionally been invested into flat weighted ensembles, certain studies, both theoretical~\citep{ruga02,ruga03} and empirical~\citep{ruga05}, have identified that the performance limits of a multi-classifier model can be stretched by suitably structuring the model into ensembles of ensembles. The other upside of working with such `deeper' ensembles is the flexibility of delegation that they possess, promoting ML solutions that leverage specialisation. The idea is that each learner within a collective, potentially an ensemble of lower-level learners in its own right, can be trained to specialise on its own non-overlapping selection of data, e.g.~by instance or feature~\citep{garu06}, according to the requests that an ensemble manager makes of a data distributor. Multi-level ensembling thus allows for a much more controlled and targeted synthesis of information. Of course, specialisation must still be managed carefully, with one word-sense disambiguation study showing that ensemble accuracy is very sensitive to the optimisation of individual experts~\citep{hohe02}. Moreover, designing a multi-layer MCPS is not trivial, relying on the right approaches for local learning and ensemble diversity~\citep{al18}. However, if done right, specialisation can be very powerful. This is evidenced by a series of publications exploring multi-layer ensemble models applied to airline revenue management~\citep{riga05, riga07, riga07a, riga09}, which, due to seasonal variations in demand, also investigated the dynamic evolution of these structures. Adaptation is further discussed in Section~\ref{Sec:Dynamic}. Similarly themed considerations likewise resulted in a generic multi-level predictor system for time series that managed to achieve significant performance in several contemporary forecasting competitions~\citep{ruga11}.

\textit{\textbf{Framework - Bundling Pipelines.}} In terms of the illustrative AutoML framework developed in this work, many of the elements discussed thus far already cater to ensembled approaches. Homogeneous ensembles can be constructed/aggregated internally within an ML component with little additional effort, as hinted at in Section~\ref{Sec:Basics}. Likewise, while heterogeneous methods cannot be represented on the ML-component level, a simple example of stacking has also been demonstrated, internal to an ML pipeline, within Fig.~\ref{Fig:PipelineExample}. However, we emphasise that the intent behind defining ML pipelines is to focus on enabling complex sequences of data transformations. There is still an implicit assumption that a one-pipeline ML solution must engage with the entire scope of a data source as part of the learning process, no matter the branching degree of its internal ML-component DAG. This is a severe limitation, as a diverse team of learners concentrating on different parts of a problem, i.e.~specialisation, may be much more effective than a single learner with a scattered attention. Now, it is arguable that heterogeneous boosting, with its sequential nature of having learners clean up the mistakes of other learners, can still be squeezed into ML-pipeline representation with sufficiently elegant inflow-data control, i.e.~constructing new candidate solutions as extensions of previous ones but training them on novel samplings. However, heterogeneous bagging relies on the distribution of dissimilar data samplings across various learners, all potentially working asynchronously.

\begin{figure}[htb!]
  \centering
  \includegraphics[width=0.95\linewidth]{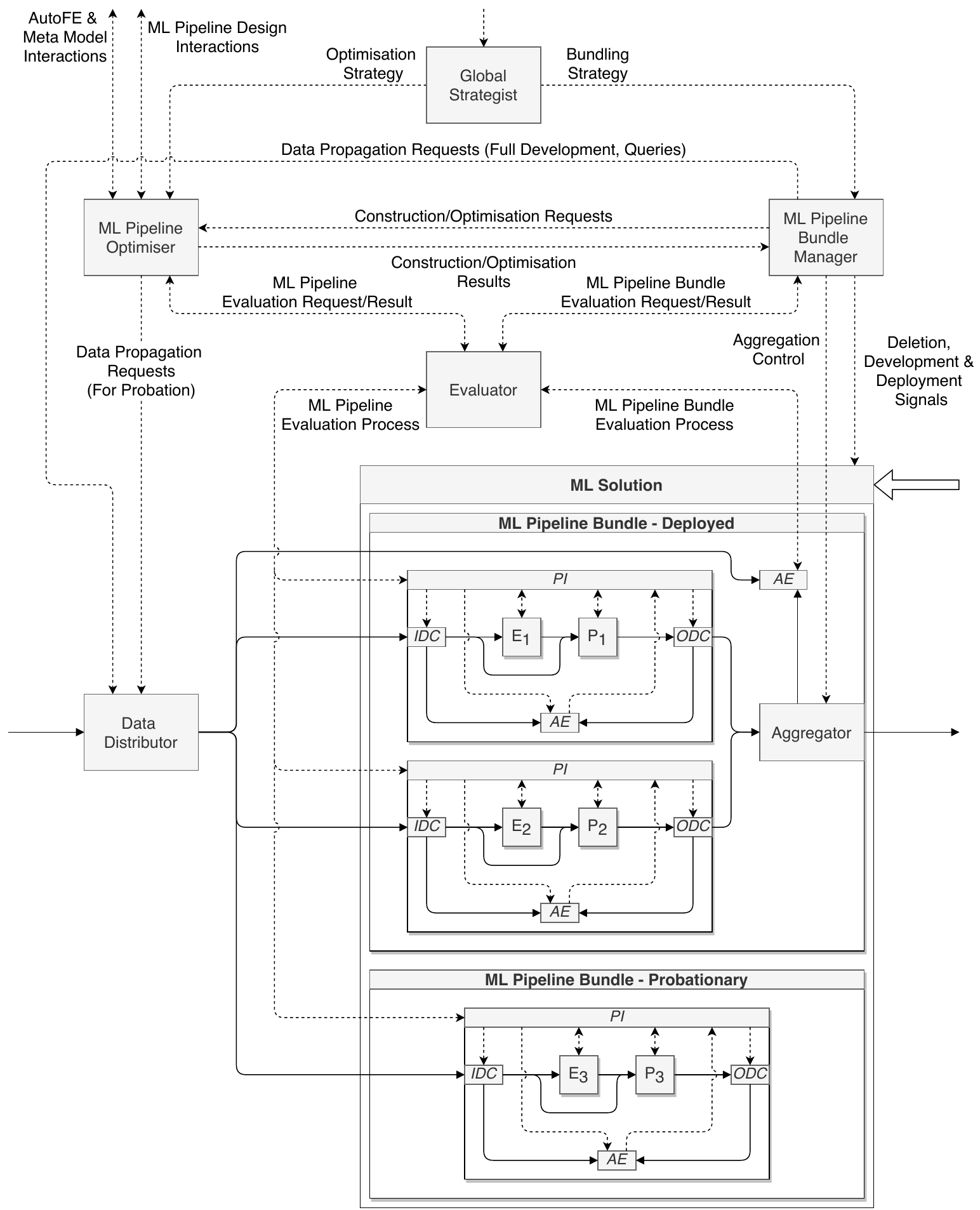}
  \caption{A schematic of a deployed two-pipeline ML bundle acting as an ML solution within an AutoML system, with a third ML pipeline kept in its own bundle for model development. Only the deployed bundle requires aggregation for an actionable output, along with a final accuracy evaluator (AE). ML components marked E\textsubscript{i} and P\textsubscript{j} represent feature-engineers and predictors, respectively. Other abbreviations denote a pipeline interface (PI), inflow-data controller (IDC), and outflow-data controller (ODC). Dashed arrows depict control and feedback signals. Solid arrows depict dataflow channels. Block arrows depict the transfer of ML components/pipelines.}
  \label{Fig:BundleControl}
\end{figure}

The upshot is that an ML solution needs a higher level of structure beyond an ML pipeline to fully support heterogeneity and complex data re-distributions. This is best served by the ability to manage multiple ML pipelines simultaneously and even ensemble them together. There is no consistent name for this specific parallelisation; here, we call the arrangement an ML-pipeline bundle, demonstrating in Fig.~\ref{Fig:BundleControl} how the structure could be utilised within an AutoML system. Within this illustration, there is a new module that oversees ML bundles, deciding when to add, remove, develop or deploy ML pipelines within the arrangement. Only one bundle is ever deployed though, and this one needs both an aggregator for individual outputs and an associated accuracy evaluator. The bundle manager controls the aggregator, e.g.~weighting the sum of outputs, and also has access to all evaluations that may assist its prescribed ensembling strategy. However, while the bundle manager does decide when a new ML pipeline must be added to the deployed ML bundle, possibly overwriting an underperformer, the pipeline optimiser is tasked with the actual construction and optimisation thereof, as expected.

Naturally, this division of responsibility does require additional modifications to the architecture. Data propagation requests are now split between the optimiser and bundle manager by whether an ML pipeline is under probation, i.e.~in the process of being optimised/validated, or under deployment, respectively. In static `one-and-done' ML processes, the entire ML solution will typically be wholly in a state of learning followed by wholly in a state of use, but the compartmentalisation depicted in Fig.~\ref{Fig:BundleControl} becomes important in dynamic `long-life' ML; see Section~\ref{Sec:Dynamic}. Furthermore, it is no longer sufficient to provide optimisation constraints upon interpreting/analysing a new ML task; strategies for both HPO and pipeline bundling must be determined. Within the illustrative schematic, these are prescribed by a global strategist. Further commentary on this is provided in Section~\ref{Sec:Discussion}.

\textbf{Combine with Caution.} Ultimately, choosing when to deploy an ensemble remains debatable. They can often improve the accuracy of a predictive ML solution markedly, but, as Section~\ref{Sec:Eval} discusses, the performance of an AutoML system cannot be judged on predictive error alone. Ensembles complicate ML models and become increasingly difficult to interpret, a challenge that regularly faces algorithm design and selection in the modern era~\citep{ru19}. One way to counter this is to train a simpler interpretable model to mimic the decisions of a more complex one, or use a more direct process of compression if available, and this concept of `knowledge distillation' is a research topic in its own right. It has been heavily investigated, for example, in the context of fuzzy neural networks~\citep{ga02, ga04, eaga11}, with similar approaches tried for decision trees~\citep{eaga12}. In relation to AutoML, distillation has also recently been explored with AutoGluon~\citep{famu20}. In the long run, the procedure may be worth implementing as one of several high-level processes to enact in the relative downtime of an autonomous system, i.e.~seeking condensed versions of deployed ML pipelines. Regardless, the capacity to build ML-pipeline bundles/ensembles remains a recommended upgrade for any AutoML system that seeks to pursue long-term maintenance/adaptation schemes.

\section{Persistence and Adaptation}
\label{Sec:Dynamic}

From here on in, all reviewed research is framed within the context of working towards AutonoML, not just AutoML. This distinction is drawn because, to date, the vast majority of AutoML research has focussed on automating the construction of ML models for fixed-term ML tasks. Broadly put, an AutoML system is engineered to be a powerful tool, turned off after use, whereas we contend that an AutonoML system should instead be viewed as an artificial brain, continuously developing while active. Unsurprisingly, many proposed upgrades for ML autonomy are biologically inspired~\citep{al16}. However, conceptually upgrading AutoML to become self-governing is highly ambitious and certainly not a direct route. Thus, for practical purposes, we propose a fundamental characteristic to define an AutonoML system: the capacity to persist and adapt~\citep{zlbi12}.

\textbf{Data Streams and Concept Drift.} Persistence is the idea that an AutonoML system should be capable of multi-level operations in the long term. Technically, keeping a trained ML model under deployment is a trivial form of persistence that all AutoML systems are capable of. However, support for persistent learning is much rarer. Few AutoML systems are designed for the streaming of inflow data, limiting ML operations to datasets of known finite size. This is understandable, as dealing with data from dynamic systems upstream of an ML process is very challenging, with the potential to incur a lot of technical debt~\citep{scho15}. On the other hand, the `Internet of Things' (IOT) is a growing driver of big data analytics~\citep{deha18}; the need to embed ML systems into long-life dynamic environments is likely to increase in the coming years.

At a low level, streamed data is nothing new to ML, with varying types of online-learning algorithms in existence~\citep{boga07}. Moreover, many tunable data transformations that traditionally work in one-off fashion have incremental variants, such as PCA~\citep{hama98}. For instance, one SMBO proposal highlights its use of Mondrian forests as an online alternative to random forests~\citep{kije16}. There is even at least one entire software environment dedicated to supporting ML for streamed data, e.g.~Massive Online Analysis (MOA)~\citep{biho10}. Thus, inevitably, researchers have begun studying the extension of AutoML processes into streaming contexts. This was the driving motivator for a 2018 NIPS competition into lifelong AutoML, as well as a recently released evaluatory framework named automl-streams~\citep{im20}.

Importantly, stream-ready AutonoML is not designed to inherently outperform standard AutoML in stable contexts, i.e.~static inflow-data distributions. For this kind of scenario, each incoming instance of data provides decreasing marginal information about the statistics of its source, obeying a law of diminishing returns. If the frequency of inflow data is sufficiently high, then buffering a representative sample becomes cheap enough, in terms of runtime, to annul the headstart advantage of online learning. In these stable contexts, stream-ready AutonoML instead proves its value for rapid-deployment operations, where a valid solution must be provided instantly and in the absence of historical data~\citep{kaga10a}, irrespective of quality. Section~\ref{Sec:Basics} supported this idea, proposing that, for a general implementation, data transformations should be initialised immediately within their ML-component shells. In fact, we suggest that, given a pool of ML models/algorithms, a solid AutonoML strategy should bias its initial selections to instance-based incremental learners before exploring more complex batch-based models, thus maintaining respectable predictors while allowing data to accumulate. Such strategies work best when ML-pipeline bundles are part of the design, detailed in Section~\ref{Sec:Ensemble}, because one or more ML pipelines can then be developed in parallel to one or more that are currently deployed; the foreground/background partitioning of these operations resembles the page-flip method used in computer graphics. Not unexpectedly, this solution-refinement process is a logical one for user-interactive systems that minimise response latency, e.g.~ones where users may wish to steer data exploration, and is employed by the Northstar data-science system within its CASH-solving module, Alpine Meadow~\citep{kr18, shzg19}.

Where the AutonoML paradigm becomes critically necessary is for ML tasks applied within dynamic contexts, specifically scenarios for which `concept drift' occurs~\citep{gazl14}. Technically, this refers to the desirable mapping between query-space $\mathcal{Q}$ and response-space $\mathcal{R}$ changing over time, e.g.~epidemiological notifications being updated to include asymptomatic cases. Notably, real concept drift is often conflated with virtual drift, whereby the data distribution that drives ML-model training changes, e.g.~notifications being assessed within a new city. Theoretically, virtual drift does not affect the ideal mapping. Nonetheless, given that an ML model, $M:\mathcal{Q} \to \mathcal{R}$, is already an imperfect approximation of the maximally desirable mapping, we use `concept drift' here loosely; in practice, both forms of drift require ML models to be re-tuned. Thus, if it is to persist usefully in these contexts, an AutonoML system should be able to identify any form of drift and respond appropriately.

\begin{figure}[htb!]
  \centering
  \includegraphics[width=0.975\linewidth]{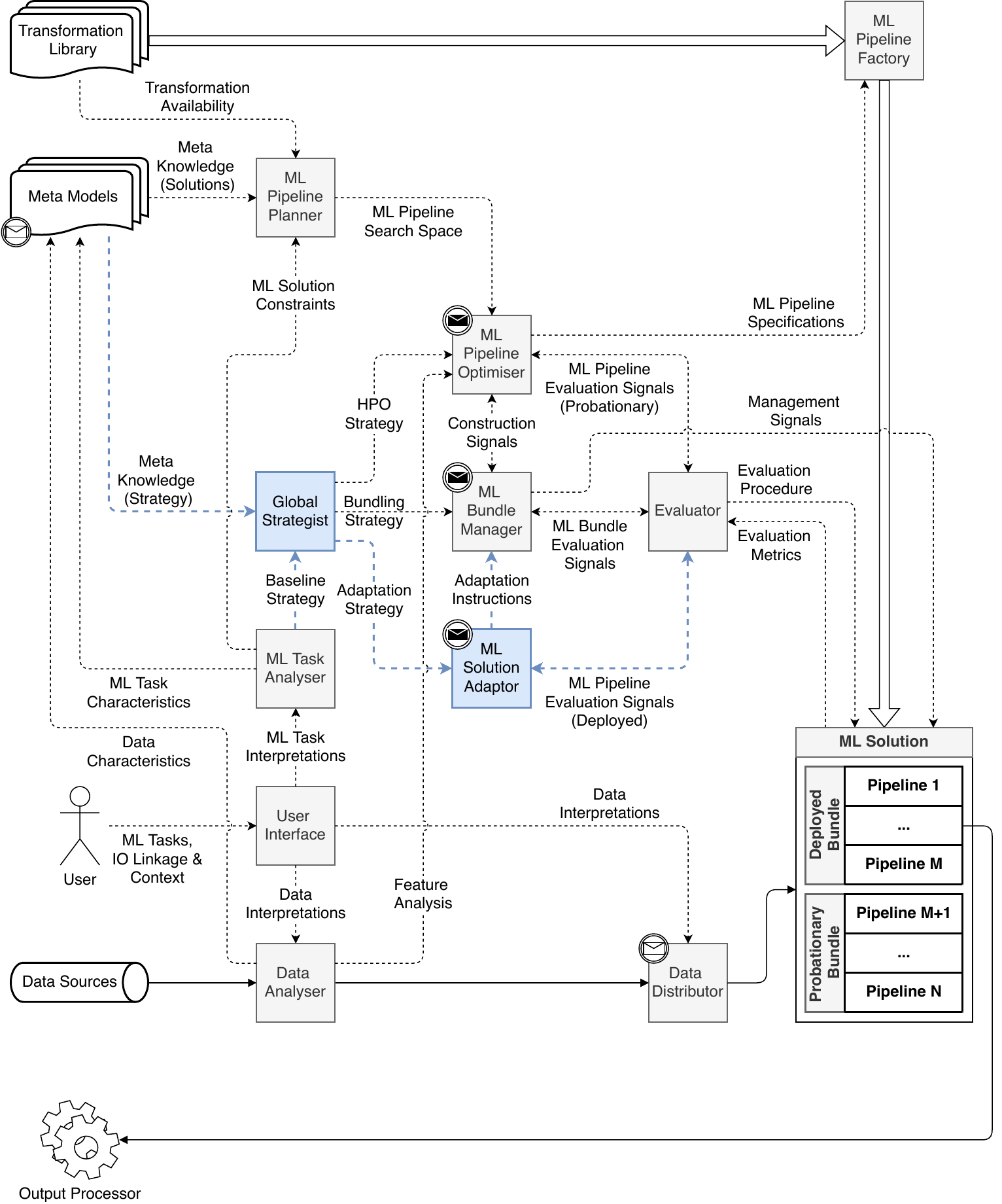}
  \caption{(Colour online) A high-level schematic of a dynamic AutonoML system suited to streamed data, where an ensemble of deployed pipelines is monitored for concept drift and reactive strategies are enacted. This architecture is an upgrade to that of Fig.~\ref{Fig:PriorKnowledge}, with blue highlights emphasising the main new additions related to adaptation. Dashed arrows depict control and feedback signals. Solid arrows depict dataflow channels. Block arrows depict the transfer of ML components/pipelines. Black/white envelope symbols represent data propagation requests to the data distributor and developmental feedback to the meta-models.}
  \label{Fig:DynamicLearning}
\end{figure}

\textit{\textbf{Framework - Continuous Learning.}} On a conceptual level, adaptive strategies can be integrated into an AutoML framework in the manner shown by Fig.~\ref{Fig:DynamicLearning}. This schematic builds upon the CASH-solving system supported by meta-learning, as depicted in Fig.~\ref{Fig:PriorKnowledge}, albeit with the additional inclusion of the ML-pipeline bundles introduced by Fig.~\ref{Fig:BundleControl}. Here, just as specialisation required an ML-bundle manager to oversee the ML-pipeline optimiser, a new adaptor module has its own elevated responsibility over the bundle manager. It primarily monitors the performance of individual deployed ML pipelines, polling the evaluator with some frequency over the lifetime of the ML solution, and, when the threshold for drift is passed, it instructs the bundle manager to deal with the failing pipeline accordingly. Based on the employed strategy, this may involve shifting extant ML pipelines between foreground deployment and background probation, not just the outright addition/subtraction of individual learners to/from memory.

As before, this new partitioning of responsibility does require updates to our illustrative architecture. Upon encountering a new ML problem, initial task analysis may suggest a baseline approach, but the global strategist is now responsible for prescribing individual tactics regarding HPO, pipeline ensembling, and adaptation. To reflect that any of these can be improved by prior experience, meta-models now feed their strategic knowledge into the global strategist for further dissemination. In turn, reciprocal meta-model development is extended across both the bundle manager and solution adaptor. As for dataflow control, the data distributor similarly receives signals from all three solution-developing modules to support their activities. Given the potential for undue clutter, Fig.~\ref{Fig:DynamicLearning} uses shorthand symbols to depict both meta-model feedback and data propagation requests.

\textbf{Strategies for Adaptation.} Returning to a survey of academic literature, the topic of adaptation has been a thread of ML research for decades. For instance, catastrophic interference has long been a challenge for updatable neural networks~\citep{mcco89}, in that previously learned knowledge is so easily erased by adjustments to the trained ML model. As a side note, one way to mitigate this effect is in fact to use the specialisation techniques introduced in Section~\ref{Sec:Ensemble}, so that adjustments in knowledge are well segregated; this was demonstrated by early adoption of a two-level ensemble within an adaptive monitoring scheme for a water distribution network~\citep{gaba99}. Elsewhere, biological inspirations for adaptive classifiers~\citep{fapa86} managed to spawn an entire sub-field dedicated to artificial immune systems. Amongst all these investigations, some studies have expressed a recurring principle, that adaptability is the capacity to cope with uncertainty~\citep{al16}. This view meshes well with the idea of fuzzy logic, where the rigidity of decision boundaries is softened, and so, unsurprisingly, the development of neuro-fuzzy theory has aligned well with researching ML solutions that are robust and easily adaptable. Indeed, fuzzy neural networks have previously been proposed as the basis for evolving intelligent systems~\citep{ka01, anka05}, and their online-learning algorithms continue to be improved~\citep{khch20}.

Of course, most ML algorithms do not have in-built mechanisms that differentiate between contexts. Instance-based incremental learners may evolve as rapidly as possible, closely tracking the receipt of new data, but even these algorithms are typically incapable of controlled non-catastrophic forgetting. One approach for designing exceptions to this rule is to include a domain-encoder in some manner. This is exemplified by a deep-learning architecture that, with some novel gradient-reversal tweaks, branches off a domain classifier alongside standard label-predicting layers~\citep{gale15}. Loosely expressed, the domain-classifier serves to absorb the influence of differing contexts during training, promoting domain-invariant features for label classification. While this domain classifier arguably internalises only one dimension of concept drift, the more recent Memory-based Parameter Adaptation (MbPA) method~\citep{spja18} works to embed and save previously encountered labelled data, potentially capturing multiple contexts. Parameters for an output network are adjusted based on how embeddings of new instances compare to those that have already been memorised.

From an idealised perspective, the paucity of specialised algorithms is largely irrelevant to AutonoML, which aims to be universally applicable; any strategist in such a framework should employ generic adaptation mechanisms that work irrespective of base learner. Dynamic Weighted Majority (DWM) is one such general method, bundling together an ensemble of experts~\citep{koma07}. If the ensemble errs, a new expert is added, training only on subsequent data that ideally encapsulates a drifted concept. If an expert errs, adhering too rigidly to outdated concepts, its prediction is given less weight within the ensemble. If confidence in an expert is sufficiently low, it is booted from the ensemble altogether. Another alternative is the Paired Learners (PL) method, which only requires two ML-pipelines~\citep{bama08}. Specifically, a stable learner trains on all the data it encounters after its initialisation, while a reactive learner trains on a moving window of data. If the reactive learner sufficiently outperforms the stable learner, the latter is reset with a new expert. Beyond this, there are many more approaches to adaptation that have been reviewed elsewhere, e.g.~in the context of soft sensors~\citep{kagr11}. Some recent works have argued that maintaining a static collection of stream-based heterogeneous learners, and only adapting the ensemble weighting, is already competitive with other methods in accuracy while being computationally cheaper~\citep{riho15, riho17}. However, more complex methods of note include the `incremental local-learning soft-sensing algorithm' (ILLSA)~\citep{kaga10} and its batch-based successor, the Simple Adaptive Batch Local Ensemble (SABLE) method~\citep{baga15}.

\textbf{Modularising Monitoring and Maintenance.} The principle behind SABLE is worth calling particular attention to, as the procedure is actually expressed as a composite of multiple adaptive mechanisms, e.g.~batch-based retraining or adding a new expert. The order of these processes is important in optimising predictive accuracy over time~\citep{baga17}. Recent experiments with SABLE have been combined with investigations into batched versions of DWM and PL, similarly decomposed into atomic adaptive mechanisms that could arguably be mixed and matched~\citep{baga18}. Given a strong drive towards integrated adaptive systems~\citep{gale05}, this customisation of strategies suits the ethos of AutonoML, although, as with the meta-models discussed in Section~\ref{Sec:Meta}, questions remain as to whether automating such a high-level selection process is tractable.

At the very least, a modular approach for adaptation is appealing; such a philosophy aligns with the design of a general adaptive MCPS architecture, published in 2009~\citep{kaga09}, that serves as partial inspiration for the illustrative one developed in this review. This plug-and-play framework is worth highlighting, as it describes a predictive system operating with constrained ML pipelines/components, employing a meta-learning module and even catering to both multi-objective evaluation and expert knowledge; see Sections~\ref{Sec:Eval} and~\ref{Sec:User}, respectively. Most pertinently, it is a system that employs a dynamic and hierarchical aggregation of learners, leveraging lessons from prior research~\citep{riga05, riga07, riga07a} that was detailed in Section~\ref{Sec:Ensemble}, and has been successfully instantiated several times in the context of soft sensors~\citep{kaga09a, kaga09b, kaga09c}. However, its publication preceded the surge of progress in HPO/CASH over the last decade, resulting in an architecture that is rudimentary in general purpose HPO, while being sophisticated in both automatic adaptation and the integration of multiple relevant ML mechanisms into a coherent framework. It is symbolic of the fact that, while we treat AutonoML as a successor to AutoML, research and development do not obey convenient orderings.

While devising strategies to maintain ML-model accuracy in shifting contexts is important, there is another key question to consider: when should an adaptive mechanism be triggered? A sustained concept drift, whether real or virtual, is different from an anomaly~\citep{chba09}; any detector should be capable of distinguishing and ignoring the latter. Naturally, there have historically been many approaches to change detection. The Page-Hinkley test is one such example from the 1950s, sequentially analysing shifts in the average of a Gaussian signal~\citep{pa54}. It finds frequent use in adaptive mechanisms, e.g.~triggering the pruning of regression-based decision rules based on their errors~\citep{alfe13}. On that note, many detection techniques are model-based, assessing concept drift by the deterioration of predictor accuracy. For instance, the Drift Detection Method (DDM) essentially alerts a context change if the error rate of an ML algorithm passes a threshold~\citep{game04}. The Early Drift Detection Method (EDDM) is more interested in the distance between errors, thus picking up on slower drifts more effectively~\citep{baca06}. The downside with such methods is that they can be computationally expensive, especially in resource-constrained hardware environments. This is particularly true for many non-incremental ML algorithms, which need to re-process historical data just to ingest a new batch of instances. Alternatively, reminiscent of the filter/wrapper duality of AutoFE, some drift detection methods operate solely on the data, e.g.~identifying outliers and shifts based on local densities~\citep{pola07}.

\textbf{Forecasting the Rise of AutonoML.} Despite all this research into the triggers and processes of adaptation, it is possible to question whether it is too speculative to declare AutonoML as the next step in ML automation. This is certainly open to debate, but the last decade has already teased synergies between the surge in AutoML developments and the theories of dynamic contexts. For instance, while ensemble-based strategies such as SABLE are usually homogeneous in terms of predictor type, HPO studies have considered the feasibility of changing hyperparameters over time~\citep{chtr13}. One proposal does this via the online Nelder-Mead algorithm, triggering the search when a Page-Hinkley test identifies concept drift~\citep{vega18}. Then there are efficiency-based investigations tailored for the idea of an MCPS. These include identifying whether segments of a pipeline can be adapted independently~\citep{zlga14} and, if so, how these adjustments must be propagated through the rest of an ML pipeline~\citep{sabu16b}. The interplay of meta-learning and adaptation also remains of interest~\citep{albu15}, with the former potentially used to recommend an expert~\citep{im20} or, at a higher level, an adaptation mechanism.

Crucially, it has been the last couple of years that have truly hinted at burgeoning interest in adaptive AutoML. While batch-based and cumulative strategies were evaluated with AutoWeka4MCPS for a chemical-processing case study in 2016~\citep{sabu16a}, at least three publications have recently been released to push beyond this body of work. One studies the extension of Auto-sklearn to data streams, complete with drift detection~\citep{maes19}, while another examines adaptive mechanisms more broadly, leveraging both Auto-sklearn and TPOT~\citep{im20}. A third works towards a more extensive multi-dataset benchmark, adding H2O AutoML and GAMA to the list, while also using EDDM to identify concept drift and trigger adaptation~\citep{ceva21}. Thus, it appears that the transition to AutonoML has already begun. Nonetheless, numerous challenges remain~\citep{zlbi12}. These include ensuring that adaptive systems are user-friendly, trustworthy, and capable of expert intervention. They must also scale well to big data that is realistic and messy, and, in the long run, work in general settings. In practice, adaptation must also be deployed sparingly, perhaps only when a cost-benefit analysis of an operation identifies a net positive~\citep{zlbu15}. Intelligent evaluation is key to optimising all aspects of AutonoML.

\section{Definitions of Model Quality}
\label{Sec:Eval}

Recalling Section~\ref{Sec:Basics}, the purpose of ML is to learn a model from data that approximates some maximally desirable mapping, $\mathcal{Q} \to \mathcal{R}$. In certain cases, i.e.~unsupervised learning and data-exploratory processes, this ground-truth function can seem arbitrarily subjective and vague; if two data scientists are scouring data for `interesting' information, would they agree on their criteria for interest? Would they even know what they are looking for until they find it? Without meta-learning the brain chemistry of a user, or at least the expression of their preferences, an AutoML system cannot estimate the quality of an exploratory ML model until it serves a user-defined purpose elsewhere. Nonetheless, for predictive tasks, e.g.~those employing supervised learning, the desired $\mathcal{Q} \to \mathcal{R}$ mapping is much more objective and immediately discernible from inflow data, e.g.~hinted at by way of class label. Thus, unsurprisingly, there is a tendency across ML research to focus on the concept of `accuracy'. This is understandable, as minimising some metric for the dissimilarity between model and ground-truth function is a universal goal across all ML tasks. However, the automation of ML is ultimately in service of human requirement, and AutonoML, as a field, cannot ignore the many ways by which model quality can be judged.

\textbf{The Rigours and Systems of Evaluation.} It is first worth stressing that the topic of accuracy evaluation itself is in no way settled. As mentioned previously in discussion of the no-free-lunch theorems~\citep{wo96, woma97}, the performance of an ML model and ML algorithm is intrinsically tied with the data environments that they operate upon. This can make it difficult to trust general claims about, for instance, NAS methodologies that limit their evaluations to ImageNet. Would those same approaches be as effective on a radically different dataset? What even determines their sensitivity in performance? While meta-learning often tries to intuit these variations of model quality in automated ways, sometimes even opportunistically~\citep{ngke21}, there have been several attempts to analyse the space of all ML problems more rigorously. For example, one work scatters datasets from the well-known UCI repository across a construction of such a space, subsequently examining the shape of algorithmic performance `footprints', i.e.~where an ML model has empirically performed well~\citep{muvi17}. The research even considers how to generate new artificial datasets to flesh out unsampled parts within this topology. Of course, any attempt to define such a problem-instance space faces the usual issues around which meta-features to use. Nonetheless, the motivation of this work is important; one cannot truly assess the general performance of low-level ML models/algorithms or high-level AutoML mechanisms without benchmarks that are sufficiently `diverse' -- whatever that may truly mean -- in both datasets and, more abstractly, problem contexts. This fundamental challenge should be kept in mind when considering almost any metric.

As for discussing metrics specifically, it is convenient that, from a conceptual perspective, broadening the scope of model performance does not require any major changes in the illustrative architecture developed thus far, i.e.~Fig.~\ref{Fig:DynamicLearning}. While ML components and ML pipelines have all incorporated an accuracy evaluator, the high-level evaluator module has been left intentionally abstract from the beginning, linking up with the interfaces of both. There is no reason why the ML-pipeline bundle cannot be fleshed out as a proper wrapper either; Fig.~\ref{Fig:BundleControl} only avoided this for the sake of simplicity. Thus, assuming low-level interfaces have easy access to other model characteristics beyond internally evaluated accuracy, e.g.~runtime or memory footprint, higher-level modules are able to optimise the solution according to varied objectives. Admittedly, direct assessment in the Fig.~\ref{Fig:DynamicLearning} schematic does remain limited to the ML solution, with high-level strategies being only indirectly evaluable. This is likely fine in practice, although, in theory, perhaps high-level strategies should in fact be directly evaluated and optimised; see Section~\ref{Sec:Discussion}.

\textbf{Metrics Beyond Accuracy.} Notably, when the algorithm-selection problem was first codified in the 1970s, various ways of evaluating models were proposed, including metrics based on complexity and robustness~\citep{ri76}. The vast majority of HPO methods and CASH-solvers over the last decade have disregarded these concepts, although not without cause; error metrics are, on the whole, easier to standardise and calculate across different types of ML model. Even so, the idea that model quality could be based on alternative metrics has persisted throughout the years, with the fusion of multiple criteria likewise studied at least as early as in the 1970s. For instance, despite describing not-for-profit programmes, one publication discussing the `efficiency' of decision making units by combined characteristics~\citep{chco78} is often cited as a forerunner for multi-objective evaluation in KDD systems. Certainly, by the 1990s and the formalisation of KDD, the list of proposed metrics by which to judge an ML-model was broadening, including attributes such as novelty and interpretability, and it remained an open research question as to if and how these could be measured and combined in weighted sum~\citep{nasc97}.

In recent years, with the advent of HPO methods, sections of the ML and ML-adjacent communities have asserted the need for multi-objective optimisation~\citep{hone17}. A thesis on multi-layer MCPS frameworks explores this topic in depth~\citep{al18}, noting that common approaches may attempt to linearly combine objectives, hierarchically optimise them, or, in the absence of clear priorities, seek solutions that are Pareto-optimal, i.e.~where the improvement of one criterion necessarily damages another. In practical terms, the mlrMBO package is one implementation among SMBOs that caters to multiple criteria~\citep{howa15, biri17}, while, for evolutionary methods, multi-objective genetic algorithms have been well established for many years~\citep{depr02}.

Returning to evaluation criteria, typically the first extension researchers explore beyond accuracy is runtime; there is a limit to user patience for any practical ML task, beyond which increased accuracy is not appreciated. Argued to have been largely ignored until the early 2000s~\citep{brso03}, the gradual inclusion of runtime into model-evaluation schemes~\citep{snla11, snla12, riab15, caab17, ab17} may have mirrored the development of the user-focussed KDD paradigm and the maturation of HPO, e.g.~the shift of AutoML from lengthy academic experiments to public-use toolkits. Of course, observing the development time of an ML pipeline and its ML components is trivial, and meta-learning techniques have been proposed to make runtime estimates based on the transfer of previously accumulated experience~\citep{huxu14, maka17}. As a side note, runtime here refers to model development costs; latency as part of model deployment is discussed in Section~\ref{Sec:Resource}.

Beyond accuracy and runtime, many performance metrics have been proposed for ML solutions, although, to date, consolidative surveys are largely absent. In part, this is because many proposals are only applicable to restrictive subsets of use cases, or have at least been coupled tightly to particular contexts. For instance, FE transformations can be judged based on associated errors and tuning time, but optimising feature-selection percentage, i.e.~preferencing fewer features~\citep{waso13, pale17}, is specific to AutoFE pipeline segments. As another example, an automated learning system for semi-supervised learning (Auto-SSL) uses SVM-reminiscent `large margin separation' as a criterion for HPO, where inflow data is mostly unlabelled~\citep{liwa19}. Furthermore, as adaptation research into streaming data and changing contexts continues, it is likely that evolving AutonoML systems will require new descriptors for the quality of ML solutions, e.g.~metrics relating to temporal stability and robustness.

Crucially, we have primarily surveyed technical performance metrics in this section. These would be what developers traditionally focus upon, and they do not reflect end-user experience in its entirety~\citep{xats20}. Although a deep review into AutoML and mainstream engagement is better approached as a standalone work, user interfacing is further discussed in Section~\ref{Sec:User}. Even then, it must be acknowledged that ML models are often embedded into applications managed by businesses, and these products frequently operate in social contexts replete with human biases. Thus, there are other ways to judge a model that are often more challenging to directly quantify. For instance, a focus on profit will lead to questions around whether the benefit of running an ML model will counter the monetary costs of deployment and maintenance~\citep{zlbu15}. Socially conscious businesses may similarly assess the carbon footprint of, for example, running a NAS algorithm~\citep{pago21}. These evaluations are complicated by being derived from delayed processes; they must often be estimated if an AutoML mechanism is to be guided by them in a timely manner. As for issues of bias and fairness, which underlie whether ML models can be trusted by society~\citep{ru19, drwe20}, a discussion around these topics can be extensive. It is certainly possible to set up automatic mechanisms that both discover and prevent discrimination~\citep{habo16}, provided that stakeholders agree how fairness should be defined on a technical level. The problem is that there are many possible definitions for fairness, and, in many scenarios, they are orthogonal or, worse yet, contradictory~\citep{veru18}.

\textbf{Mastering Multiple Objectives.} For now, research strands into ML performance evaluation remain arguably disorganised, at least from the perspective of informing high-level control mechanisms for an AutonoML system. Typical ML benchmarks focus on minimising both loss functions and processing times, which do not necessarily encapsulate the entirety of human requirement. In fact, in seeming response to the age of deep learning and extensively layered DNNs, it is a tenable view that model complexity has recently become a major marker for solution quality, or, more precisely, the lack thereof. Specifically, while the MCPS paradigm has enabled arbitrarily complicated ML solutions to be built, discussed in Section~\ref{Sec:MCPS}, there is an implicit understanding in the ML community that no ML model should be unjustifiably complex. Efforts towards a general meta-learning architecture have espoused this view~\citep{jagr08, jagr11}, and the TPOT package demonstrates a practical Pareto-based method of controlling ML-pipeline size~\citep{olba16}. From an ideological perspective, complexity is eschewed by intensifying social desires for explainable/interpretable ML models~\citep{gegi19, ru19}. Indeed, while the power of ensembling for maximising predictive accuracy is well evidenced, as discussed in Section~\ref{Sec:Ensemble}, the process of knowledge distillation into simpler models is partially driven by the opaque nature of these ensembles~\citep{famu20}. However, there are also pragmatic reasons for maintaining simple models, especially when computational resources are in short supply.

\section{Resource Management}
\label{Sec:Resource}

All sections up to this point have implicitly discussed AutoML and its extension to dynamic contexts, AutonoML, as fundamental developments in data science. The emphasis has been on how novel strategies and theoretical advances in various topics, e.g.~optimisation and knowledge transfer, have enabled or upgraded the automation of workflow operations for an ML task, with special focus on designing and maintaining high-quality ML pipelines within a diversity of contexts. However, while the conceptual promise of AutonoML has the potential to revolutionise how society engages with ML on a day-to-day basis, the present reality suggests that practical application is still far behind theory. There may be a proliferation of AutoML packages as of the early 2020s, both open-source and commercial, but it is arguable that many lack the application stability to support general uptake. For instance, several benchmarking studies~\citep{baal18, zohu21, ermu20} have, between them, encountered process failures for the following: Auto-WEKA, Hyperopt-sklearn~\citep{kobe14, kobe19}, Auto-sklearn, auto\_ml, TPOT, H2O AutoML, Auto-Tuned Models (ATM)~\citep{swdr17}, AutoGluon-Tabular, Google Cloud Platform (GCP) Tables, and Sagemaker AutoPilot. This list is virtually guaranteed to not be exhaustive. Thus, it is clear that the challenges remaining are not simply theoretical; AutonoML is a problem of engineering. Managing the automation of ML when subject to resource-constrained hardware is a research thread of its own.

\begin{figure}[htb!]
  \centering
  \includegraphics[width=0.925\linewidth]{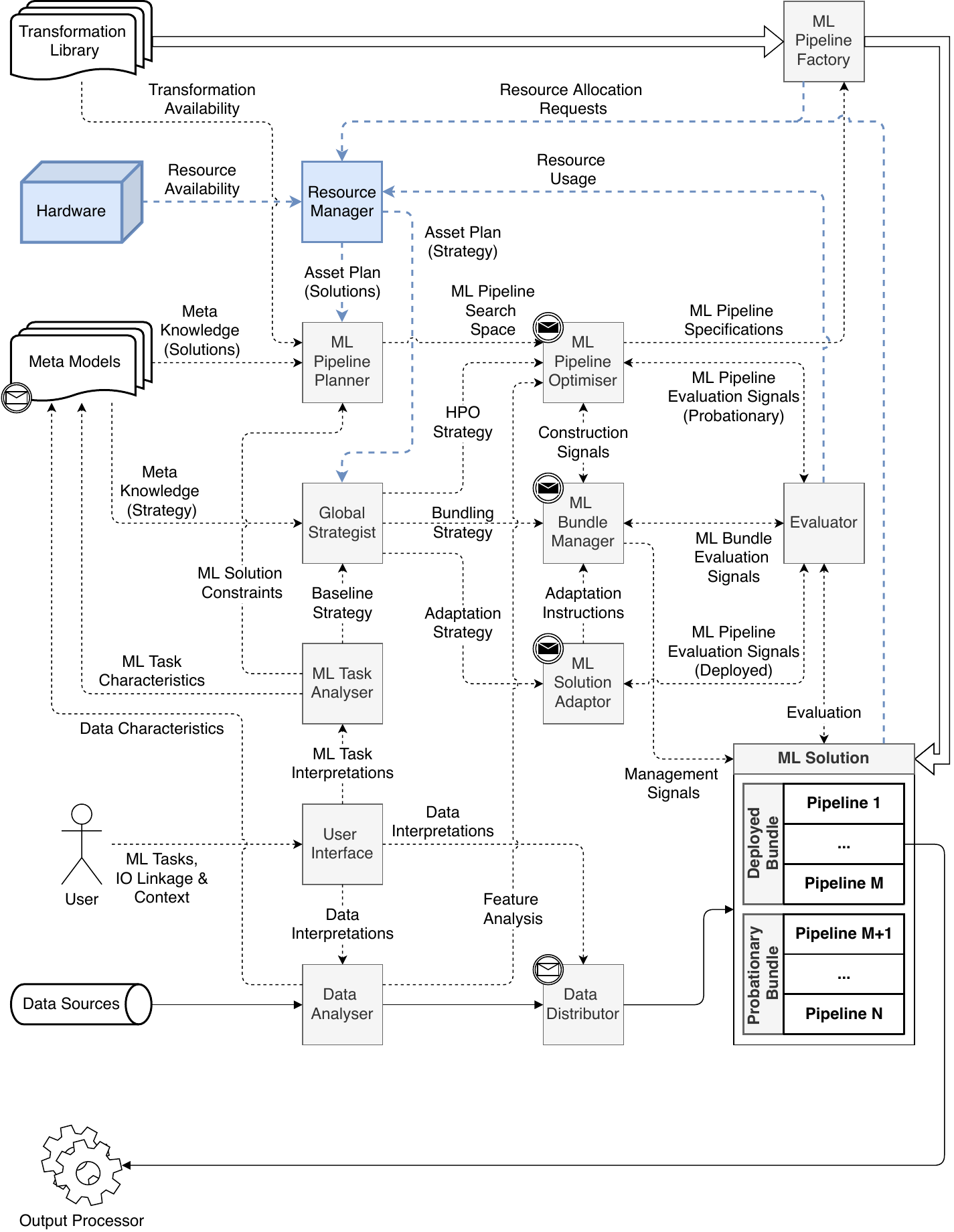}
  \caption{(Colour online) A high-level schematic of a resource-aware AutonoML system, with a manager responsible for asset allocation and strategy. This architecture is an upgrade to that of Fig.~\ref{Fig:DynamicLearning}, with blue highlights emphasising the main new additions related to resource control. Dashed arrows depict control and feedback signals. Solid arrows depict dataflow channels. Block arrows depict the transfer of ML components/pipelines. Black/white envelope symbols represent data propagation requests to the data distributor and developmental feedback to the meta-models.}
  \label{Fig:Resources}
\end{figure}

\textit{\textbf{Framework - Exploiting Hardware Availability.}} Naturally, it is challenging to proclaim anything too specific on this topic; hardware particulars, including layerings and intercommunications, vary greatly from IOT devices to large-scale servers. However, broadly speaking, an AutonoML system cannot achieve optimal performance without direct awareness of and influence on its hardware environment. An abstraction of these features is illustrated by the use of a resource manager in Fig.~\ref{Fig:Resources}, a schematic that extends the adaptive AutonoML shown in Fig.~\ref{Fig:DynamicLearning}. In terms of routine operations, this management module is tasked with connecting requests for resources, e.g.~memory allocations for data transformations, to the hardware that is available. However, it also represents the opportunity to guide the search for ML solutions, assessing the limitations/benefits of available hardware and evaluating how resources are used in the process of ML activities. With this knowledge, a well-designed resource manager can both constrain solution space directly, e.g.~restrict ML-pipeline complexity for low-memory settings, and tailor high-level strategies, e.g.~trial DNN candidates only when GPUs are free. This representation also implies that strategies, e.g.~adaptive mechanisms, can themselves be adapted over the course of an ML task in response to changing resource circumstances. Overall, most recent developments in resource-conscious AutoML can be projected in some way onto this conceptual framework.

\textbf{The Pressures of Time.} Returning to a survey of related research, runtime appears to be the primary focus of efficiency-based studies in AutoML. While Section~\ref{Sec:Eval} discussed runtime as a metric for grading ML models/algorithms, resource-management research is more interested in how the system as a whole should function around scarce processing time. In particular, `anytime algorithms' are procedures designed specifically around this concern, providing valid solutions to a problem if interrupted. Incremental ML algorithms are generally most suitable for anytime-compliance, whereas batch-based learning has more to lose if cancelled; this idea is related to the discussion of rapid deployment in Section~\ref{Sec:Dynamic}. Importantly, while low-level training processes are good candidates for anytime mechanisms, a robust AutonoML system should be safely interruptible at all levels. As an example, one proposed meta-learning framework acquires its meta-model for ML algorithms offline, but then, upon being applied to a dataset, is able to evaluate an ordered list of recommendations until interrupted, returning the best solution found thus far~\citep{glma13}.

Of course, while resilience against interruption is a necessity, efficient operation within time constraints is the ideal. For instance, research into fusing BO with MAB approaches was partially driven by time-budget considerations~\citep{hosh14, fakl17}, which is perhaps why some consider BO-HB to be a state-of-the-art CASH-solver~\citep{yash20}. Even so, working within a budget may still be more practically challenging than expected, as, curiously, a recent study found circumstances in which popular AutoML packages, TPOT and H2O AutoML, did not even adhere to user-provided time limits~\citep{ermu20}. Other high-level systems have also been the focus of budget-conscious experimentation, with, for instance, the economic principle of return on investment (ROI) used to assess adaptive mechanisms~\citep{zlbu15}. The ROI in this study retrospectively pits gains in ML-model accuracy against processing time, additionally including several other costs, e.g.~random access memory (RAM) and cloud-service communication costs.

\textbf{AutoML and Distributed Computing.} Excluding rare exception, the earliest years of modern AutoML research appear to avoid serious discussion about hardware, with efficiency-based developments targeting algorithms in a general manner. However, given the nature of DNNs, the popularisation of NAS has forced developers to be conscious about interfacing with hardware. For instance, the Auto-Keras package automatically limits DNN size based on GPU memory limits~\citep{jiso19}. Furthermore, there has been growing interest in designing AutoML systems and packages for clusters of machines, despite the typical challenges of developing distributed ML~\citep{dupa17}, e.g.~designing/using an appropriate `programming model' beyond MapReduce~\citep{ka16}. For some portions of the ML community, this interest appears to be motivated by the needs of time-critical big-data biomedical fields~\citep{lu16a}, as is exemplified by the application of distributed HPO to a lung-tissue classification problem~\citep{scmu16}.

Consequently, recent years have witnessed AutoML frameworks being developed for distributed big-data settings, e.g.~KeystoneML~\citep{sp16, spve17} and ATM~\citep{swdr17}. Notably, an early forerunner of this movement is MLbase~\citep{krta13}, which is supported by a MAB-based CASH-solver called `Training supported Predictive Analytic Queries' (TuPAQ)~\citep{spta15}. Beyond using a cluster of machines to evaluate ML models in parallel, TuPAQ is also designed for data-parallel scenarios in which a dataset is too large to fit on one machine, thus forcing model updates to rely on the fusion of partial statistics derived from data subsets that are processed on individual worker machines. Accordingly, many of the design choices behind TuPAQ are geared towards optimising against this engineering challenge, such as a mechanism to discard unpromising models before they are fully trained. Such stopping rules are not uncommon, and they are employed, for example, by another cluster-based hyperparameter optimiser named Sherpa~\citep{heco18}. 

The recent trend of companies providing AutoML as a cloud-based service only complicates matters further. In these settings, an AutoML system must run multiple ML tasks across a pool of processors in asynchronous fashion. This extension of standard AutoML to multiple devices and multiple tenants (MDMT) requires novel strategies for resource coordination~\citep{yuka19}. The cost of data transfers between hardware can become significant in such distributed settings, meaning that hyperparameter configurations must be allocated intelligently between workers when solving CASH. Task schedulers based on runtime estimates have been proposed to support this allocation~\citep{taas18}. Beyond model-parallel and data-parallel approaches, certain frameworks like PipeDream have also explored whether breaking apart an ML pipeline, technically a DNN, may be more efficient, where individual machines are dedicated to the training of individual ML pipeline segments~\citep{hana18}.

A notable extension of distributed ML that has gained attention in the last few years is the concept of federated learning, where data samples located on local machines cannot be transferred elsewhere. Because of this restriction, the data may not be independent and identically distributed either. Any associated research has been attributed to a wave of `privacy-aware' ML~\citep{gegi19}, on account of data scientists only having access to locally trained models and not their underlying data. Recently, the federated-learning community has highlighted a growing need for HPO and other AutoML processes in this context, albeit stressing that communication and computing efficacy for mobile devices cannot be overwhelmed by the traditional objective of model accuracy~\citep{kamc19}. Consequently, the year 2020 has witnessed the application of NAS to federated learning, with, for example, the FedNAS framework, which deploys one NAS-solver per local machine and then coordinates them via a central server~\citep{hean20}. Alternatively, the DecNAS framework targets low-resource mobile devices, noting the FedNAS assumption that each local machine is GPU-equipped. In the absence of hardware resources, DecNAS uses subgroups of devices only to train/test candidate models, while the actual searcher is situated at a higher level, i.e.~a central cloud~\citep{xuzh20}.

\textbf{The Spectrum of Deployment Environments.} There are further engineering challenges to consider that are often overlooked in theoretical investigations of AutoML. For instance, most benchmarks do not consider inference latency, i.e.~the time it takes to convert a query to response. Traditionally a negligible concern in academic experiments over the last few decades, inference latency has gradually become an important issue due to the complexity of DNN models, so much so that some NAS algorithms specifically take into account the environment a constructed network is expected to be deployed within~\citep{cazh19}. Furthermore, this latency can become significant enough in MDMT settings and industrial applications to warrant an automated control system of its own, e.g.~InferLine~\citep{crse18}, which is responsible for query distribution and provisioning ML bundles. Occasionally, this execution lag can be further diminished by designing and incorporating bespoke hardware, e.g.~a Tensor Processing Unit (TPU)~\citep{joyo17} or a Neural Processing Unit (NPU)~\citep{foov18}. In such a case, an ideal AutonoML system should be capable of leveraging optional hardware when appropriate.

On the other end of the spectrum, hardware restrictions can be necessarily severe, such as when designing ML solutions in IOT settings~\citep{deha18}. In this context, an AutonoML system must decide whether to centralise its ML model and treat integrated circuits with embedded Wi-Fi as pure sensors, or whether some ML components should be installed on these limited-memory microchips. Research into IOT-based AutoML is presently sparse, but this only serves to suggest that many applications of interest remain for AutonoML research.

\section{User Interactivity}
\label{Sec:User}

How should artificial intelligence (AI) be designed to interact with humans? This recurring question is inherently profound, but, within the AutonoML context, it is a pragmatic one. Traditionally, a core principle behind AutoML is to reduce manual involvement in all ML processes. Indeed, a long-term goal of AutonoML is to allow a user to specify a task of interest, hook up relevant data sources, and then relax; the onus is on the system to generate and maintain a high-quality ML solution. However, an ideal framework should also facilitate user participation whenever desired and wherever possible. In fact, a lack of user-friendly control options among contemporary packages may be adversely affecting AutoML uptake altogether~\citep{lema19}.

\textbf{Human in the Loop.} Notably, in some scenarios, human interaction is intrinsic to an ML task; these are generally called human-in-the-loop (HITL). Although this topic is broad, it typically involves a user engaging with an ML system, either by validating query responses, e.g.~confirming the classification of an email as spam, or by providing data that best serves the learning process, e.g.~deciphering scanned text via `Completely Automated Public Turing tests to tell Computers and Humans Apart' (CAPTCHAs)~\citep{ahma08}. The process of an ML system intelligently requesting additional information from an oracle/teacher is called active learning, and it shares faint similarities with the persistent learning and adaptive mechanisms discussed in Section~\ref{Sec:Dynamic}. Frequently applied to NLP, studies of active learning have often considered how to optimise the feedback loop between human and AI, e.g.~by seeking the most impactful corrections to a model~\citep{cukr06} or by accelerating feature extraction between requests~\citep{anan16}. Some efforts have even explored using RL to outright learn active-learning strategies, i.e.~tactics for deciding which unlabelled instances of data should form the basis of annotation requests~\citep{fali17}. Essentially, while reliance on human intelligence in HITL settings may seem incompatible with the concept of autonomy, progress towards effectively leveraging this interaction lies firmly within the AutonoML sphere.

More broadly, the design of AutoML systems regularly involves questions about where to link in human control and expertise. Some ML platforms target the ML pipeline, with, for instance, the hybrid Flock system embedding a crowd of users as an FE component~\citep{chbe15}. This crowd is asked to nominate and label a feature space for inflow data associated with an ML task, thus encoding attributes that may be challenging for a machine to extract, e.g.~the attitude of a person within an image. In theory, though, every control element of an AutonoML system could be influenced by a human expert, as was considered within an adaptive architecture proposed in 2009~\citep{kaga09}. Baking this flexibility into an implementation is an engineering challenge, but it has been suggested that AutoML uptake would be served well by at least designing a few varying levels of autonomy, e.g.~user-driven, cruise-control, and autopilot~\citep{lema19}. One recent study has even attempted to systematise which user-driven interactions should be supported; this pool of basic actions was partially drawn from a decomposition of published scientific analyses~\citep{giho19}.

\begin{figure}[htb!]
  \centering
  \includegraphics[width=0.925\linewidth]{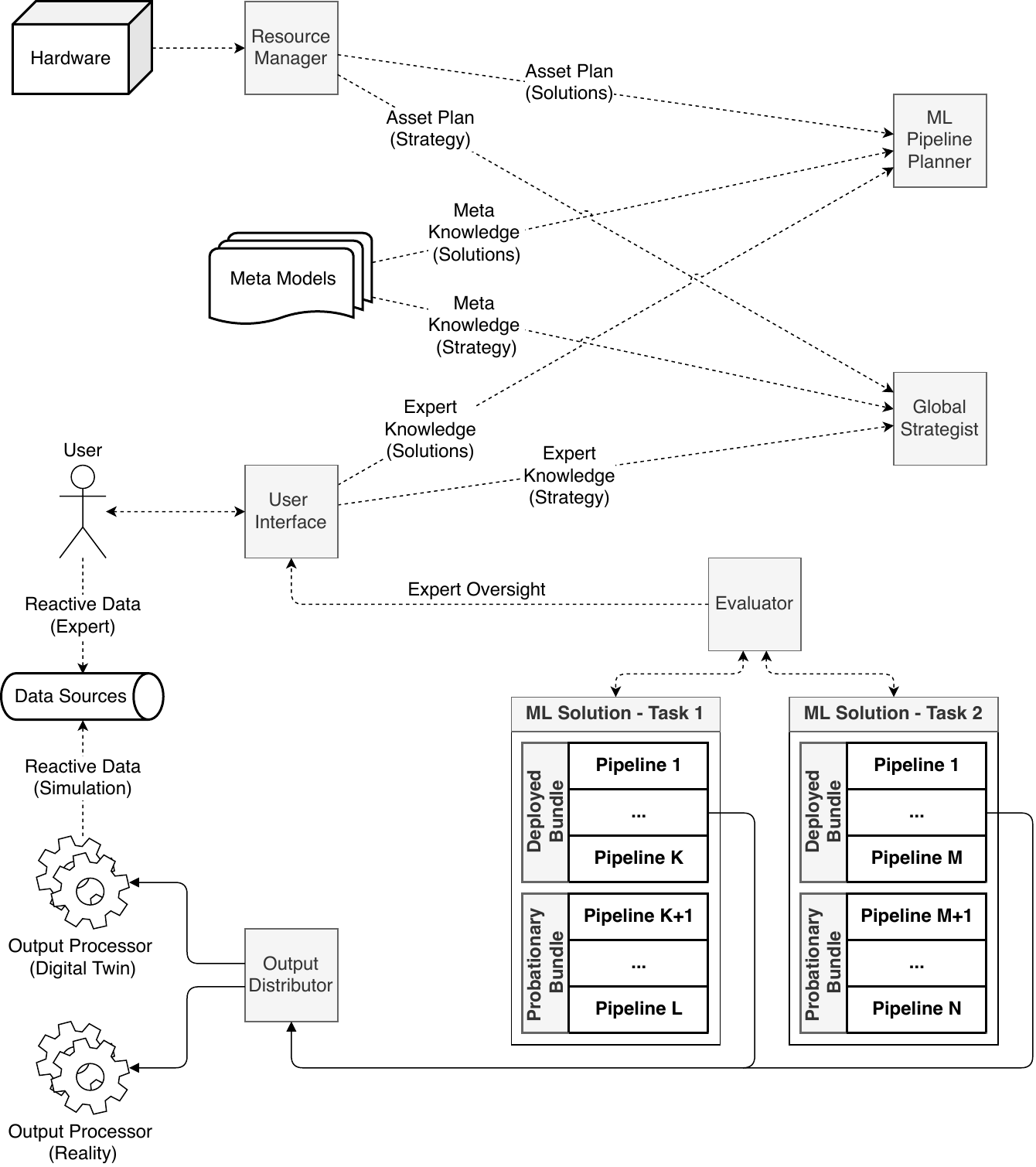}
  \caption{A high-level schematic of expert-knowledge and digital-twinning mechanisms in an AutonoML system. These upgrades should be applied in the architectural context of Fig.~\ref{Fig:Resources}. Dashed arrows depict control and feedback signals. Solid arrows depict dataflow channels.}
  \label{Fig:Expert}
\end{figure}

\textit{\textbf{Framework - Exploiting Expert Knowledge.}} Given the expanding complexity of the illustrative framework developed in this monograph, we avoid presenting another cumulative depiction of AutonoML. However, the salient elements of advanced user interfacing are shown in Fig.~\ref{Fig:Expert}. This representation can easily be integrated into resource-aware AutonoML, portrayed in Fig.~\ref{Fig:Resources}. Crucially, the diagram emphasises the analogies between expert knowledge and meta-knowledge, suggesting that there is no conceptual difference in how prior knowledge is used to tune regular/hyper/system parameters, whether accumulated manually or automatically. In this review, we have essentially partitioned preparatory high-level ML operations into two objectives: determining a solution search space and determining how to search throughout that space. Within this representation, the analogies are thus reflected by informative signals being directed to both the ML-pipeline planner and global strategist. This is a consistent design choice; Section~\ref{Sec:Resource} suggested that recommendations based on the practicalities of resource availability should be likewise distributed. Beyond all that, Fig.~\ref{Fig:Expert} also portrays active-learning processes, with model evaluation exposed to the user. User response may be to tweak either the model or how a new model is found, but data sources may also be supplemented or adjusted as a reaction.

\textbf{Supporting Democratised AutonoML.} Regardless of design, if this growing focus on interactivity appears to contradict the principles of automation, it is explained by the following: concurrently to the field of AutoML surging in prominence, the last decade has also witnessed a parallel movement towards AI accountability~\citep{gegi19, ru19}. A popular idea is that an AutonoML system can be treated as a black box, but, if required, its processes should also be completely observable and, ideally, comprehensible~\citep{xats20}. For instance, the Automatic Statistician project merges the philosophies of AutoML and explainable AI by aiming to generate human-readable reports from both data and models~\citep{stsm19}. Moreover, the idea of interpretability is appealing even during the development of an ML solution; a recent clinician-supported study suggests that predictive performance is more effectively fine-tuned by a user if that user assesses/augments the internal human-understandable rules of a model rather than individual instances of inflow data~\citep{gefr20}.

Of course, whether as a black box or with its machinery exposed, facilitating ease-of-use for an AutonoML system, thereby encouraging lay usage, brings new risks into focus. Specifically, there are intensifying calls to safeguard users from poor or even dangerous conclusions arising from data-science illiteracy~\citep{boko19}. Northstar, an AutoML system motivated by user interactivity and democratisation~\citep{shzg19}, exemplifies consideration of these topics by proposing a module to Quantify the Uncertainty in Data Exploration (QUDE), which acts as a sophisticated error-detection mechanism for user interactions~\citep{kr18}. Ultimately, it is evident that, if AutonoML is to be successfully integrated into society, the design of a UI will need just as much attention as any other subsystem.

\section{Towards General Applicability}
\label{Sec:AGI}

Over the course of many sections, numerous practical challenges have been raised with respect to the automation of ML. Most of these stem from limitations in the resources that are presently available. Without some technological paradigm shift, e.g.~quantum computing~\citep{biwi17}, searching for a data-transformation pipeline that suits some ML task, particularly a predictive one, will continue to be typically measured in hours, if not by longer timescales. Nonetheless, with research continuing to explore and develop this space, evaluating AutoML packages by the quality of their solutions, i.e.~any metric discussed in Section~\ref{Sec:Eval}, may eventually become less compelling. In the long run, the utility of AutonoML may be determined by another factor: generalisability.

\textbf{The Many Modes of Data.} By and large, most AutoML packages have not been developed with flexible data inflow in mind. With many data transformations expecting inputs of a consistent size and shape, even ingesting arbitrary images was considered a significant deep-learning advance, with spatial pyramid pooling promoted as an effective alternative to substandard cropping/warping processes~\citep{hezh14}. Thus, unsurprisingly, extensive manipulations of heterogeneous data do not appear particularly common in ML. By the time disparate data passes through the preprocessing segments of an ML-pipeline, it has usually been imputed/concatenated into a uniform vector space for the remaining FE and predictive segments. The approach is similar even when dealing with time series; several chemical-processing studies exemplify this by synchronising and combining data streams from various sensors into vectorised snapshots of an industrial environment~\citep{kaga09, sa17}. In fairness, this can occasionally be a design choice, as these same studies do actually involve frameworks capable of constructing ML-pipeline bundles and partitioning inflow data in sophisticated ways~\citep{ruga11}, theoretically enabling the delegation of individual ML experts to distinct sensors.

Regardless of implementation, it is clear that being able to generalise inflow data extends the power of AutonoML, especially in IOT settings~\citep{deha18}. There is a growing appreciation of this fact, with, for example, one recently proposed AutoML system promoting its relatively novel ability to ingest different kinds of data, including tables, text, images, and audio~\citep{giho19}. In fact, learning from multimodal data has recently become of significant interest, with, for instance, AutoGluon-Tabular being merged with transformer architectures to process free-form text in data tables~\citep{shmu21}; the system has been ranked first and second in several Kaggle and MachineHack competitions. Moreover, the field of AutoDL, which is partially motivated by the need to handle diverse forms of data, has likewise begun to examine multimodal NAS~\citep{alhu20, yucu20}. This is why, from the very beginning in Section~\ref{Sec:Basics}, the approach in this review towards conceptualising AutonoML has emphasised the need to treat inflow data as generally as possible.

\textbf{Accommodating New Problem Contexts.} In the long term, the ideal AutonoML system is one that does not just cater to generic data inflow; it is an application that is flexible in the types of tasks it can handle and, as Fig.~\ref{Fig:Expert} suggests, ideally able to operate several at the same time. This means extending beyond the supervised learning that a vast majority of present AutoML packages focus on. This also means considering how to contextualise novel ML operations beyond model selection as they arise. Section~\ref{Sec:Ensemble} has already mentioned knowledge distillation as a potential task worthy of its own module. Another example is shown in Fig.~\ref{Fig:Expert}, which attempts to represent one manifestation of the concept recently termed `digital twinning', heralded at least as early as in the 1990s under the speculative label of `mirror worlds'~\citep{ge91}, whereby high-fidelity simulations of physical systems act as proxies for reality. In this schematic depiction, ML solutions can be fed consecutive data that is simulated as a reaction to previous model outputs, where, for various reasons, the acquisition of such data may be infeasible in the real world.

Now, granted, there have been substantial efforts to automate solution search within less typical ML contexts, particularly recently. For instance, one AutoML framework specifically targets semi-supervised learning~\citep{liwa19}. Another one applies the principles of NAS to the trending topic of generative adversarial networks (GANs)~\citep{goch19}, with related work attempting to automatically distil a GAN~\citep{fuch20}. However, most of these efforts currently appear to be proof-of-principle, constructed in bespoke manner for the novel contexts, as opposed to extending existing capabilities. Of course, there is a valid discussion to be had whether attempting to develop a one-size-fits-all system is even the right functional approach. Some may argue that a better path to progress is in facilitating the quick development of bespoke AutoML systems, just as an AutoML system is ideally intended to streamline the production of bespoke ML models. Perhaps AutoML development libraries would be more pragmatically useful, in similar fashion to how PyTorch and Tensorflow support DNN experimentation. Indeed, there have already been various attempts to abstract out aspects of AutoML mechanisms in the hope of supporting standardised implementations, e.g.~PyGlove using symbolic programming to decouple search space, search strategy and candidate model for NAS~\citep{pedo21}. Nonetheless, it must be acknowledged that the learning machine AutoML is attempting to support, i.e.~the brain of a data scientist, is itself capable of grappling with a diversity of learning tasks, regardless of human efficiency. Mimicking this capability with a singular system would be quite an achievement.

\textbf{Hints for a Future of Flexibility.} Ultimately, a general-purpose AutonoML implementation may not be as blue-sky as one might expect. In the years prior to the optimisation community pushing AutoML, KDD researchers dwelled significantly upon ontologies of ML tasks while devising recommender systems. In essence, mappings were studied between ML tasks and the structures of ML pipelines that best suited those problems~\citep{fero13}. While such notions of hard constraints have trickled into the AutoML sphere, the technical achievements behind CASH have yet to be fully explored for the wide-ranging KDD paradigm. It is conceivable that, with good design of a specification language, an AutonoML system could accommodate a variety of ML task types, each of which would be translated into ML-pipeline structural constraints by a KDD approach and then fleshed out by a CASH solver. Naturally, various tasks may require unique ML components to be imported from distinct ML libraries, but this is not an obstacle; it is possible to mesh codebases even for disparate programming languages. One recent approach to this wrapped up environment-specific ML components as web services~\citep{mowe18b, mowe18}. Furthermore, with robust mechanisms for persistent learning and the long-term accumulation of meta-knowledge, there is potential for long-life AutonoML to accelerate model development within unfamiliar contexts, leveraging associations between the meta-features of an ML task and its optimal solution.

Admittedly, the major caveat to the AutonoML paradigm as it currently stands is that it can only evolve via combinatorial exploration within pre-coded spaces of ML components, adaptive mechanisms, ML-task types, and so on. This means that the extended generality of a framework like AutoCompete~\citep{thkr15}, designed to tackle arbitrary machine learning competitions, remains subject to the hard limit of ML scenarios that the developers either have experienced or can imagine. Nonetheless, and this is no foreign idea to the ML community~\citep{li18}, future developments in the field of AutoML/AutonoML may well have a bearing on the broader quest for artificial general intelligence (AGI).
\section{Discussion}
\label{Sec:Discussion}

Each section of this review has honed in on a particular facet of ML, primarily high-level, surveying the benefits/challenges that its automation may produce/encounter in the pursuit of maximally desirable data-driven models. However, unlike most contemporary reviews of AutoML, there has been an intended emphasis on integration woven throughout this work, with a concurrently designed conceptual illustration demonstrating one way in which to blend these ML mechanisms together. Unsurprisingly, there is more to discuss regarding the big picture of integration.

\textbf{AutonoML and the Challenges of Integration.} The overwhelming majority of existing large-scale AutoML frameworks, if not all, are designed prescriptively from above. In essence, developers select phenomena they wish to see in the system, e.g.~HPO and meta-learning, and then build around these, often maintaining some sort of modularity. Inevitably, this approach necessitates design decisions, either implicitly or explicitly, that prompt a debate about encapsulation and intercommunication. As an example, one may question whether Fig.~\ref{Fig:ModelSelection} actually serves as an authoritative blueprint for a CASH-solving AutoML. For instance, is the separation between ML-solution planner and optimiser overkill when an MCPS is not involved? Should not the evaluator be responsible for some data propagation requests, rather than leaving them all to the optimiser? Perhaps instantiating an IDC per ML component is too large of a footprint, and the onus of intra-component data distribution should also be on the high-level distributor? Ultimately, pragmatics will determine implementational specifics, and a Unified Modeling Language (UML) diagram of a resulting AutoML codebase may look dramatically different from the networked abstractions in Fig.~\ref{Fig:ModelSelection}.

On the other hand, each schematic in this work has been presented as a broad conceptual depiction of ML/AutoML/AutonoML operations; the demonstrative power of a schematic depends simply on whether it captures the salient features of the associated mechanisms. By this metric, Fig.~\ref{Fig:ModelSelection} is not expected to be controversial in its depiction of solving CASH, and any legitimate tweaks to the diagram must maintain the same expression of that process. That raises the question: design choices aside, do all the diagrams in this review definitively reflect the appropriate application of an AutoML mechanism? In truth, and this is why the large-scale integration of AutonoML is deserving of research attention, probably not. The optimal interplay of processes is unknown.

To elaborate on this, it is first worth noting that a fundamental criticism of AutoML is that automation merely shifts the level at which human involvement is required~\citep{li18}. Base-level ML algorithms select the parameters of a model. CASH-solvers select the parameters of an ML algorithm, i.e.~hyperparameters. So then, what selects the parameters of a CASH-solver, i.e.~hyper-hyperparameters? Certainly, each time this search is elevated by a level, the time required to solve an ML problem balloons out; it may presently be infeasible to optimise the high-level optimisers. However, relying on user-defined system parameters could face the same negative criticism earned by the use of default model parameters~\citep{baca17}. There may be an eventual need to discuss telescoping optimisations and thresholds for diminishing utility.

Crucially, matters are complicated when multiple methodologies interact. The illustrative framework developed in this monograph has been grounded in the debatable assumption that HPO, specialisation and adaptation are best enacted on the same level above the ML model. Indeed, there is a simple logic to the equal treatment of associated strategies that is depicted in Fig.~\ref{Fig:DynamicLearning}. Yet, perhaps the parameters of specialisation/adaptation strategies should be optimised. Perhaps HPO/adaptation strategies should operate differently across a bundle of ML pipelines, i.e.~be specialised. Alternatively, perhaps HPO/bundle strategies should be adapted for concept drift. At the moment, beyond a disinclination for complex architecture, there is no systematic way to determine when any of these arrangements are worthwhile or not, especially when benchmarking AutoML remains in its infancy. If, in fact, an AutonoML framework does turn out to involve sophisticated multi-level networks of mechanisms in order to perform optimally, the field will begin to overlap with the area of complex systems, insinuating possibilities regarding emergent behaviour and the process of learning. We defer further speculation on the topic.

Importantly, it must be acknowledged that, irrespective of engineering, certain trade-offs are likely to remain constants in the design of an ML/AutoML/AutonoML system. These include:
\begin{itemize}
    \item Speed versus accuracy -- Low runtime and low loss are archetypal competing objectives, the contest of which, given the computationally expensive loop of model/algorithm probation for CASH, will remain a challenging disincentive for layperson uptake. Having said that, the incredible improvement in computational power, memory and availability of large amounts of data means that the computationally hungry AutoML research emerging in recent years has become tractable, and the speed with which ML models of equivalent accuracy can be trained has also been dramatically increased over recent years. Naturally, there is an ongoing need for research in this space to take advantage of inevitably improving computational power. 
    \item Stability versus plasticity -- Highlighted by the extremes of entrenchment and catastrophic forgetting~\citep{mcco89}, any adaptive system faces this dilemma when seeking an equilibrium between learning new concepts and maintaining extant knowledge. While this classical dilemma related to learning systems is unlikely to go away, it has been ameliorated by ongoing progress in the thorough and continuous validation of both acquired knowledge and the relevance of deployed ML models.
    \item Control versus security -- Given an AutoML package, broadening the scope of user access and influence is known to improve model performance for a human expert. On the other hand, the opposing result holds for an amateur~\citep{yaxi17}. This emphasises just how important the AutoML/AutonoML research facet of user interactivity is, as discussed in Section~\ref{Sec:User}, along with, more broadly, the debate around HITL challenges for AI systems in general. These have not yet been addressed in any significant way.
    \item Generality versus simplicity -- The capacity for one system to tackle any ML task is appealing. However, without solid design principles to manage model complexity and feature bloat, the associated codebase can suffer from being too heavyweight for bespoke contexts. These competing attributes are also related to the tension between the accuracy afforded by complexity and the requirement for models to be both transparent and interpretable. In essence, there is a need for progress in both directions. For instance, flexible multi-objective optimisation approaches are expected to extend AutoML/AutonoML beyond a communally prevalent focus on predictive accuracy, enhanced in recent years by additional considerations of model/algorithmic complexity and both runtime and memory usage, but these approaches must also be robust and easily applicable.
\end{itemize}
Unsurprisingly, finding the right balance for each of the above trade-offs is a significant research endeavour in its own right.

As a final comment on integration, it may end up that the `prescribed-from-above' approach proves entirely inappropriate for AutonoML systems of the future. Admittedly, there is an appeal to designing an AutonoML framework in a modular manner, insofar as the operation of all desirable mechanisms is clearly demarcated and interpretable, but, as DNNs have exemplified, sometimes optimal performance is hidden within obtuse arrangements of granular base elements. If ML algorithms can be designed from scratch by evolutionary means~\citep{reli20}, then perhaps an AutonoML architecture itself can and should be grown from first principles, where desired mechanisms arise organically from the pursuit of well-designed objectives. For now, though, this remains a theoretical curiosity, with pragmatism likely to continue pushing the prevalent top-down approach of AutoML design. Biomimetics may be an appealing direction for AutonoML~\citep{al16}, but fashioning a tool remains easier than growing a brain.

\textbf{AutoML and a Brief Assessment of Impact.} Regardless of how the ML community decides to approach the architectural integration, it seems clear that, in the long term, AutonoML has the potential to revolutionise the way that ML is practiced, with significant implications for everyday decision-making across diverse facets of the human experience. That said, history has also witnessed at least two AI winters, primarily on the back of unmet expectations, and there is debate whether the limitations of deep learning are severe enough to herald another in the near future~\citep{sc19}. Thus, while crucial advances in HPO, meta-learning, NAS, adaptation, and so on, are arguably academically impressive, the evolution of the field is not served by overhype.

Therefore, a standard question to ask is this: has the automation of ML had an impact on society? This cannot be answered of AutonoML, as, with the exception of early pioneering work on this subject in 2009--2010~\citep{kaga09, kaga10}, ML-model search in dynamic contexts has only begun garnering broader attention within the last year or so~\citep{maes19, im20, ceva21}. However, CASH was formalised in 2013~\citep{thhu13}, meaning that the modern notion of AutoML has now had a while to diffuse through the data-science community and the mainstream. Certainly, there has been a recent proliferation of both GitHub repositories and vendor-related websites touting AutoML services. Books are being written to promote AutoML to the general public, e.g.~with tutorials for Auto-Keras~\citep{soji22}. Some frameworks have even successfully crossed over from academia into business within this timeframe, with the evolution of Northstar~\citep{kr18, shzg19} into the `einblick' company as an example of commercialisation.

Nonetheless, as of the early 2020s, the answer to the question on impact remains inconclusive. Notably, while it is expected that any publication proposing a new AutoML implementation or subsystem will promote its successes wherever evident, there are likewise strong indications that automation does not, well, automatically improve ML outcomes. This has been the case from the beginning, with a 2013 graduate-school project attempting to apply Auto-WEKA to a Kaggle competition, assuming the perspective of a layperson, which eventually reached the following conclusion: ``The automated procedure is better than the average human performance, but we do not achieve performance near the top of the ladder, and we cannot even compete with the single default model trained on all the data.''~\citep{an13}

Granted, AutoML has come a long way, and modern benchmarks, e.g.~run by the AutoGluon team~\citep{ermu20}, show that automated systems can dominate data scientists in Kaggle competitions. Yet mixed sentiments persist, reoccurring within the few independent evaluations of AutoML packages that have been made, and results can often be spun one way or another. An example is provided by a 2020 benchmark~\citep{habl20}, which uses the performance on several OpenML datasets~\citep{vari14, bica17} to assess the following four systems: Auto-sklearn, TPOT, H2O AutoML, and AutoGluon. The study positively claims that at least one package performs equal to or better than humans for seven out of twelve datasets. However, upon excluding three ML tasks where humans and almost all tested packages are capable of perfect accuracy, the results can also be interpreted with equal validity to imply that AutoML systems fail to outperform humans in five, almost six, out of nine contexts.

Conclusions are likewise sparse when solely comparing AutoML packages against each other rather than against human practitioners. One analysis, based on four popular options, finds Auto-sklearn dominant for classification datasets and TPOT best for regression problems, but it cautions that results have extremely high variances~\citep{baal18}. Another benchmarking study of seven open-source systems is even more circumspect, noting variations in performance but concluding that there is no perfect tool that dominates all others on a plurality of ML tasks~\citep{trwa19}. Proprietary systems are even more difficult to evaluate, however a 2017 investigation into ML-service providers is likewise unable to identify an outlier~\citep{yaxi17}. Returning to open-source packages, another comparison of four major systems across thirty-nine classification tasks~\citep{gile19} found that, after four hours of operation, all could be outperformed by a random forest for certain problems. The packages also struggled with datasets involving high dimensionality or numerous classes.

Accordingly, scepticism is warranted with any declaration of a state-of-the-art system. There are numerous deficiencies in academic culture, with, for instance, publish-or-perish pressures occasionally leading to shortcuts being taken when making research claims, the generality of which is sometimes unsupported~\citep{gegi19}. Computer science is also a field that moves fast, and publications are often released without adequate information that would allow results to be easily reproduced~\citep{pe11}.

In fairness, though, AutoML appears most burdened by the fact that there does not yet seem to be a consensus about where its greatest advantages lie. Of the limited independent benchmarks that exist, most adhere to traditional evaluations of predictive accuracy after set amounts of processing time. However, given that a distribution of models engineered by a large crowd of humans on an ML task is likely to have a strong maximal accuracy, perhaps the role of AutoML is one of efficiency instead, i.e.~achieving a reasonable result in a short time. If so, such an assessment would mirror conclusions drawn about easy-to-use off-the-shelf forecasting algorithms and their utility for non-experts~\citep{lega08}. Passing judgement is further complicated by the fact that an AutoML/AutonoML system effectively involves the interplay of many high-level concepts, some that may not be shared with other packages. Depending on implementation, these subsystems and constraints may also be difficult if not impossible to turn off or modify, respectively. One may then ask the following: on what basis should a one-component system with a warm-start function be compared against an MCPS without meta-learning? In short, rigorously comprehensive benchmarking is substantially less developed than the state of AutoML engineering, despite the existence of strong supporting infrastructure~\citep{vari14, bica17}, making it challenging to verify that any theoretical claims propagate into real-world problems. Even ablation studies, appropriate for examining such complex systems, are rare enough to stand out when they arise~\citep{ermu20}.

The upshot is that, if AutoML is to truly become a disruptive technology, there is more work to be done beyond just the technical front. Indeed, a recent study of the online OpenML platform shows that less than 2\% of its users and workflows have adopted associated techniques~\citep{lema19}. Some hints of why this is the case are present in interviews with data scientists, which reveal that, while time-savings resulting from automation are appreciated, there remain several negative sentiments around AutoML~\citep{wawe19}. Concerns include the loss of technical depth for operators, a focus on performance rather than insight, the inability to infer domain knowledge, and issues of trust. The authors of the study concluded that, in general, data scientists are open to their work being augmented, not automated, by AI. Likewise, a deeper study into trust has suggested that users may be more open to accepting AutoML if visualisations are prioritised for input-data statistics, FE processes and final model evaluations, although, curiously, the sampled data scientists were less concerned about the internals of the learning process, including HPO~\citep{drwe20}. Of course, these are but preliminary perspectives from the academic side. An accurate picture of social engagement with AutoML, along with recommendations to promote it, will require a much broader review of practical application and uptake.
\section{Conclusions}
\label{Sec:Conclusions}

In this monograph, we have surveyed a series of research threads related to AutoML, both well-established and prospective. In so doing, we have identified both benefits and obstacles that may arise from integrating them into one centralised architecture.

To support this review, we have also steadily built up a generic conceptual framework that illustrates how this integration may be achieved, all with the aim of supporting an effective data-driven learning system that needs minimal human interaction. The following is a summary of this process:
\begin{itemize}
    \item Section~\ref{Sec:Basics} introduced a fundamental `ML component', attempting to encapsulate a superset of low-level processes applicable to any ML task, e.g.~the data-driven algorithmic development of a model.
    \item Section~\ref{Sec:CASH} schematised a component-selection mechanism, identifying that the search for an optimal ML model/algorithm and associated hyperparameters is the core goal of AutoML. This section reviewed high-level optimisation strategies.
    \item Section~\ref{Sec:MCPS} upgraded the framework to revolve around an `ML pipeline', acknowledging that an ML model is more accurately described as a composition of transformations. This section reviewed the management of multi-component ML solutions.
    \item Section~\ref{Sec:NAS} described NAS as a neural-network specialisation of the multi-component framework detailed thus far. This section reviewed NAS.
    \item Section~\ref{Sec:Features} considered the segmentation of ML pipelines, specifically focussing on optimising pre-processing transformations. This section reviewed AutoFE.
    \item Section~\ref{Sec:Meta} schematised a meta-learning mechanism, designed to incorporate knowledge from previous ML experiments. This section reviewed the leveraging of meta-knowledge.
    \item Section~\ref{Sec:Ensemble} upgraded the framework to revolve around `ML-pipeline bundles', enabling high-level model combinations and, eventually, the ability to juggle ML pipelines between development and deployment. This section reviewed the management of ensembles.
    \item Section~\ref{Sec:Dynamic} upgraded the framework to strategise for dynamic contexts, defining persistent learning and adaptation as core prerequisites for AutonoML. This section reviewed managing both streamed data and concept drift.
    \item Section~\ref{Sec:Eval} elaborated the evaluation of an ML pipeline and associated mechanisms. This section reviewed metrics for model quality.
    \item Section~\ref{Sec:Resource} considered framework improvements to manage available hardware. This section reviewed automating operations subject to limited resources.
    \item Section~\ref{Sec:User} considered framework improvements to facilitate the inclusion of expert knowledge and user control. This section reviewed human-AI interactivity.
    \item Section~\ref{Sec:AGI} emphasised the flexibility of the framework to handle both arbitrary inflow data and ML tasks. This section reviewed attempts to generalise AutoML.
\end{itemize}
Finally, Section~\ref{Sec:Discussion} discussed AutoML/AutonoML from two angles: the overarching challenges of integration and the current state of mainstream engagement.

As this work is primarily intended to be a review of associated research fields, we have had to keep the depth of technical discussions balanced. Accordingly, while the illustrative architecture presented in this work is broadly encompassing and logically motivated, it should not necessarily be seen as an authoritative blueprint for an automated learning system; it is but a preliminary attempt to synthesise numerous topics that have not necessarily been previously combined.

Ultimately, our goal has simply been to show that AutonoML is, in principle, a feasible pursuit as of the current year, provided that a diverse array of concepts are identified and intelligently integrated. Now, granted, the field of AutoML, like deep learning, is thriving on its own in the modern era, especially with the availability of computational power, memory, and data. However, identifying and synthesising these concepts are not trivial; in our view, AutoML cannot properly evolve towards AutonoML without much deeper thought on this topic. The true challenges of venturing into the realm of complex adaptive systems, specifically through the development/emergence of an AutonoML framework as a living and perpetually evolving entity, still lie ahead. This review is intended to stimulate discussion on what is required and how we should approach such an endeavour.

\bibliographystyle{ACM-Reference-Format}

\end{document}